\documentclass[acmsmall]{acmart}
\usepackage{mydef}
\usepackage{dsfont}
\AtBeginDocument{%
  }

\setcopyright{acmlicensed}
\copyrightyear{2024}
\acmYear{2024}
\acmDOI{XXXXXXX.XXXXXXX}

\acmJournal{JACM}

\begin{document}

\title{Towards Graph Contrastive Learning: A Survey and Beyond}

\author{Wei Ju}
\email{juwei@pku.edu.cn}
\affiliation{%
  \institution{Peking University}
  \country{China}
  \postcode{100871}
}

\author{Yifan Wang}
\email{yifanwang@uibe.edu.cn}
\affiliation{%
  \institution{University of International Business and Economics}
  \country{China}
  \postcode{100029}
}

\author{Yifang Qin}
\email{qinyifang@pku.edu.cn}
\author{Zhengyang Mao}
\email{zhengyang.mao@stu.pku.edu.cn}
\affiliation{%
  \institution{Peking University}
  \country{China}
  \postcode{100871}
}

\author{Zhiping Xiao}
\email{patricia.xiao@cs.ucla.edu}
\affiliation{%
  \institution{University of California, Los Angeles}
  \country{USA}
  \postcode{90095}
}


\author{Junyu Luo}
\email{luojunyu@stu.pku.edu.cn}
\author{Junwei Yang}
\email{yjwtheonly@pku.edu.cn}
\author{Yiyang Gu}
\email{yiyanggu@pku.edu.cn}
\affiliation{%
  \institution{Peking University}
  \country{China}
  \postcode{100871}
}

\author{Dongjie Wang}
\email{wangdongjie@ku.edu}
\affiliation{%
  \institution{University of Kansas}
  \country{USA}
  \postcode{66045}
}

\author{Qingqing Long}
\email{qqlong@cnic.cn}
\affiliation{%
  \institution{Chinese Academy of Sciences}
  \country{China}
  \postcode{100190}
}



\author{Siyu Yi}
\email{siyuyi@mail.nankai.edu.cn}
\affiliation{%
  \institution{Nankai University}
  \country{China}
  \postcode{300071}
}

\author{Xiao Luo}
\email{xiaoluo@cs.ucla.edu}
\affiliation{%
  \institution{University of California, Los Angeles}
  \country{USA}
  \postcode{90095}
}


\author{Ming Zhang}
\email{mzhang_cs@pku.edu.cn}
\affiliation{%
  \institution{Peking University}
  \country{China}
  \postcode{100871}
}

\renewcommand{\shortauthors}{Ju et al.}

\begin{abstract}
In recent years, deep learning on graphs has achieved remarkable success in various domains. However, the reliance on annotated graph data remains a significant bottleneck due to its prohibitive cost and time-intensive nature. To address this challenge, self-supervised learning (SSL) on graphs has gained increasing attention and has made significant progress. SSL enables machine learning models to produce informative representations from unlabeled graph data, reducing the reliance on expensive labeled data. While SSL on graphs has witnessed widespread adoption, one critical component, Graph Contrastive Learning (GCL), has not been thoroughly investigated in the existing literature. Thus, this survey aims to fill this gap by offering a dedicated survey on GCL. We provide a comprehensive overview of the fundamental principles of GCL, including data augmentation strategies, contrastive modes, and contrastive optimization objectives. Furthermore, we explore the extensions of GCL to other aspects of data-efficient graph learning, such as weakly supervised learning, transfer learning, and related scenarios. We also discuss practical applications spanning domains such as drug discovery, genomics analysis, recommender systems, and finally outline the challenges and potential future directions in this field.
\end{abstract}

\ccsdesc[500]{Computing methodologies~Neural networks}
\ccsdesc[500]{Computing methodologies~Learning latent representations}

\keywords{Graph Contrastive Learning, Graph Neural Network, Deep Learning}


\maketitle

\section{Introduction}

Graph-structured data is ubiquitous and pervasive across various domains, ranging from social networks~\cite{alemany2022review,sharma2022survey} to recommender systems~\cite{ju2022kernel,qin2024learning,wu2022graph}, biological networks~\cite{zhang2024motif,fang2023knowledge}, and knowledge graphs~\cite{chen2023zero,yang2023poisoning}. With the rise in popularity and remarkable success of Graph Neural Networks (GNNs), deep learning on graphs has garnered significant attention in numerous fields~\cite{kipf2016semi,wu2020comprehensive,ju2024survey_realworld,ju2024comprehensive}. However, despite the widespread adoption of GNNs, a fundamental challenge persists – the majority of GNN models are tailored towards (semi-)supervised learning scenarios~\cite{kipf2016semi,gilmer2017neural,khosla2020supervised,luo2023rignn}. This necessitates access to a substantial amount of labeled data, which significantly restricts the applicability of graph deep learning methods in practice. This limitation is particularly pronounced in domains such as healthcare and molecular chemistry. In these fields, acquiring labeled data requires specialized expertise and extensive manual annotation efforts. Moreover, the graph data in these domains are typically limited, costly to acquire, or inaccessible. For instance, in healthcare, constructing patient interaction networks or disease progression graphs may require intricate knowledge of medical procedures and conditions, along with exhaustive documentation and annotation efforts~\cite{li2022graph_healthcare}. Similarly, in molecular chemistry, identifying the properties of compounds requires expertise in chemical synthesis and experimental validation, as well as extensive data collection and analysis resources.~\cite{ju2023few}.

To tackle the issues of scarce and inaccessible labeled data, self-supervised learning (SSL) has emerged as a promising solution~\cite{chen2020simple,he2020momentum,grill2020bootstrap,devlin2018bert,schiappa2023self}. SSL works by using pretext tasks to automatically extract meaningful representations from unlabeled data, thus reducing the reliance on manual annotations. By devising pretext tasks that exploit inherent structures within the data itself, SSL can uncover rich information from unannotated datasets, leading to improved model performance and generalization~\cite{jing2020self,liu2021self}. In recent years, SSL has made significant strides in computer vision (CV) and natural language processing (NLP), demonstrating promising prospects for future applications. SSL methods in CV leverage semantic invariance under image transformations to learn visual features. For instance, models like SimCLR~\cite{chen2020simple} and Moco~\cite{he2020momentum} focus on maximizing agreement between differently augmented views of the same image, enabling the model to capture robust and invariant features across variations. In NLP, SSL relies on language pretext tasks for pretraining. Recent advancements, notably exemplified by models like BERT~\cite{devlin2018bert}, utilize large-scale language models trained on tasks such as masked language modeling and next sentence prediction, achieving state-of-the-art performance across multiple tasks.

Inheriting the successes of SSL in CV and NLP, there's a growing interest in extending SSL to graph-structured data~\cite{hu2020gpt,velivckovic2019deep,hassani2020contrastive,qiu2020gcc,hu2019strategies,you2020does,luo2022clear}. However, applying SSL directly to graphs presents significant challenges. First, while CV and NLP primarily deal with Euclidean data, graphs introduce non-Euclidean complexities, making traditional SSL approaches less effective~\cite{wu2020comprehensive}. Second, unlike the independent nature of data points in CV and NLP, graph data intertwines through intricate topological structures, requiring innovative methodologies to effectively leverage these relationships~\cite{ju2024comprehensive,ju2024survey}. Thus, designing graph-specific pretext tasks that seamlessly integrate node features with graph structures becomes a crucial yet challenging concern. 

In recent years, several literature reviews on graph SSL have put forward a comprehensive framework~\cite{jin2020self,wu2021self_survey,liu2022graph,xie2022self}. They summarize a novel paradigm that emphasizes the use of carefully designed pretext tasks to extract meaningful graph representations efficiently. These reviews have categorized the pretext tasks into various types, such as \emph{contrastive-based, generative-based, and predictive-based approaches}. Contrastive-based SSL approaches aim to learn effective representations by comparing positive and negative examples in the embedding space~\cite{velivckovic2019deep,hassani2020contrastive,qiu2020gcc}. Generative-based SSL approaches focus on reconstructing input data and utilizing it as a supervisory signal, and aim to produce representations that capture underlying structures and patterns in the graph data~\cite{hu2020gpt,you2020does}. Predictive-based SSL techniques involve predicting certain aspects of the graph structure or node attributes, serving as auxiliary tasks to guide representation learning~\cite{hu2019strategies,peng2020self}.

Despite the comprehensive coverage provided by existing literature reviews on graph SSL paradigms, they often lack in-depth analysis of specific aspects. This deficiency may stem from the broad scope of the field and the multitude of techniques being developed concurrently. For example, graph contrastive learning (GCL) stands out as one of the most extensively researched paradigms currently. However, existing literature on graph SSL often only covers the fundamental principles of GCL, without fully exploring its potential in various contexts and downstream applications. 

Towards this end, in this survey, our primary focus is to provide a comprehensive overview of GCL. \textbf{Importantly, to the best of our knowledge, there has not been a dedicated survey specifically studying GCL}. The overall structure of the paper is described in Figure \ref{fig:overall_structure}. Technically, we initially summarize the fundamental principles of GCL in SSL, including augmentation strategies, contrastive modes, and contrastive optimization objectives. Following that, we explore the extensions of GCL into other aspects of data-efficient learning, such as weakly supervised learning, transfer learning, and other relevant scenarios. Additionally, we discuss real-world applications and outline the challenges and potential future directions in the field. The core contributions of this survey can be summarized as follows:

\begin{itemize}
    \item Research on GCL is extensive and continually gaining momentum. However, there is currently a lack of comprehensive reviews specifically focused on GCL research. By offering this overview, our goal is to address a critical gap in the literature and provide valuable insights.
    \item We provide a thorough elucidation of the foundational principles of GCL in SSL. This encompasses a detailed exploration of augmentation strategies, contrastive modes, and optimization objectives, shedding light on the core mechanisms that drive the effectiveness of GCL.
    \item We extend the exploration beyond to cover scenarios such as weakly supervised learning, transfer learning, and diverse data-efficient learning environments, highlighting GCL's capacity to enhance learning efficiency and effectiveness.
    \item We discuss real-world applications where GCL has been successfully employed, spanning fields such as drug discovery, genomics analysis, recommender systems, social networks and traffic forecasting, showcasing its practical relevance and impact.
    \item We highlight the challenges facing the field of GCL, while also outlining promising avenues for future research and development, indicating the exciting prospects that lie ahead.
\end{itemize}

\noindent\textbf{Differences from Existing Surveys.}
Our survey distinguishes itself by focusing exclusively on GCL within the field of graph SSL. 
Building on the comprehensive frameworks of Jin et al.~\cite{jin2020self} and Liu et al.~\cite{liu2022graph}, we delve into the often-neglected differences of GCL, such as augmentation strategies, contrastive modes, and optimization objectives within SSL.
To the best of our knowledge, in recent years, the surveys most related to ours are~\cite{wu2021self_survey,liu2022graph,zhu2021empirical}.
Specifically, Wu et al.~\cite{wu2021self_survey} and Liu et al.~\cite{liu2022graph} adopt a similar logic to introduce the graph SSL technique with subtle differences. They categorize graph SSL into different types and present associated mathematical summaries. 
Our survey, however, exclusively discusses the GCL domain with an in-depth analysis including not just theoretical foundations but also the practical implementations of GCL. 
Regarding~\cite{zhu2021empirical}, Zhu et al. investigate the key design elements of GCL through controlled experiments, contributing insights and the library to enhance GCL implementation.
In contrast, we broadly discuss the techniques of GCL in distinct realistic contexts like weakly supervised learning and transfer learning. We further introduce the real-world applications of GCL, such as drug discovery, and genomics analysis. 
It is important to note that the most recent survey~\cite{liu2022graph} covers methods only up to 2022, and does not include the latest developments from the past two years. Thus, this survey fills the gaps left by previous studies and provides a valuable resource for understanding the landscape of GCL.

\tikzstyle{leaf1}=[draw=black, 
    rounded corners,minimum height=1.2em,
    edge=black!10, 
    text opacity=1, align=center,
    fill opacity=.3,  text=black,font=\scriptsize,
    inner xsep=2pt, inner ysep=3.6pt,
    ]
\tikzstyle{leaf2}=[draw=black, 
    rounded corners,minimum height=1.2em,
    edge=black!10, 
    text opacity=1, align=center,
    fill opacity=.5,  text=black,font=\scriptsize,
    inner xsep=2pt, inner ysep=3.6pt,
    ]
\tikzstyle{leaf3}=[draw=black, 
    rounded corners,minimum height=1.2em,
    edge=black!10, 
    text opacity=1, align=center,
    fill opacity=.8,  text=black,font=\scriptsize,
    inner xsep=2pt, inner ysep=3.8pt,
    ]
\tikzstyle{leaf4}=[draw=black, 
    rounded corners,minimum height=1.2em,
    text width=4.4em, 
    edge=black!10, 
    text opacity=1, align=center,
    fill opacity=1,  text=black,font=\scriptsize,
    inner xsep=2pt, inner ysep=3.8pt,
    ]
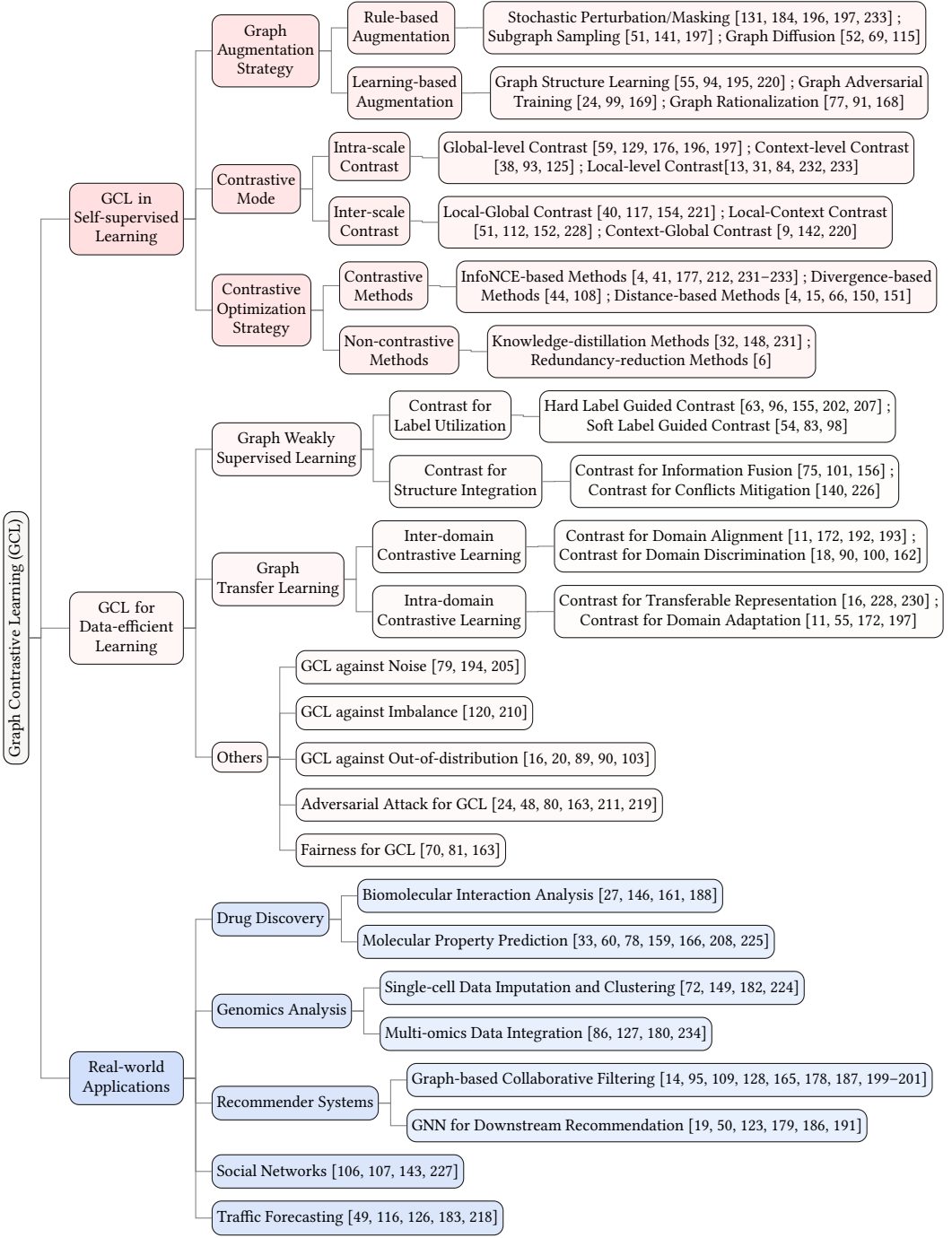
\begin{figure}[t]
\centering
\begin{forest}
  for tree={
  forked edges,
  grow=east,
  reversed=true,
  anchor=base west,
  parent anchor=east,
  child anchor=west,
  base=middle,
  font=\scriptsize,
  rectangle,
  draw=black, 
  edge=black!50, 
  rounded corners,
  minimum width=2em,
  s sep=5pt,
  inner xsep=3pt,
  inner ysep=2pt
  },
  [Graph Contrastive Learning (GCL),rotate=90,anchor=north,edge=black!50,fill=myred,fill opacity=1,draw=black,
    [GCL in \\ Self-supervised \\ Learning,leaf4,edge=black!50, fill=mypurple, minimum height=1.2em
        [Graph \\ Augmentation \\ Strategy,leaf3,edge=black!50,fill=mypurple, 
            [Rule-based \\ Augmentation,leaf2,fill=mypurple,
                [Stochastic Perturbation/Masking \cite{rong2019dropedge,you2020graph,xu2021infogcl,you2021graph,zhu2021graph}
                {;}\\
                Subgraph Sampling \cite{you2020graph,sun2021mocl,jiao2020sub}
                {;}
                Graph Diffusion \cite{kondor2002diffusion,page1998pagerank,jin2021multi}, leaf1,fill=mypurple
                ]
            ]
            [Learning-based \\ Augmentation,leaf2,fill=mypurple
                [Graph Structure Learning \cite{zhang2024motif,liu2022towards,yin2022autogcl,jin2023empowering}
                {;}
                Graph Adversarial \\ Training \cite{feng2022adversarial,luo2023self,wu2023graph}
                {;}
                Graph Rationalization \cite{li2022let,wei2023boosting,liu2024towards}
                ,leaf1,fill=mypurple
                ]
            ]
        ]
        [Contrastive \\ Mode,leaf3,edge=black!50,fill=mypurple, 
            [Intra-scale\\ Contrast,leaf2, fill=mypurple
            [Global-level Contrast \cite{you2020graph,you2021graph,ren2021label,xia2022simgrace,ju2023unsupervised}
            {;}
            Context-level Contrast \\ \cite{qiu2020gcc,han2022generative,liu2023multi}
            {;}
            Local-level Contrast\cite{zhu2020deep,zhu2021graph,chen2023attribute,liu2023b2,gong2023ma}
            ,leaf1,fill=mypurple
            ]
            ]
            [Inter-scale \\ Contrast,leaf2,fill=mypurple
            [Local-Global Contrast \cite{velivckovic2019deep,hassani2020contrastive,park2020unsupervised,zhang2023contrastive}
            {;}
            Local-Context Contrast \\ \cite{jiao2020sub,zhu2021transfer,mavromatis2020graph,tu2023hierarchically}
            {;} 
            Context-Global Contrast \cite{sun2021sugar,cao2021bipartite,zhang2024motif}
            ,leaf1,fill=mypurple
            ] 
            ]
        ]
        [Contrastive \\ Optimization \\ Strategy,leaf3,edge=black!50, fill=mypurple, 
            [Contrastive \\ Methods, leaf2,fill=mypurple
            [{InfoNCE}-based Methods \cite{bachman2019learning,zhu2021empirical,zhu2020deep,zhu2021graph,zhang2022localized,xia2022progcl,he2024new}
            {;} 
            Divergence-based \\ Methods \cite{manning1999foundations,hjelm2018learning}
            {;}
            Distance-based Methods \cite{bachman2019learning,chen2020simple,tian2020contrastive,tian2020makes,khosla2020supervised}
            ,leaf1,fill=mypurple
            ] 
            ]
            [Non-contrastive \\ Methods, leaf2,fill=mypurple  
            [Knowledge-distillation Methods \cite{thakoor2021bootstrapped,zhu2021empirical,grill2020bootstrap}
            {;}\\
            Redundancy-reduction Methods \cite{bardes2021vicreg}
            ,leaf1,fill=mypurple
            ]
            ]
        ]
    ]
    [GCL for \\ Data-efficient \\ Learning,leaf4,edge=black!50, fill=myyellow, minimum height=1.2em
        [Graph Weakly \\ Supervised Learning,leaf3,edge=black!50, fill=myyellow, minimum height=1.2em,
            [Contrast for \\ Label Utilization,leaf2,fill=myyellow
            [Hard Label Guided Contrast \cite{wan2021contrastive,long2021theoretically,ju2022kgnn,yue2022label,yu2023semi}
            {;}\\
            Soft Label Guided Contrast \cite{liu2021noise,lu2023pseudo,jin2021node}
            ,leaf1,fill=myyellow
            ]
            ]
            [Contrast for \\ Structure Integration,leaf2,fill=myyellow
            [Contrast for Information Fusion \cite{luo2022dualgraph,wan2021contrastive_2,li2021comatch}
            {;}\\
            Contrast for Conflicts Mitigation \cite{sun2019infograph,zhou2023smgcl}
            ,leaf1,fill=myyellow
            ]
            ]
        ]
        [Graph \\ Transfer Learning,leaf3,edge=black!50, fill=myyellow, minimum height=1.2em,
            [Inter-domain \\Contrastive Learning,leaf2,fill=myyellow
                [Contrast for Domain Alignment \cite{yin2022deal,chen2023universal,yin2023coco,wu2022attraction}
                {;}\\ 
                Contrast for Domain Discrimination \cite{liu2023good,luo2023towards_node,wang2023cross,ding2021few}
                ,leaf1,fill=myyellow
                ]
            ]
            [Intra-domain \\Contrastive Learning,leaf2,fill=myyellow
                [Contrast for Transferable Representation \cite{zhu2023graphcontrol,zhu2021transfer,chen2022learning}
                {;}\\
                Contrast for Domain Adaptation \cite{you2020graph,jin2023empowering,wu2022attraction,chen2023universal}
                ,leaf1,fill=myyellow
                ]
            ]
        ]
        [Others,leaf3,edge=black!50, fill=myyellow, minimum height=1.2em,
            [GCL against Noise \cite{li2024contrastive,yuan2023learning,yin2023omg},leaf2,fill=myyellow]
            [GCL against Imbalance \cite{zeng2023imgcl,qian2022co},leaf2,fill=myyellow]
            [GCL against Out-of-distribution \cite{duan2023graph,liu2023good,luo2022deep,liu2023flood,chen2022learning},leaf2,fill=myyellow]
            [Adversarial Attack for GCL \cite{feng2022adversarial,zhang2022unsupervised,zhang2023graph,in2023similarity,lin2023certifiably,wang2022uncovering},leaf2,fill=myyellow]
            [Fairness for GCL \cite{wang2022uncovering,kose2022fair,ling2022learning},leaf2,fill=myyellow]
        ]
    ]
    [Real-world \\ Applications,leaf4,edge=black!50, fill=mygreen, minimum height=1.2em
        [Drug Discovery,leaf3,edge=black!50, fill=mygreen, minimum height=1.2em,
            [Biomolecular Interaction Analysis \cite{wang2024predicting,yao2023semi,tao2023prediction,gao2023similarity}
            ,leaf2,fill=mygreen]
            [
            Molecular Property Prediction \cite{li2022geomgcl,zheng2023casangcl,zang2023hierarchical,wang2023molecular,ju2023few,wang2022improving,gu2023hierarchical}
            ,leaf2,fill=mygreen]
        ]
        [Genomics Analysis,leaf3,edge=black!50, fill=mygreen, minimum height=1.2em,
            [Single-cell Data Imputation and Clustering \cite{xiong2023scgcl,lee2023deep,tian2023scgcc,zheng2024subgraph},leaf2,fill=mygreen]
            [Multi-omics Data Integration \cite{liu2024muse,rajadhyaksha2023graph,xie2023mtgcl,zong2022const},leaf2,fill=mygreen]
        ]
        [Recommender Systems,leaf3,edge=black!50, fill=mygreen, minimum height=1.2em,
            [Graph-based Collaborative Filtering \cite{liu2021contrastive,xia2022hypergraph,yu2022graph,yu2023xsimgcl,ren2023disentangled,mao2021simplex,yu2021self,wang2021self,chen2023heterogeneous,yang2022knowledge},leaf2,fill=mygreen]
            [GNN for Downstream Recommendation \cite{xia2021self,yang2023debiased,du2022socially,yi2022multimodal,qin2023disenpoi,jiang2023rcenr},leaf2,fill=mygreen]
        ]
        [Social Networks \cite{ma2022towards,zhou2023detecting,ma-etal-2022-open-topic,sun2022rumor},leaf3,edge=black!50, fill=mygreen, minimum height=1.2em,
        ] 
        [Traffic Forecasting \cite{zhang2023spatial,pan2023spatial,qu2023st,xu2023gan,ji2024contrastive},leaf3,edge=black!50, fill=mygreen, minimum height=1.2em,
        ] 
    ]
  ]
\end{forest}
\caption{An overview of the taxonomy for existing GCL models.}
\label{fig:overall_structure}
\end{figure}
\section{Preliminary}
To promote the introduction of GCL frameworks in the following sections, we first provide important notations for graphs, outline the principle of graph neural networks and the basic concept of contrastive learning, and introduce various downstream tasks of GCL in this section.

\subsection{Notations}
Let $G = (\mathcal{V}, \mathcal{E}, \mathbf{X})$ be an attribute graph, where $\mathcal{V} = \{ v_1, v_2, \dots, v_{|\mathcal{V}|} \}$ is the set of nodes, and $\mathcal{E} \subset \mathcal{V} \times \mathcal{V}$ is the set of edges, $\mathbf{X}$ represents the feature matrix. Each node $v_i$ in the graph is associated with an attribute vector $x_i \in \mathbb{R}^F$, where $F$ denotes the attribute dimension. The attribute vectors of all the nodes form the feature matrix of the graph as $\mathbf{X} \in \mathbb{R}^{|\mathcal{V}| \times F}$. The set of edges can also be described using an adjacency matrix $\mathbf{A} \in \mathbb{R}^{|\mathcal{V}| \times |\mathcal{V}|}$, where $\mathbf{A}_{ij}=1$ if $(v_i, v_j)\in\mathcal{E}$, while $\mathbf{A}_{ij}=0$ otherwise. Consequently, the graph can be equivalently expressed as $G=(\mathbf{A}, \mathbf{X})$.

\subsection{Graph Neural Network}
Graph Neural Networks (GNNs)~\cite{kipf2017semi,hamilton2017inductive,velivckovic2018graph,luo2023towards_node} have developed into a crucial backbone for learning representations of graph-structured data. The majority of popular GNN layers are based on the message-passing mechanism~\cite{gilmer2017neural}, iteratively aggregating information from neighboring nodes to capture the structural characteristics of the graph. 
Specifically, 
let $\mathbf{h}_v^{(l)}$ represents the embedding of $v \in G$ at layer $l$, and the propagation rule of message passing can be written as: 
\begin{equation}
    \mathbf{h}_v^{(l)}=\mathcal{U}^{(l)}\left(\mathbf{h}_v^{(l-1)}, \mathcal{A}^{(l)}\left(\left\{\mathbf{h}_u^{(l-1)}, \forall u \in \mathcal{N}(v)\right\}\right)\right),
\end{equation}
where $\mathcal{A}^{(l)}$ denotes the message aggregating function, $\mathcal{U}^{(l)}$ denotes the embedding updating function, and $\mathcal{N}(v)$ is the neighborhood of node $v$. After $L$ iterative propagations, we can obtain the node-level representation $\mathbf{h}_v^{(L)}$, while the graph-level representation $\mathbf{h}_G$ can be further generated by integrating all the node representations with the $\operatorname{READOUT}$ function, 
\begin{equation}
    \mathbf{h}_{G} = \operatorname{READOUT}(\{\mathbf{h}_v^{(L)}, \forall v \in \mathcal{V}\}),
\end{equation}
which can be a simple $\operatorname{sum}$ function or any other global graph pooling operations~\cite{zhang2018end,baek2020accurate}.

\subsection{Contrastive Learning}
Contrastive learning (CL)~\cite{chen2020simple,he2020momentum,caron2020unsupervised} has become a popular self-supervised learning approach, which explores inherent similarities and differences between different objects, thereby reducing the reliance on labeled data. The core idea of contrastive learning is to narrow the distance between similar objects (positive pairs) while widening the distance between dissimilar objects (negative pairs) in the embedding space. 

For instance, for an object $x_i$, a common CL paradigm involves first generating an augmented view of it, i.e., $x_i'$, through a kind of augmentation strategy that preserves semantics, such as rotations in the case of image data~\cite{chen2020simple}. The augmented view is treated as the positive sample, while the other objects $x_j (j \neq i)$ in the dataset are considered negative samples. Then a parameterized encoder $f_{\theta}$ is employed to map various objects and their augmented views into embeddings in the representation space, i.e., $\mathbf{h}_i$ and $\mathbf{h}_i^\prime, i \in \{1,2, \dots, B\}$, where $B$ denotes the batch size. Finally, the model can be optimized by minimizing a contrastive loss function $\mathcal{L}_{con}$, which maximizes the consistency between positive sample pairs while minimizing the consistency between negative sample pairs. The general loss objective can be written as,
\begin{equation}
    \theta^{*} =  \underset{\theta}{\arg\min} \frac{1}{B} \sum_{i=1}^{B} \mathcal{L}_{con} \left(\mathbf{h}_{i}, \mathbf{h}'_{i}\right).
 \end{equation}
A typical way for constructing a contrastive loss function involves leveraging the softmax function to distinguish between positive and negative samples, such as NT-Xent~\cite{chen2020simple} and InfoNCE~\cite{oord2018representation}, which can be formalized as follows,
\begin{equation}
    \mathcal{L}_{con} \left(\mathbf{h}_{i}, \mathbf{h}'_{i}\right) = -\log \frac{e^{\operatorname{sim}\left(\mathbf{h}_{i}, \mathbf{h}'_{i}\right) / \tau}}{e^{ \operatorname{sim}\left(\mathbf{h}_{i}, \mathbf{h}'_{i}\right) / \tau}+\sum_{j \neq i} e^{\operatorname{sim}\left(\mathbf{h}_{i}, \mathbf{h}_{j}\right) / \tau}},
 \end{equation}
where $\operatorname{sim}(\cdot,\cdot)$ denotes a function measuring the similarity between embeddings, 
$\tau$ denotes the temperature parameter that controls the sensitivity of penalties on hard negative samples~\cite{wang2021understanding}.

\subsection{Downstream Tasks}

The graph encoder optimized by contrastive learning can map graphs into informative and highly discriminative vector representations,  facilitating the application to various downstream tasks. Based on the granularity of the prediction targets, these downstream tasks can be broadly categorized into node-level tasks and graph-level tasks.

\smallskip\textbf{Node-level tasks.} These tasks focus on predicting properties of nodes and relationships among nodes, including node classification~\cite{kipf2016semi}, node ranking~\cite{agarwal2006ranking}, node clustering~\cite{yi2023redundancy}, and link prediction~\cite{qin2024polycf}. For instance, the goal of node clustering is to partition the $n$ unlabeled nodes in a graph into $K$ disjoint clusters $\{C_1, \ldots, C_K\}$, where nodes within the same cluster exhibit similar semantics while those in different clusters have distinct semantics.
Among these node-level tasks, prediction problems involve pairwise relationships between nodes, such as predicting the existence of relationships (edges) between two nodes and predicting the type of these relationships, also referred to as edge-level tasks. The most typical edge-level prediction task is link prediction~\cite{zhang2018link}, whose objective is to forecast unobserved connections between nodes in a partially observed graph. Formally, given a graph $\mathcal{G}=(\mathcal{V},\mathcal{E}, \mathbf{X})$ with partial edges $\mathcal{E}_\mathcal{O} \subset \mathcal{E}$ observed, the model is expected to infer whether an edge exists between node $v_i$ and $v_j$, where $(v_i, v_j) \notin \mathcal{E}_\mathcal{O}$.

\smallskip\textbf{Graph-level tasks.} These tasks focus on predicting properties, relationships, and structures for entire graphs, including graph classification~\cite{luo2023towards,ju2024hypergraph}, graph matching~\cite{fey2019deep}, graph generation~\cite{zhu2022survey,wang2022disencite}, and graph-level clustering~\cite{ju2023glcc}. For example, graph classification aims to predict the property category to which an entire graph belongs. Formally, given a training dataset $\mathcal{D} = \{G_i, y_i\}_{i=1}^{N}$, where $y_i$ is the category label for graph $G_i$, graph classification aims to learn a model to predict the class $\hat{y}_j$ for each graph $G_j$ in the test set precisely. For graph generation, the goal is to learn the distribution $p(\mathcal{G})$ of given graph data $\mathcal{G}=\{G_i\}_{i=1}^{N}$, and then sample new graphs from this distribution $G_{\text{new}} \sim p(\mathcal{G})$, which is beneficial for accelerating the process of drug discovery~\cite{shi2020graphaf}.


\section{GCL in Self-supervised Learning}

\begin{figure*}[t]
    \centering
    \includegraphics[width=\linewidth]{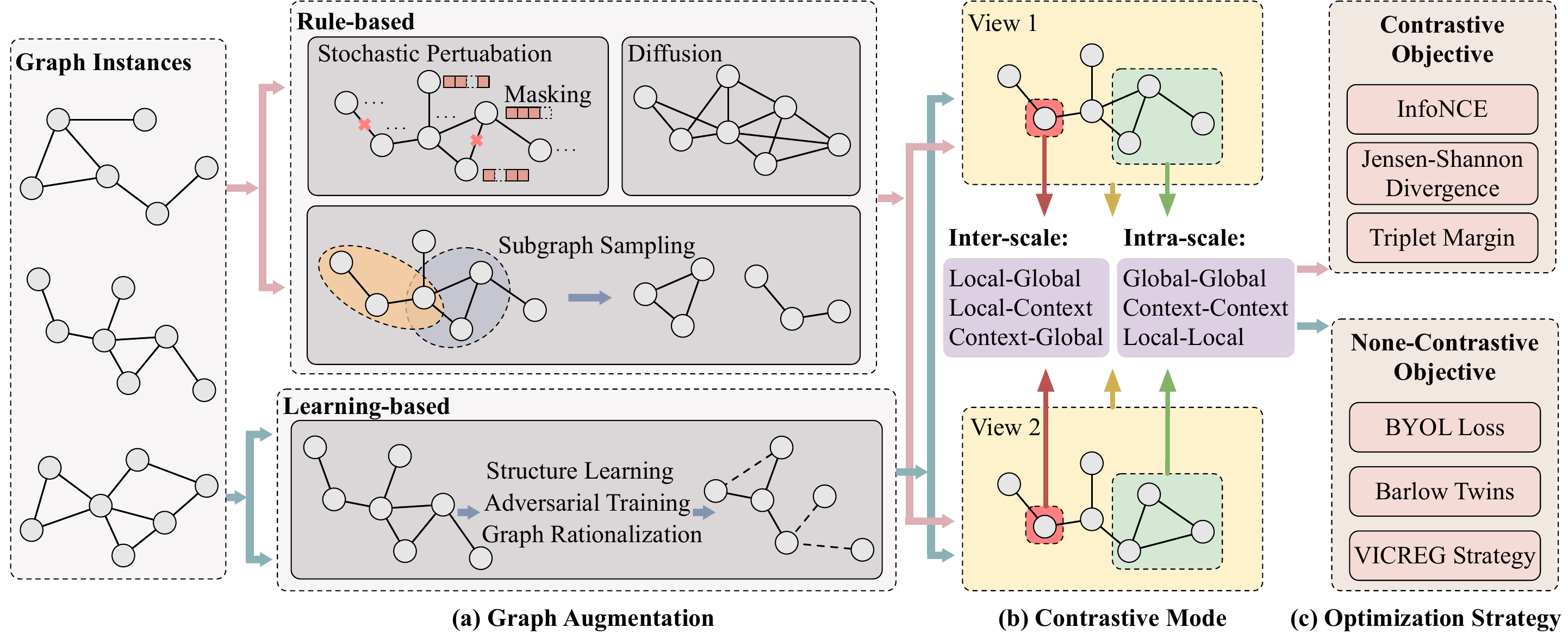}
    \caption{The general framework of graph contrastive learning (GCL). A contrastive method can be determined by defining its data augmentation strategy to generate different views, contrastive mode for the alignment between instances at different scales, and corresponding different contrastive optimization strategies.}
    \label{fig:gcl_framework}
\end{figure*}

In this section, we systematically categorize the basic principles of GCL in SSL, including augmentation strategies, contrastive modes, and contrastive optimization objectives, and present the associated mathematical summaries. The overall framework is shown in Fig.~\ref{fig:gcl_framework}.

\label{section::self_supervised}

\subsection{Graph Augmentation Strategy}
Graph augmentation aims to generate compatible, identity-maintaining positive examples of the specified graph for GCL. Despite significant advancements in data augmentation for contrastive learning within computer vision and natural language processing, directly transferring these methods to the graph domain is challenging due to the non-Euclidean, and irregular structure of graph data. Current graph augmentation strategies can be roughly divided into \emph{ruled-based} and \emph{learning-based} approaches, depending on whether learning is involved in the augmentation process. The graph augmentation strategies in GCL are summarized in Table~\ref{tab:augmentation}. 

\subsubsection{Rule-based Augmentation.}
A commonly used approach for graph-data augmentation is modifying the graph data following pre-defined rules. Depending on their complexity, rules for graph augmentation can vary from simple strategies such as \emph{stochastic perturbation/masking}~\cite{rong2019dropedge,you2020graph,xu2021infogcl,thakoor2022large} or randomly \emph{sampling subgraphs}~\cite{you2020graph,qiu2020gcc,jiao2020sub} to more sophisticated strategies that utilize \emph{graph diffusion} processes following specific diffusion kernels~\cite{gasteiger2019diffusion,kondor2002diffusion}.

\smallskip\textbf{Stochastic Perturbation/Masking.}
This category of methods perturbs/masks a certain fraction of nodes/edges and their corresponding features from the graph data to integrate diverse priors. Edge perturbation (dropping/adding) methods randomly remove/add a certain portion of edges from the input graph during training~\cite{rong2019dropedge,you2020graph,xu2021infogcl,thakoor2022large}, as follows:
\begin{equation}
    \tilde{\mathbf{A}}=(1-r)\cdot\mathbf{A}+r\cdot(1-\mathbf{A}),
\end{equation}
where $r$ denotes the perturbation ratio. Similarly, the node-dropping method randomly removes a certain portion of nodes from the graph to get the augmented graph~\cite{you2020graph,xu2021infogcl}. Instead of manipulating the graph structure, the feature masking method randomly masked off node features for the augmentation~\cite{you2020graph,xu2021infogcl,thakoor2022large}, defined as:
\begin{equation}
    \tilde{\mathbf{X}} = (1-L)\odot \mathbf{X}+L\odot M,
\end{equation}
where $M$ and $L$ denote the mask and location indicator matrix. Given masking ratio $r$, $L_{vi}$ ($i$-th elements of node $v$) is set to 1 with probability $r$ and 0 with probability $1-r$. Besides, DGI~\cite{velivckovic2019deep} also employs a corruption function to get the augmented graph by a row-wise shuffling of the node feature matrix, formulated as follows: $\tilde{X}=X[idx,:]$, where $idx$ is a random arrangement list from 1 to $\mathcal{V}$. To avoid manually selecting augmentations, JOAO~\cite{you2021graph} formulates a bi-level optimization framework for augmentation strategy selection for GCL. GCA~\cite{zhu2021graph} identifies important edges and feature dimensions for adaptive edge dropping and feature masking.

\smallskip\textbf{Subgraph Sampling.}
The key idea is to derive augmented graph instances from the input graphs that optimally retain certain desired properties. This is achieved by maintaining a subset of nodes and their associated connections, formulated as:
\begin{equation}
    \tilde{\mathbf{A}}, ~~\tilde{\mathbf{X}} =\mathbf{A}[\mathcal{S}, \mathcal{S}], ~~\tilde{\mathbf{X}}[\mathcal{S}, :],
\end{equation}
where $\mathcal{S}\subset\mathcal{V}$ represents the node subset. GraphCL~\cite{you2020graph} and GCC~\cite{qiu2020gcc} use subgraphs as contrastive instances and sample subgraphs for augmentation with random walks on the graph. SUBG-CON~\cite{jiao2020sub} samples central nodes and their surrounding nodes as the subgraph for the node-wise augmentation. MoCL~\cite{sun2021mocl} utilizes biomedical domain knowledge to generate augmented molecular graphs by substituting substructures like functional groups. 

\smallskip\textbf{Graph Diffusion.}
This kind of strategies create the augmented graph with the global structural knowledge of the input graph, which establishes connections between nodes and their indirectly connected neighbors. A generalized graph diffusion process can be formulated as:
\begin{equation}
    \tilde{\mathbf{A}}=\sum\nolimits_{k=0}^{\infty}\theta_k \mathbf{T}^k,
\end{equation}
where $\mathbf{T}\in\mathcal{R}^{|\mathcal{V}|\times |\mathcal{V}|}$ is the transformation matrix derived from adjacent matrix $\mathbf{A}$ (e.g., $\mathbf{A}\mathbf{D}^{-1}$ or $\mathbf{D}^{-\frac{1}{2}}\mathbf{A}\mathbf{D}^{-\frac{1}{2}}$). $\theta_k$ denotes the global-local coefficient, pre-defined by specific diffusion variants such as heat kernel~\cite{kondor2002diffusion} (i.e., $\theta_k=e^{-t}\frac{t^k}{K!}$) and Personalized PageRank (PPR)~\cite{page1998pagerank,hassani2020contrastive,jin2021multi} (i.e., $\theta_k=\alpha(1-\alpha)^k$), where $\alpha$ and $t$ is the random walk teleport probability and diffusion time, respectively.

\begin{table}[t]
\caption{Summary of graph augmentation strategies for GCL.}
\label{tab:augmentation}
\centering
\setlength{\tabcolsep}{2pt}
\resizebox{0.85\columnwidth}{!}{
\begin{tabular}{lcc}
\toprule
Method & Augmented Data & Core Idea \\
\midrule
DropEdge~\cite{rong2019dropedge} & Edge & Stochastic Perturbation/Masking \\
GraphCL \cite{you2020graph} & Node/Edge/Feature/SubGraph & Stochastic Perturbation/Masking \\
BGRL~\cite{thakoor2022large} & Node/Edge & Stochastic Perturbation/Masking \\
DGI~\cite{velivckovic2019deep} & Feature & Stochastic Perturbation/Masking \\
JOAO~\cite{you2021graph} & Node/Edge/Feature/SubGraph & Stochastic Perturbation/Masking \\
GCA~\cite{zhu2021graph} & Edge/Feature & Stochastic Perturbation/Masking  \\
GCC~\cite{qiu2020gcc} & Subgraph & Subgraph Sampling \\
SUBG-CON~\cite{jiao2020sub} & Subgraph & Subgraph Sampling \\
MOCL~\cite{sun2021mocl} & Subgraph & Subgraph Sampling \\
PPR~\cite{hassani2020contrastive} & Graph &  Graph Diffusion \\
MERIT~\cite{jin2021multi} & Edge/Feature/Graph/SubGraph &  Graph Diffusion \\
MICRO-Graph~\cite{zhang2024motif} & SubGraph &  Graph Structure Learning \\
DeCA~\cite{li2022graph} & Graph & Graph Structure Learning \\
SUBLIME~\cite{liu2022towards} & Graph & Graph Structure Learning \\
AutoGCL~\cite{yin2022autogcl} & Graph & Graph Structure Learning \\
AD-GCL~\cite{suresh2021adversarial} & Graph & Graph Adversarial Training \\
ARIEL~\cite{feng2022adversarial} & Graph & Graph Adversarial Training \\
GraphACL~\cite{luo2023self} & Graph & Graph Adversarial Training \\
RGCL~\cite{li2022let} & Graph & Graph Rationalization \\
GCS~\cite{wei2023boosting} & Graph & Graph Rationalization \\
\bottomrule
\end{tabular}
}
\end{table}

\subsubsection{Learning-based Augmentation}
Rather than relying on stochastic rule-based augmentations, which may introduce noise into GCL, learning-based graph augmentation approaches aim to learn augmentation strategies in a data-driven manner, which include \emph{graph structure learning}, \emph{graph adversarial training}, and \emph{graph rationalization} methods.

\smallskip\textbf{Graph Structure Learning.}
Since the graph structures are often incomplete and noisy, graph structure learning treats graph structure as learnable parameters and endeavors to identify an augmented graph for the downstream task:
\begin{equation}
    \tilde{\mathbf{A}}, ~~\tilde{\mathbf{X}} = \mathcal{T}(\mathbf{A},\mathbf{X}),
\end{equation}
where $\mathcal{T}(\cdot)$ represents the graph structure learning function. MICRO-Graph~\cite{zhang2024motif} formulates graph motif learning as a clustering process and leverages extracted graph motifs to sample subgraphs for GCL. DeCA~\cite{li2022graph} jointly learns graph community partition and graph representation considering the community structures. SUBLIME~\cite{liu2022towards} and AutoGCL~\cite{yin2022autogcl} design a bootstrapping GCL mechanism that automatically generates the representative graph view of the original graph. 

\smallskip\textbf{Graph Adversarial Training.}
Instead of learning optimal structures, graph adversarial training expects the model to withstand adversarial perturbations, leading to improved generalization:
\begin{equation}
    \tilde{\mathbf{A}}=\mathcal{T}'(\mathbf{A}), ~~\mathcal{T}'=\argmin_{\mathcal{T}}\mathcal{D}(f(\mathcal{T}(\mathbf{A})),f(\mathbf{A}),
\end{equation}
where $f{\cdot}$ and $\mathcal{D}(\cdot,\cdot)$ are the graph encoder and graph agreement discriminator. 
AD-GCL~\cite{suresh2021adversarial} employs an adversarial method to produce augmented graph views, reducing information redundancy from the original graph. 
ARIEL~\cite{feng2022adversarial} generates the adversarial graph view by maximizing the graph contrastive loss and formulates the generated view as hard training samples. GraphACL~\cite{luo2023self} introduces an adversarial generation branch that actively generates a bank of hard negative graph samples for GCL. GACN~\cite{wu2023graph} develops a graph view generator and discriminator to learn the augmented graph views through a min-max game and the views are further used for GCL.

\smallskip\textbf{Graph Rationalization.}
Rationale serves as a representative subset of a graph and is utilized independently for GCL, inherently making these methods interpretable. We define the rationale as:
\begin{equation}
    \tilde{\mathbf{A}}=\mathcal{T}(\mathbf{A}), ~~f(\mathcal{T}(\mathcal{A}))=f(\mathcal{A}).
\end{equation}
RGCL~\cite{li2022let} introduces a rationale generator that identifies discriminative subset nodes within the original graph, thereby generating rationale-aware views for GCL. GCS~\cite{wei2023boosting} proposes an iterative refinement procedure to identify the semantically discriminative structures of the graph depending on contrastive saliency. 

\subsection{Contrastive Mode}

GCL enhances the alignment between instances with similar semantic content by leveraging various graph views at different scales. These views range from graph-level view to subgraph-level view and down to node-level view, in increasing order of granularity. We further categorize the contrastive mode into \emph{intra-scale} and \emph{inter-scale contrast}. To be specific, \emph{intra-scale contrast} focuses on comparing two views at the same granularity, such as node versus node, maintaining consistency in scale across comparisons. Conversely, the two views in \emph{inter-scale contrast} have different levels of granularity, such as node versus graph, facilitating comparisons from local (node-level) to global (graph-level) perspectives within graph contexts. The contrastive mode methods in GCL are summarized in Table~\ref{tab:contrastive_mode}.

\subsubsection{Intra-scale Contrast}

The intra-scale contrastive learning approach can be subdivided into three distinct perspectives: global-level, context-level, and local-level intra-scale contrast.

\smallskip\textbf{Global-level Intra-scale Contrast.}
Methods within this category typically employ discrimination between graph representations~\cite{you2020graph,you2021graph,suresh2021adversarial,suresh2021adversarial,ren2021label,xia2022simgrace,ju2023unsupervised}. As an instance, GraphCL~\cite{you2020graph} utilizes four distinct graph augmentation methods to generate varied views. Formally, let $\mathbf{h}_{g_i}=f_{\theta}\left(\mathbf{A}_i, \mathbf{X}_i\right)$ denote the graph representation of the original view and $ \tilde{\mathbf{h}}_{g_i}=f_{\theta}\left(\tilde{\mathbf{A}}_i, \tilde{\mathbf{X}}_i\right)$ be the augmented representation. GraphCL employs contrastive loss denoted as $\mathcal{L}_{con}(\cdot)$ to draw the representations of two graphs that share similar views closer:
\begin{equation}
   \theta^{*} =  \underset{\theta}{\arg\min} \frac{1}{|\mathcal{G}|} \sum_{g_i \in \mathcal{G}} \mathcal{L}_{con} \left(\mathbf{h}_{g_i}, \tilde{\mathbf{h}}_{g_i}\right).
\end{equation}
JOAO~\cite{you2021graph} further improves GraphCL by introducing joint augmentation optimization, which optimizes both the selection of augmentations and the contrastive objectives simultaneously. AD-GCL~\cite{suresh2021adversarial} introduces adversarial edge dropping as an augmentation strategy to minimize the redundant information captured by encoders. 
Unlike previous methods that require data augmentations, SimGRACE~\cite{xia2022simgrace} uses the original graph as input and pairs a GNN model with its perturbed variant as dual encoders to obtain two correlated views for contrast without relying on data augmentations. HGCL~\cite{ju2023unsupervised} utilizes contrastive learning across multiple scales to capture the hierarchical structural semantics in graphs, and uses a Siamese network and momentum-update encoder to reduce the need for a large amount of negative samples.

\smallskip\textbf{Context-level Intra-scale Contrast.}
To explore the universal topological properties across various networks and uncover semantic relationships in multi-scale structures, researchers propose to perform contrast at the subgraph level~\cite{qiu2020gcc,han2022generative,liu2023multi}.
GCC~\cite{qiu2020gcc} introduces a pre-training task for graphs that focuses on identifying and transferring structural patterns across various graph networks. Initially, it extracts subgraphs from the $r$-hop ego networks of nodes. To elaborate, the set of $r$-hop neighbors surrounding a vertex $v_i$ is denoted as $N_{v_i} = \{u: d(u, v_i) \leq r\}$, with $d(\cdot,\cdot)$ representing the function that calculates the shortest path distance between nodes. The subgraph generated from $N_{v_i}$ is known as the $r$-ego network. GCC then uses contrastive learning to differentiate between subgraphs from a specific vertex and others, which empowers the GNN to capture and generalize universal patterns across various graphs. Han et al.~\cite{han2022generative} argue that manually crafted negative samples may fail to adequately represent a graph's local structures. Consequently, they introduce a framework GSC that combines adaptive subgraph generation with contrastive learning for self-supervised graph representation, employing optimal transport distance as a similarity metric to more effectively capture and differentiate the intrinsic structures of graphs. Similarly, MSSGCL~\cite{liu2023multi} also raises doubts about the efficacy of forming positive and negative pairs through graph augmentation. By employing subgraph sampling, it generates views at various scales and constructs diverse contrastive pairs according to semantic associations.

\smallskip\textbf{Local-level Intra-scale Contrast.}
Methods in this category primarily aim to learn node-level representations by aligning representations at the node level~\cite{zhu2020deep,zhu2021graph,chen2023attribute,liu2023b2,gong2023ma}.
For example, GRACE~\cite{zhu2020deep} initially creates views of the nodes based on both structural and attribute aspects, and the embeddings are denoted as $\mathbf{h}_{v_i}$ and $\mathbf{h}_{u_i}$. It then employs a contrastive loss to enhance the consistency between node embeddings across these views:
\begin{equation}
\mathcal{L}\left(\mathbf{h}_{v_i}, \mathbf{h}_{u_i}\right)=\log \frac{e^{\theta\left(\mathbf{h}_{v_i}, \mathbf{h}_{u_i}\right) / \tau}}{\underbrace{e^{\theta\left(\mathbf{h}_{v_i}, \mathbf{h}_{u_i}\right) / \tau}}_{\text {positive }}+\underbrace{\sum\nolimits_{k=1}^N \mathbb{I}_{[k \neq i]} e^{\theta\left(\mathbf{h}_{v_i}, \mathbf{h}_{w_k}\right) / \tau}}_{\text {inter- and intra-view negatives }}},
\end{equation}
where $\mathbf{h}_{w_k} = \{\mathbf{h}_{v_k}, \mathbf{h}_{u_k}\}$ indicates the inter- and intra-view negatives.
GCA~\cite{zhu2021graph} advances this field by incorporating an adaptive augmentation technique for graph-structured data, utilizing rich priors for both the topological and semantic information of the graph. Different from previous work that utilizes two correlated views for GCL, ASP~\cite{chen2023attribute} integrates three distinct graph views (original, attribute, and global views) into its framework for joint contrastive learning across these views. To address the challenge of selecting representative samples for more effective training, B2-sampling~\cite{liu2023b2} employs balanced sampling to identify the most representative negatives and biased sampling to correct the most error-prone negative pairs. 

\begin{table}[t]
\caption{Summary of contrastive mode for GCL. ``GIC", ``CIC”, and ``LIC” stand for global-level intra-scale, context-level intra-scale, and local-level intra-scale contrast, respectively. While ``LGC", ``LCC”, and ``CGC” represent local-global inter-scale, local-context inter-scale, context-global inter-scale contrast, respectively.}
\label{tab:contrastive_mode}
\centering
\setlength{\tabcolsep}{2pt}
\resizebox{0.9\columnwidth}{!}{
\begin{tabular}{lcc}
\toprule
Method & Pretext Task & Core Idea \\
\midrule
GraphCL \cite{you2020graph} & GIC & Augmentation-based graph-level contrast  \\
AD-GCL \cite{suresh2021adversarial} & GIC & Adversarial edge dropping  \\
SimGRACE \cite{xia2022simgrace} & GIC & Dual encoders \\
HGCL \cite{ju2023unsupervised} & GIC &  Multi-scale hierarchical contrast \\
GCC \cite{qiu2020gcc} & CIC &  Sample ego networks; Contrast among sub-graphs \\
GSC \cite{han2022generative} & CIC &  Adaptive subgraph generation \\
MSSGCL \cite{liu2023multi} & CIC &  Multi-scale subgraph sampling \\
GRACE \cite{zhu2020deep} & LIC & Use structural and attribute views  \\
GCA \cite{zhu2021graph} & LIC & Adaptive augmentation using rich priors  \\
ASP \cite{chen2023attribute} & LIC & contrast among three views  \\
B2-sampling \cite{liu2023b2} & LIC & Balanced and biased sampling  \\
DGI \cite{velivckovic2019deep} & LGC & Contrast node and graph embeddings\\
MVGRL \cite{hassani2020contrastive} & LGC & Use graph diffusion to create structural views\\
CGKS \cite{zhang2023contrastive}  & LGC & Graph pyramid of coarse-grained views\\
SUBG-CON \cite{jiao2020sub}  & LCC & Contrast central nodes and their adjacent subgraphs \\
EGI \cite{zhu2021transfer}  & LCC & Enhance MI between structural and node embeddings \\
GIC \cite{mavromatis2020graph}  & LCC & Contrast nodes and their cluster embeddings \\
SUGAR \cite{sun2021sugar}  & CGC & Distill subgraph patterns into a sketched graph version  \\
BiGI \cite{cao2021bipartite}  & CGC & Optimize subgraph-graph MI in bipartite graphs  \\
MICRO-Graph \cite{zhang2024motif}  & CGC & Employ a sampler to identify informative subgraphs  \\

\bottomrule
\end{tabular}
}
\end{table}

\subsubsection{Inter-scale Contrast}
Inter-scale approaches discriminate between different levels of graph topologies, which contain three sub-categories: local-global, local-context, and context-global contrast, each targeting contrasts at varying granularities.

\smallskip\textbf{Local-Global Inter-scale Contrast.}
This category of methods focuses on capturing both local and global information, aiming to maximize the mutual information between these two scales of representations~\cite{velivckovic2019deep,hassani2020contrastive,park2020unsupervised,zhang2023contrastive}.
DGI~\cite{velivckovic2019deep} pioneers the approach of contrasting node-level embeddings against graph-level representations. It first uses two encoders to obtain graph and node embeddings, i.e. $\tilde{\mathbf{h}}_{G} = R(f_{\theta_1}(\tilde{\mathbf{A}}, \tilde{\mathbf{X}}))$ and $\mathbf{H} = {f_{\theta_2}(\mathbf{A}, \mathbf{X})}$, where $R(\cdot)$ is the graph readout function. The goal is to maximize the mutual information between local representations and their corresponding global summaries:
\begin{equation}
    \theta_1^{*}, \theta_2^{*} = \underset{\theta_1, \theta_2}{\arg\min} \frac{1}{|\mathcal{V}|} \sum_{v_i \in \mathcal{V}} \mathcal{L}_{con} \left(\mathbf{h}_{G}, \mathbf{h}_{v_i}\right).
\end{equation}
MVGRL~\cite{hassani2020contrastive} employs graph diffusion to create an augmented structural view of a graph, and adopts a discriminator to contrast nodel-level and graph-level representations across views. 
To leverage the complex and hierarchical nature of graphs, CGKS~\cite{zhang2023contrastive} develops a graph pyramid of coarse-grained graph views, each crafted through a topology-aware graph coarsening layer. It then employs a novel joint optimization strategy featuring a pairwise contrastive loss to facilitate knowledge interactions across different scales of the graph pyramid.

\smallskip\textbf{Local-Context Inter-scale Contrast.}
Methods in this category mainly differ from local-global contrast by emphasizing the encoding of sampled subgraphs rather than the entire graph~\cite{jiao2020sub,zhu2021transfer,mavromatis2020graph,tu2023hierarchically}. SUBG-CON~\cite{jiao2020sub} leverages the significant relationship between central nodes and their adjacent subgraphs, aiming to capture structural information from regional neighborhoods and avoiding bias towards fitting the overall graph structure. Similarly, EGI~\cite{zhu2021transfer} focuses on training a GNN encoder to optimize the mutual information between defined structural information and node embeddings, which facilitates the extraction of high-level transferable knowledge from graphs. Besides contrasting local-global views, GIC~\cite{mavromatis2020graph} also aims to enhance the mutual information between node representations and their cluster embeddings, which is achieved by grouping nodes according to their embeddings and adjusting them to be more aligned with their cluster summaries. HCHSM~\cite{tu2023hierarchically} identifies a limitation in earlier GNN pre-training strategies, where treating all nodes equally can result in neglecting the significant yet challenging boundary samples. Therefore, it implements a multi-hierarchical contrastive strategy that incorporates multiple levels of intrinsic graph features, thereby enhancing the ability to distinguish representations of difficult samples.

\smallskip\textbf{Context-Global Inter-scale Contrast.}
Another notable category in inter-scale contrast is context-global contrast, which seeks to enhance the mutual information between subgraph representations and the overall graph representation~\cite{sun2021sugar,cao2021bipartite,zhang2024motif}.
SUGAR~\cite{sun2021sugar} distills essential subgraph patterns from the original graph into a sketched version, and employs a self-supervised approach that enhances subgraph embeddings to reflect global graph structures through mutual information maximization. BiGI~\cite{cao2021bipartite} generates a comprehensive global representation from two prototype representations and employs a subgraph-level attention mechanism to encode sampled edges into local representations. It enhances the global significance of nodes within a bipartite graph by optimizing the mutual information between their local and global representations. MICRO-Graph~\cite{zhang2024motif} tackles the challenge of subgraph-level contrast by identifying and sampling semantically significant subgraphs, particularly in molecular graphs. It then employs a sampler to identify informative subgraphs, which are to train GNNs via graph-to-subgraph contrastive learning.

\subsection{Contrastive Optimization Strategy}

To optimize the GCL, we need to define a contrastive objective, which captures the positive samples' similarity, while showing the negative samples' discrepancy.
Following some recent reviews on self-supervised learning strategies, we categorize the strategies into \emph{contrastive} and \emph{non-contrastive methods} respectively~\cite{shwartz2023compress,zhu2021graph}. The \emph{contrastive methods} are those that require concrete negative samples, and the \emph{non-contrastive methods} are those that can optimize without any negative samples. The contrastive optimization strategies in GCL are summarized in Table~\ref{tab:optimization}.  

\subsubsection{Contrastive Methods}
These strategies require both positive and negative samples. 
Some methods are based on {InfoNCE}~\cite{oord2018representation,bachman2019learning}, some of them are based on divergence of the distributions~\cite{manning1999foundations}, while some others are based on direct distance of positive versus negative samples~\cite{shah2022max}.

\smallskip\textbf{{InfoNCE}-based Methods.}
From data $x$ and its context $c$, directly modeling $p(x|c)$ might be sub-optimal, since the model can be distracted by information unrelated to the task goal. 
Therefore, some researchers seek to maximally preserve the mutual information between $x$ and $c$,
\begin{equation}\label{eq:mutual_info_infoNCE}
    I(x;c) = \sum\nolimits_{x,c} p(x, c) \log \frac{p(x|c)}{p(x)}\,.
\end{equation}
And they propose {InfoNCE}~\cite{oord2018representation,bachman2019learning} based on the Noise-Contrastive Estimation (NCE) technique~\cite{gutmann2010noise}, which has long been widely utilized in various fields. 
InfoNCE-based methods typically use random perturbation to generate augmented views. 
In the augmented views, InfoNCE regards any two nodes that are generated from the same source node as positive pairs, and any other pairs as negative pairs.
{InfoNCE} can handle tasks in other fields as well, such as speech, image, and text~\cite{oord2018representation}.
Given a set of $N$ random samples $X = \{ x_1, \dots, x_N \}$, where it contains only one positive sample and $N-1$ negative samples from the proposed distribution $p(x)$, {InfoNCE} optimizes:
\begin{equation}\label{eq:objective_infoNCE}
    \mathcal{L}_N(x) = -\mathbb{E}_{X \sim \{p(x)\}^N}\Big[ \log\frac{f(x,c)}{\sum_{x_j \in X} f(x_j, c)} \Big]\,,
\end{equation}
and optimizing the loss in Equation~\eqref{eq:objective_infoNCE} is equivalent with estimating the density ratio
\begin{equation}
    f(x,c) \propto \frac{p(x|c)}{p(x)}\,.
\end{equation}

To be more specific, in GCL, for every node $\mathbf{v}_i$ with positive set $\mathcal{P}(\mathbf{v}_i)$ where we sample $P$ positive samples from $\mathcal{P}(\mathbf{v}_i)$, and negative set $\mathcal{Q}(v_i)$ where we get the $Q$ negative samples, {InfoNCE} is often implemented as
\begin{equation}\label{eq:concrete_objective_infoNCE}
    \mathcal{L}(\mathbf{v}_i) = -\frac{1}{P}\sum_{\mathbf{p}_j \in \mathcal{P}(\mathbf{v}_i)}\log \frac{e^{\theta(\mathbf{v}_i, \mathbf{p}_j) / \tau}}{e^{\theta(\mathbf{v}_i, \mathbf{p}_j) / \tau} + \sum_{ \mathbf{q}_j \in \mathcal{Q}(\mathbf{v}_i) } e^{\theta(\mathbf{v}_i, \mathbf{q}_j) / \tau} }\,,
\end{equation}
where the function $\theta$ measuring the embeddings' similarity can be implemented as cosine similarity.

It is demonstrated that {InfoNCE} is able to provide robust and effective performance improvements under many contrastive learning settings~\cite{zhu2021empirical}. However, the limitation it faces is that it still needs many negative samples.
There are many works that improve the standard InfoNCE approach. For example, GRACE~\cite{zhu2020deep} and GCA~\cite{zhu2021graph} work on improving data augmentation strategies, Local-GCL~\cite{zhang2022localized} and ProGCL~\cite{xia2022progcl} improve its sampling approaches, while PiGCL~\cite{he2024new} alleviate the influence of implicit conflicts from the negative samples.

\smallskip\textbf{Divergence-based Methods.}
Some strategies compare the divergence between the positive samples' distribution and the negative samples'. 
When defining the graph contrastive optimization objective, it is common to use Jensen-Shannon Divergence (JSD).
JSD is also called Information Radius ({IRad})~\cite{manning1999foundations}. It is an extension of the Kullback–Leibler divergence (KLD) with non-negligible differences, such as the fact that JSD is symmetrized, smoothed, and always has a finite value. JSD of distributions $X$ and $Y$ is
\begin{displaymath}
    \mathrm{JSD}(X \Vert Y) = \frac{1}{2} \mathrm{KLD}(X \Vert M) + \frac{1}{2} \mathrm{KLD}(Y \Vert M)\,,
\end{displaymath}
where $M = \frac{1}{2}(X + Y)$ is a mixture distribution of $X$ and $Y$.
Using JSD to define the GCL objective:
\begin{equation}
    \mathcal{L}_\mathrm{JSD}(\mathbf{v}_i) = 
    \frac{1}{P} \sum_{\mathbf{p}_j \in \mathcal{P}(\mathbf{v}_i)}
    \log(d(\mathbf{v}_i, \mathbf{p}_j)) 
     +
    \frac{1}{Q} \sum_{\mathbf{q}_j \in \mathcal{Q}(\mathbf{v}_i)}
    \log(d(\mathbf{v}_i, \mathbf{q}_j))\,,
\end{equation}
where the discriminator function $d(\mathbf{a}, \mathbf{b})$ is usually inner product of $\mathbf{a}, \mathbf{b}$ followed by a sigmoid activation function.
Hjelm et al.~~\cite{hjelm2018learning} have proposed a variant of JSD objective, where the $\log(\cdot)$ function is replaced by a softplus function $\mathrm{sp}(x) = \log(1+e^x)$:
\begin{equation}
    \mathcal{L}_\mathrm{SP-JSD}(\mathbf{v}_i) = 
    - \frac{1}{P} \sum_{\mathbf{p}_j \in \mathcal{P}(\mathbf{v}_i)}
    \mathrm{sp}(- d(\mathbf{v}_i, \mathbf{p}_j)) 
     -
    \frac{1}{Q} \sum_{\mathbf{q}_j \in \mathcal{Q}(\mathbf{v}_i)}
    \mathrm{sp}( - d(\mathbf{v}_i, \mathbf{q}_j))\,,
\end{equation}

\smallskip\textbf{Distance-based Methods.}
Some other strategies compare the positive and negative samples directly.
For example, the {Triplet Margin} (TM) optimization strategy directly enforces an increase in the relative distance between positive and negative example pairs~\cite{schroff2015facenet}. TM objective is widely used for computing relative similarity between samples as well, especially in metric learning~\cite{schultz2003learning}.
In a problem where we define an anchor $\mathbf{v}_i$, a corresponding positive sample $\mathbf{p}_i$ and negative sample $\mathbf{q}_i$, the TM loss is formulated as:
\begin{equation}
    L(\mathbf{v}_i, \mathbf{p}_i, \mathbf{q}_i) = \max \{ d(\mathbf{v}_i, \mathbf{p}_i) - d(\mathbf{v}_i, \mathbf{q}_i) + \epsilon, 0 \}\,,
\end{equation}
where $\epsilon$ is a margin value, and $d(\mathbf{x}, \mathbf{y}) = \lVert \mathbf{x} - \mathbf{y} \rVert_p$ is a norm. To be more specific, the TM loss on a GCL sample's node $\mathbf{v}_i$ is usually
\begin{equation}
    \mathcal{L}_\mathrm{TM}(\mathbf{v}_i) 
    = \max \Big\{ \frac{1}{P} \sum_{\mathbf{p}_j \in \mathcal{P}(\mathbf{v}_i)} \lVert \mathbf{v}_i - \mathbf{p}_i \rVert
    - 
    \frac{1}{Q} \sum_{\mathbf{q}_j \in \mathcal{Q}(\mathbf{v}_i)} \lVert \mathbf{v}_i - \mathbf{q}_i \rVert
    + \epsilon, 0 
    \Big\}\,.
\end{equation}

Most of these strategies are linked to the InfoMax principle~\cite{linsker1988self}, which seeks to maximize the Shannon Mutual Information (MI) between the same set of nodes across different views. Methods based on {InfoNCE} and {JSD} have been shown to successfully approximate the lower bound of MI, whereas the TM strategy is empirically found to raise the mutual information between positive samples, without much strong theoretic guarantee~\cite{poole2019variational}.

Among those methods, {InfoNCE} achieves the best overall performance as long as there are sufficient negative samples~\cite{zhu2021empirical}, and therefore is widely adopted in recent studies on contrastive learning~\cite{bachman2019learning,chen2020simple,tian2020contrastive,tian2020makes,khosla2020supervised}.

\smallskip\textbf{Other Methods.}
There are many other graph contrastive optimization methods requiring negative samples, such as DGI~\cite{velivckovic2019deep}. DGI-based methods have a similar objective function as is defined in Equation~(\ref{eq:objective_infoNCE}), except that instead of sampling from any random negative samples, it samples from a target node's neighborhood. In addition, it enforces an inductive bias that adjacent nodes should have similar representations.
InfoNCE can be more easily extended to other fields than DGI, while DGI puts more emphasis on graph structure.

\begin{table}[t]
\caption{Summary of contrastive optimization strategies for GCL.}
\label{tab:optimization}
\centering
\resizebox{1\columnwidth}{!}{
\begin{tabular}{lcc}
\toprule
Method & Category & Core Idea \\
\midrule
{InfoNCE}~\cite{oord2018representation} & Contrastive Method & NCE-based, proved to follow {InfoMax} principle  \\
GRACE~\cite{zhu2020deep}  & Contrastive Method & InfoNCE-based  \\
GCA~\cite{zhu2021graph}  & Contrastive Method & InfoNCE-based  \\
GCL~\cite{zhang2022localized} & Contrastive Method & InfoNCE-based  \\
ProGCL~\cite{xia2022progcl} & Contrastive Method & InfoNCE-based  \\
PiGCL~\cite{he2024new} & Contrastive Method & InfoNCE-based  \\
DGI~\cite{velivckovic2019deep} & Contrastive Method & Follow {InfoMax} principle  \\
JSD (IRad)~\cite{manning1999foundations} & Contrastive Method & Distance-based, proved to follow {InfoMax} principle \\
SP-JSD~\cite{hjelm2018learning} & Contrastive Method & JSD-based, replacing log function with softplus \\
TM~\cite{schroff2015facenet} & Contrastive Method & Distance-based, no guarantee on {InfoMax}  \\
BYOL~\cite{grill2020bootstrap} & Non-contrastive Method &  Knowledge-distillation method  \\
BGRL~\cite{thakoor2021bootstrapped} & Non-contrastive Method & Graph-specific extension of BYOL \\
{Barlow Twins}~\cite{zbontar2021barlow} & Non-contrastive Method &  Redundancy-reduction method \\
{VICReg}~\cite{bardes2021vicreg} & Non-contrastive Method & Redundancy-reduction method  \\
\bottomrule
\end{tabular}
}
\end{table}


\subsubsection{Non-Contrastive Methods}
This kind of strategies do not require the existence of any concrete negative samples, and in some previous works, people refer to them as \emph{non-contrastive methods}~\cite{shwartz2023compress}.

In the absence of concrete negative samples, we usually focus on the agreements among the positive samples. There are various different ways: \emph{knowledge-distillation methods}, such as the extension of BYOL~\cite{grill2020bootstrap} on graph settings --- BGRL~\cite{thakoor2021bootstrapped}, and \emph{redundancy-reduction methods} such as the {Barlow Twins} ~\cite{zbontar2021barlow,bielak2022graph} and {VICReg}~\cite{bardes2021vicreg}.


\smallskip\textbf{Knowledge-distillation Methods.}
BYOL~\cite{grill2020bootstrap}
stands as a self-supervised optimization strategy initially tailored for handling non-contrastive image representation learning. It surpasses many advanced contrastive approaches without access to any negative samples.
BYOL operates with two neural networks, known as the online network and the target network, which mutually enhance each other's learning. The {online} network is sometimes referred to as a {teacher} network, whereas the {target} network is sometimes referred to as offline network~\cite{zhu2021empirical}.
BGRL~\cite{thakoor2021bootstrapped} adapts BYOL to graph settings, applying graph augmentations on every node in a coherent way, instead of that for BYOL on images, where it learns to predict the projection of each image independently.
BGRL starts from an augmented view $q(\mathbf{v}_i)$ of a positive sample $\mathbf{v}_i$, and trains the {online} network $q(\cdot)$ to predict the representation of the {target} network $\mathbf{v}_i'$ -- another augmented view of the same data sample.
Then the objective is simply to maximize their cosine similarity by minimizing
\begin{equation}\label{eq:BYOL}
    \mathcal{L}_\mathrm{BYOL} = - \frac{q(\mathbf{v}_i)^{\top} \mathbf{v}_i' }{\lVert q(\mathbf{v}_i) \rVert \lVert \mathbf{v}_i'\rVert}\,.
\end{equation}
When being applied to graph settings, BYOL-based methods such as BGRL are often symmetrized, in the sense that the online network is also applied to $\mathbf{v}_i'$, making predictions from the reversed direction~\cite{thakoor2021bootstrapped,zhu2021empirical}. To avoid resulting in a trivial solution where $q(\mathbf{v}_i) = \mathbf{v}_i'$, several extra constraints are added. It was found that strategies including asymmetric dual encoders, updating the {target} encoder through an exponential moving average, or employing batch normalization, are all useful tricks to prevent such model collapse~\cite{grill2020bootstrap}.


\smallskip\textbf{Redundancy-reduction Methods.}
{Barlow Twins} strategy is a typical method that conducts redundancy reduction on the cross-correlation matrix between features of the augmented views of a sample. Taking representations of the same node from two augmented views: $\mathbf{v} \in \mathbb{R}^d$ and $\mathbf{u} \in \mathbb{R}^d$ for instance, we have $d$-dimensional features. Then we compute correlation matrix $\mathbf{C} \in \mathbb{R}^{d \times d}$ between $\mathbf{v}$ and $\mathbf{u}$, and the {Barlow Twins} loss is then computed as:
\begin{equation}
    \mathcal{L}_\mathrm{Barlow-Twins} = \sum_i (1-\mathbf{C}_{ii})^2 + \lambda \sum_i \sum_{j \neq i} \mathbf{C}_{ij}^2\,,
\end{equation}
where the trade-off term $\lambda$ is a hyper-parameter.

Taking one step further, the {VICReg} loss~\cite{bardes2021vicreg} added the variance-invariance-covariance regularization terms on top of {Barlow Twins} loss. Therefore, the {VICReg} strategy is less sensitive to normalization tricks and more stable than {Barlow Twins} strategy in practice.

{VICReg} defines a loss term with three components: (1) the {invariance} component, which measures the mean square distance between vectors, (2) the {variance} component, which enforces the embeddings of samples within a batch to be different from each other, and (3) the {covariance} component, which will decorrelates the variables of each embedding within a batch, and prevent information collapse (i.e., avoid the variables being highly correlated).
\begin{equation}\label{eq:VICReg}
    \mathcal{L}_\mathrm{VICReg} = \lambda \mathcal{L}_\mathrm{inv}(\mathbf{v}, \mathbf{u}) + \mu \mathcal{L}_\mathrm{var}(\mathbf{v}, \mathbf{u}) + \nu \mathcal{L}_\mathrm{cov}(\mathbf{v}, \mathbf{u})\,,
\end{equation}
where $\lambda$, $\mu$ and $\nu$ are hyper-parameters defining the importance of each term. 
The {invariance} component $\mathcal{L}_\mathrm{inv}(\mathbf{v}, \mathbf{u})$ is computed as the mean-squared Euclidean distance of the vectors pair.
The {variance} component $\mathcal{L}_\mathrm{var}(\mathbf{v}, \mathbf{u}) = f_v(\mathbf{v}) + f_v(\mathbf{u})$ has $f_v$ being a hinge function on the standard
deviation of the embeddings along the batch dimension:
\begin{equation}
    f_v(\mathbf{x}) = \frac{1}{d} \sum_{j = 1}^d \max(0, \gamma - s(\mathbf{x}, \epsilon))\,,
\end{equation}
with the regularized standard deviation defined as $s(\mathbf{x}, \epsilon) = \sqrt{\mathrm{Var}(\mathbf{x}) + \epsilon}$, and $\epsilon$ is just a small scalar preventing numerical instabilities.
Finally, the {covariance} component $\mathcal{L}_\mathrm{cov}(\mathbf{v}, \mathbf{u}) = f_c(\mathbf{v}) + f_c(\mathbf{u})$ has a function $f_c$:
\begin{equation}
    f_c(\mathbf{x}) = \frac{1}{d} \sum_{i \neq j} [C(\mathbf{x})]_{i,j}^2 \,,
\end{equation}
whose design is inspired by {Barlow Twins} strategy. Here, $C(\mathbf{x})$ is the covariance matrix of $\mathbf{x}$.

\smallskip\textbf{Other Methods.}
The above-mentioned strategies 
have been tested in experiments to demonstrate their effectiveness~\cite{zhu2021empirical}, whereas other graph self-supervised learning strategies that worked in other fields can potentially help with GCL as well~\cite{liu2022graph}.

\section{Graph Contrastive Learning for Data-efficient Learning}
In this section, we extend GCL to other aspects of data-efficient learning, such as weakly supervised learning, transfer learning, and other related scenarios, to maximize the potential of GCL.

\subsection{Graph Weakly Supervised Learning}

\begin{figure}[t]
    \centering
    \includegraphics[width=0.7\linewidth]{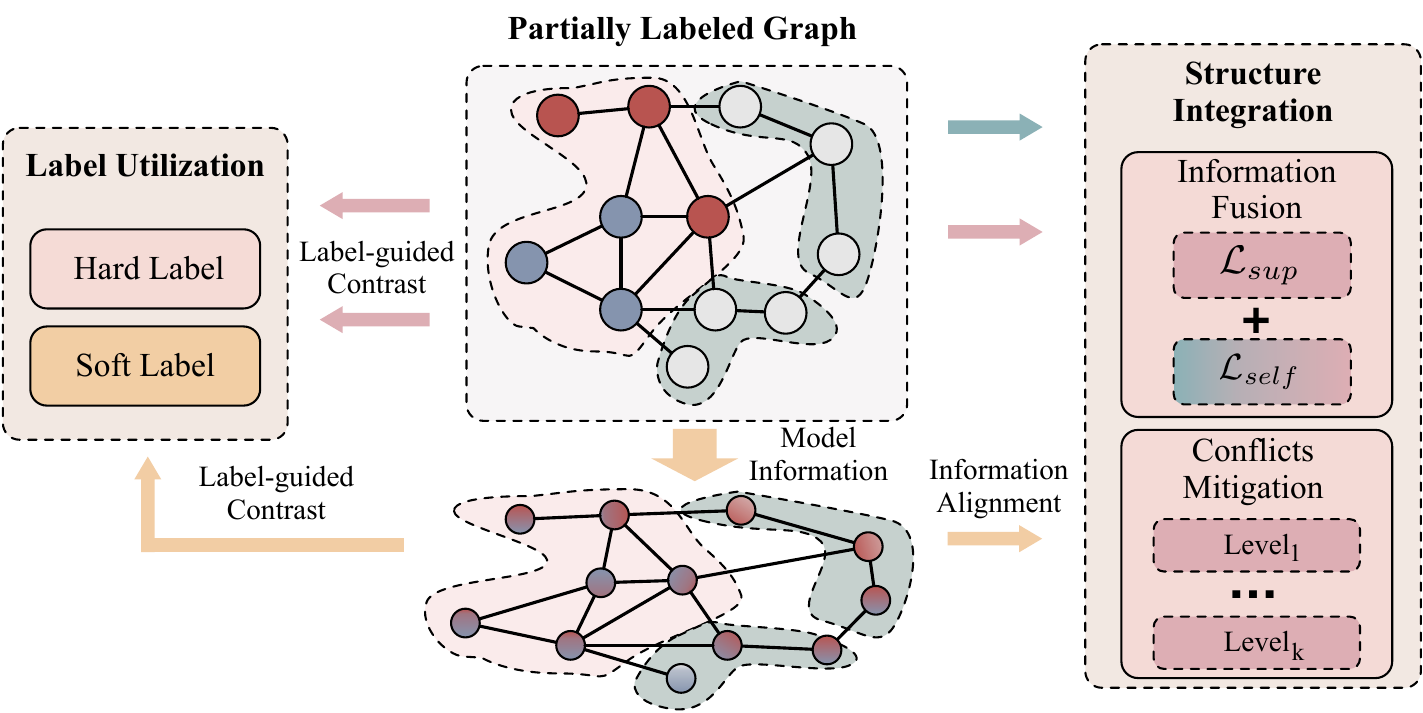}
    \caption{GCL in graph weakly supervised learning.}
    \label{fig:gcl-weakly}
\end{figure}

In many real-world scenarios, acquiring labeled data can be costly or impractical. In response to this challenge, Graph Weakly Supervised Learning (GWSL) has been proposed. GWSL utilizes both labeled and unlabeled data to enhance model performance in low-resource scenarios.
Current GWSL methods primarily utilize GCL to achieve two core functionalities: efficiently utilizing sparse labeled signals and fully exploiting unlabeled structural information as shown in Fig~\ref{fig:gcl-weakly}. The weakly supervised learning methods with GCL are summarized in Table~\ref{table:weakly}.

\subsubsection{Contrast for Label Utilization}

In this approach, the emphasis lies in attempting to utilize the sparse supervision information contained within labeled data more effectively during contrastive learning. 
Related methods can be categorized into two types: using \emph{hard labels} and using \emph{soft labels}.

\smallskip\textbf{Hard Label Guided Contrast.} In this strategy, as a representative work, CG$^3$~\cite{wan2021contrastive} performs hierarchical graph convolution and then conducts GCL within labeled graphs as:
\begin{equation}
    \mathcal{L}_{hard}\left(\mathbf{h}_i\right)=-\log \frac{\sum_{k} \mathds{1}_{\left\{y_i=y_k\right\}} \exp \left(\left\langle\mathbf{h}_i, \mathbf{h}_k\right\rangle\right)}{\sum_{j} \exp \left(\left\langle\mathbf{h}_i, \mathbf{h}_j\right\rangle\right)}.
\end{equation}
Here, $\mathds{1}_{\left\{\cdot\right\}}$ is the indicator function, which equals 1 if the condition holds. This simple operation enhances the discriminability of representation clusters corresponding to different labels (i.e., different $y$).
GSKN~\cite{long2021theoretically} and HGK-GNN~\cite{long2021hgk} are two representative structural Pseudo-Labeling GWSL works, which adopt structural kernel to capture graph topological structures and are applied in underlying contrastive learning.
KGNN~\cite{ju2022kgnn} utilizes a memory network for implicit contrastive learning, followed by the propagation of hard label information using posterior regularization:
\begin{equation}
    \mathcal{L}(\boldsymbol{\theta}, \boldsymbol{\phi})=\mathcal{L}(\boldsymbol{\theta})-\min _\phi \mathrm{KL}\left(q_\phi(y \mid x) \| p_\theta(y \mid x)\right),
\end{equation}
where $q_\phi$ is parameterized with a contrastive network and $p_\theta$ is parameterized with a traditional GNN.
GLA~\cite{yue2022label} utilizes GCL to learn label-invariant features for better propagating supervision signals between similar graphs.
Yu et al.~\cite{yu2023semi} propose a confidence-discriminative GNN, aiming to extract potential information from unlabeled data and only assign hard labels to confident data to reduce error accumulation.

\smallskip\textbf{Soft Label Guided Contrast.}
In order to better avoid the harmful bias that hard labels may introduce, a series of methods combining soft labels and contrastive learning have been proposed.
MCL~\cite{liu2021noise} uses a memory bank $\mathcal{M}$ to compute a soft prototype $\mathbf{w}$ for each class, and then utilizes prototypes to compute the similarity between each data $x_i$ and label $y_j$ as:
\begin{equation}
\mathrm{sim}(x_i,y_j) = \frac{\exp (\left<\mathbf{h}_i, \mathbf{w}_j\right>)}{\sum_k\exp (\left<\mathbf{h}_i, \mathbf{w}_k\right>)}, ~~\text{where}~~ \mathbf{w}_j = \mathop{\mathrm{Mean}}_{(\mathbf{h}_k,y_k)\in\mathcal{M},y_k=y_j}\frac{\mathbf{h}_k}{\|\mathbf{h}_k\|},
\end{equation}
and then optimizes the similarity calculation with contrastive learning methods.
PCL~\cite{lu2023pseudo} transforms the strong constraint ``a node belongs to a specific class" into a weaker constraint "two nodes do not belong to the same class'' to provide the model with better fault tolerance, and then utilizes a topologically weighted contrastive loss to better model the latter constraint.
SimP-GCN~\cite{jin2021node} designs a $k$-nearest-neighbor graph based on the node features to construct soft-label centers for improving representation learning.

\begin{table}[t]
\caption{Summary of GCL-based weakly supervised learning methods. ``HL" and ``SL" are for hard and soft label utilization. Meanwhile, ``IF" and ``CM" are for information fusion and conflict mitigation.}
\label{table:weakly}
\centering
\setlength{\tabcolsep}{1pt}
\resizebox{0.8\columnwidth}{!}{
\begin{tabular}{lcc}
\toprule
Method & Category & Core Idea \\
\midrule
CG$^3$~\cite{wan2021contrastive} & HL & Hard guidance function for GCL \\
GSKN~\cite{long2021theoretically} & HL & Label-guided and kernel-based topology contrasting \\
HGK-GNN~\cite{long2021hgk} & HL & Label-guided and kernel-based topology contrasting \\
KGNN~\cite{ju2022kgnn} & HL &  Posterior regularization for label utilization\\
GLA~\cite{yue2022label} & HL & Label-invariant features for label propagation\\ 
MCL~\cite{liu2021noise} & SL & Memory-based prototype calculation \\
PCL~\cite{lu2023pseudo} & SL & Transforming hard constraint to soft constraint\\
SimP-GCN~\cite{jin2021node} & SL & Calculating soft centers with neighbors \\
DualGraph~\cite{luo2022dualgraph} & IF & Iterative annotation and retrieval \\
CGPN~\cite{wan2021contrastive_2} & IF & Poisson Network based fusion \\
CoMatch~\cite{li2021comatch} & IF & Joint evolution for structure and label fusion \\
GAGED~\cite{zhang2022contrastive} & CM & Detecting and reducing errors \\
ASGN~\cite{hao2020asgn} & CM & Learning and selection \\
InfoGraph~\cite{sun2019infograph} & CM & Mutual information between nodes and graphs \\
SMGCL~\cite{zhou2023smgcl} & CM & Mutual information among multiple levels\\
\bottomrule
\end{tabular}
}
\end{table}

\subsubsection{Contrast for Structure Integration}

Methods in this category attempt to utilize GCL to explore the intrinsic structural information of graphs without label guidance, and then integrate it with label information.
Related approaches mainly investigate two issues: how to more fully integrate consistent information from both labels and structures to enhance representations, and how to reduce conflicts arising from inconsistent information between them.

\smallskip\textbf{Contrast for Information Fusion.}
These methods focus on how to better integrate information that is consistent between labels and structures.
The most prevalent fusion method is to directly add the label-determined supervised loss and the structure-determined self-supervised loss:
\begin{equation}
    \mathcal{L} = \mathcal{L}_{sup}+ \mathcal{L}_{self}.
\end{equation}
The method for computing $\mathcal{L}_{self}$ based on contrastive methods has been extensively discussed in Section \ref{section::self_supervised}, so we will not delve further into it here. Additionally, there are some unique fusion approaches.
DualGraph~\cite{luo2022dualgraph} iteratively utilizes annotation and retrieval to recognize and fuse consistent label and structure information. During this process, contrastive learning ensures that structure information consistently possesses discriminative representations.
CGPN~\cite{wan2021contrastive_2} designs a Poisson Network to propagate label information guided along the structure.
CoMatch~\cite{li2021comatch} learns two representations of label probabilities and contrasting structure embeddings, and then adopts a joint evolution approach to accomplish the information fusion.

\smallskip\textbf{Contrast for Conflicts Mitigation.}
This category of methods primarily focuses on reducing potential conflicts caused by inconsistent information from labels and structures.
GAGED~\cite{zhang2022contrastive} utilizes GCL to detect potential errors in node features and thereby reduce conflicts.
ASGN~\cite{hao2020asgn} adopts a strategy of learning and selection to minimize the gap between structures and labels.
An interesting observation is that there is often structural information at different levels in the graph. The inconsistency between labels and different levels of structure may vary, so aligning structures at different levels first is more advantageous to reducing conflicts with labels, formulated as:
\begin{equation}
    \mathrm{Fusion}(\mathrm{Labels}, \mathrm{Alignment}(\mathrm{level}_1, ..., \mathrm{level}_k)).
\end{equation}
Following this framework, InfoGraph~\cite{sun2019infograph} uses projection heads to align the information carried by nodes and graphs, reducing the potential bias impact. 
SMGCL~\cite{zhou2023smgcl} utilizes mutual information (MI) to align information at node-level, subgraph-level, feature-level, and topology-level simultaneously:
\begin{equation}
    \begin{aligned}
    \mathcal{L}(v_i)=&-\log \sigma\left(\operatorname{MI}\left(\mathbf{h}_i^{\prime}, \mathbf{U}_i^{\prime \prime}\right)\right)-\log \sigma\left(\operatorname{MI}\left(\mathbf{h}_i^{\prime \prime}, \mathbf{U}_i^{\prime}\right)\right)\\
    &- \log \left(1-\sigma\left(\operatorname{MI}\left(\mathbf{N h}_i^{\prime}, \mathbf{U}_i^{\prime \prime}\right)\right)\right)
    -\log \left(1-\sigma\left(\operatorname{MI}\left(\mathbf{N h}_i^{\prime \prime}, \mathbf{U}_i^{\prime}\right)\right)\right),
    \end{aligned}
\end{equation}
where $\mathbf{U}_i^{\prime}$ and $\mathbf{U}_i^{\prime \prime}$ are the representations of the subgraph related to node $i$ in two different views (feature view and topology view), $\mathbf{N u}_i^{\prime}$ and $\mathbf{N u}_i^{\prime \prime}$ are the negative representations of node $i$, which are generated by randomly disrupting the original node features.

\subsection{Graph Transfer Learning}

Graph transfer learning aims to enhance model performance when a graph domain shift exists between the source domain $\cD^{so}$ used for training and the target domain $\cD^{ta}$ for inference. Given the expensive cost of annotating graph data~\cite{yin2023coco}, graph transfer learning represents a crucial topic in graph representation learning.

\begin{figure}[t]
\centering
\includegraphics[width=0.55\linewidth]{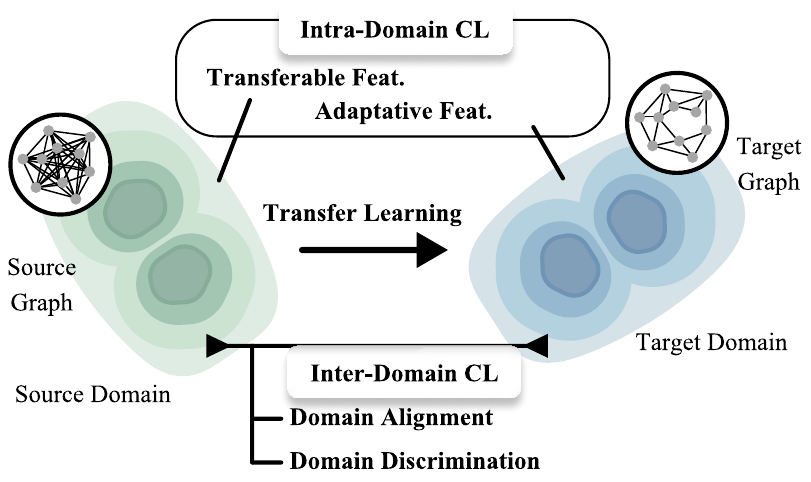}
\caption{GCL in graph transfer learning.}
\label{fig:gcl-transfer}
\end{figure}

Contrastive learning (CL) can improve graph transfer learning from the two aspects: \emph{inter-domain} and \emph{intra-domain}, as shown in Figure~\ref{fig:gcl-transfer}. For inter-domain, CL effectively aligns or discriminates between domains. For intra-domain, CL facilitates extracting transferable graph features that are invariant to domain shifts in the source domain, while allowing the model to learn adaptive features in the target domain. The graph transfer learning methods with GCL are summarized in Table~\ref{tab:transfer}.

\subsubsection{Inter-domain Contrastive Learning}
Through contrasting data from different domains, models can effectively uncover domain relationships and enable domain transformation. The objectives can be categorized as either domain alignment, bringing the domains closer, or discrimination, distinguishing the source and target domains.

\smallskip\textbf{Contrast for Domain Alignment.}
The challenge of using contrastive learning for domain alignment is to identify cross-domain consistent data points as positive samples and contrast them with negative samples.
To generate target-consistent samples from the source domain, DEAL~\cite{yin2022deal} perturbs source graphs using the semantics of the target graph, and then uses contrastive learning to align the source and target domains.
Additionally, some methods attempt to identify anchor samples in the source domain. UDANE~\cite{chen2023universal} introduces prototype contrastive learning, which aligns target domain data with source domain prototypes.
The prototype is computed as:
\begin{equation}
    \mu^{so}_k = \frac{\sum\nolimits_{i\in\mathcal{D}^{so}}{\mathds{1}_{\left\{\arg\max(y^{so}_i)=k\right\}}}\mathbf{h}^{so}_i}{\sum\nolimits_{i\in\mathcal{D}^{so}}{\mathds{1}_{\left\{\arg\max(y^{so}_i)=k\right\}}}} \,,
\end{equation}
where $\mathds{1}_{\left\{\cdot\right\}}$ is the indicator function, $\mathbf{h}^{so}_i$ is the node-level feature, and $\mu^{so}_k$ is the extracted source prototype of class $k$.
Conversely, researchers attempt to obtain source-consistent samples in the target domain. CoCo~\cite{yin2023coco} first generates pseudo-labels $\hat{y}^{ta}_j$ in the target domain. Then, it performs contrastive domain alignment using samples with the same labels across domains as positive pairs:
\begin{equation}
    \Pi(j) = \{i\lvert y^{so}_i = \hat{y}^{ta}_j,\, \mathcal{G}^{so}_i\in \mathcal{D}^{so}\}\,.
\end{equation}
Some researchers also make use of node- and graph-level representations. GCLN~\cite{wu2022attraction} performs contrastive learning between node-level representations and graph-level representations from the source and target domains, aligning cross-domain representations to achieve domain attraction.

\smallskip\textbf{Contrast for Domain Discrimination.}
To improve model robustness and security, learning to discriminate between source and target domain data, and identifying OOD graph data samples, is also necessary~\cite{liu2023good}. These methods focus on finding reliable normal and abnormal samples.
ACT~\cite{wang2023cross} generates normal and abnormal samples using neighborhood similarity information.
Utilizing average data information, GDN~\cite{ding2021few} identifies an abnormal set by contrasting it with the average sampling of a normal set. Through contrastive learning between normal and abnormal samples, these methods enable domain discrimination and anomaly awareness in the target domain.

\begin{table}[t]
\caption{Summary of graph transfer learning methods for GCL. ``AL" and ``DIS" are for inter-domain alignment and discrimination. Meanwhile, ``TR" and ``DA" are for intra-domain transferable representation learning and domain adaptation.}
\label{tab:transfer}
\centering
\resizebox{0.9\columnwidth}{!}{
\begin{tabular}{lcc}
\toprule
Method & Category & Core Idea \\
\midrule
DEAL~\cite{yin2022deal} & AL & Adversarial perturbations for domain alignment \\
\multirow{2}{*}{UDANE~\cite{chen2023universal}} & AL & Prototype contrastive learning \\
& DA & Consistent learning across backbones \\
\multirow{2}{*}{CoCo~\cite{yin2023coco}}  & AL & Pseudo-label learning \\
& DA & Consistent learning of topologies at different levels \\
\multirow{2}{*}{GCLN~\cite{wu2022attraction}}  & AL & Node-and-graph contrastive learning \\
& DA & Local-and-global consistency \\
ACT~\cite{wang2023cross} & DIS & Contrastive learning by neighborhood information \\
GDN~\cite{ding2021few} & DIS & Detection of abnormal set \\
GCC~\cite{qiu2020gcc} & TR & Graph sampling for augmentation \\
SOGA~\cite{zhu2023graphcontrol} & TR & Zero-mlps for target information learning \\
EGI~\cite{zhu2021transfer} & TR & Contrastive learning across k-hop ego-graphs \\
CICG~\cite{chen2022learning} & TR & Invariant subgraphs detection with information bottleneck \\
GraphCL~\cite{you2020graph} & DA & Maximize mutual information with augmented views \\
ALEX~\cite{yuan2023alex} & DA & Noise-robust contrastive learning by SVD \\
\bottomrule
\end{tabular}
}
\end{table}

\subsubsection{Intra-domain Contrastive Learning}
Data from different domains can play different roles in contrastive graph transfer learning. In the source domain, models can learn to extract domain-invariant features, thereby improving transferability. In the target domain, models can adapt to the target distribution, directly improving their performance in the ultimate inference scenario.

\smallskip\textbf{Contrast for Transferable Representation.} 
These methods generally operate on the source domain, heuristically extracting transferable graph patterns (e.g., subgraph structures) and using contrastive learning to improve the transferability of graph representations. 
Specifically, GCC~\cite{qiu2020gcc} and SOGA~\cite{zhu2023graphcontrol} extract subgraph instances and perform contrastive learning on them via InfoNCE, which enables the learning for transferable patterns across domains. 
Similarly, EGI~\cite{zhu2021transfer} captures substructural information via k-hop ego-graphs and performs contrastive learning by optimizing the JSD between the ego-graph embeddings.
Moreover, CICG~\cite{chen2022learning} extracts maximally domain-invariant subgraphs that are robust to domain shifts, using them as positive examples for contrastive learning. CICG discusses the problem from the information bottleneck principle~\cite{ahuja2021invariance} and models with a contrastive learning objective:
\begin{equation}
I\left(\hat{\cG};\widetilde{\cG}\lvert Y\right) \\ 
\approx \, \mathbb{E}_{\substack{
\left\{ \hat{\cG},\widetilde{\cG} \right\} \sim \mathbb{P}\left(\cG\lvert \cY=Y\right),
\left\{ \cG \right\}^{M}_{i=1} \sim \mathbb{P}\left(\cG\lvert \cY \neq Y\right)
}}
\log
\frac{
e^{\phi\left( \mathbf{s}_{\hat{\cG}}, \mathbf{s}_{\widetilde{\cG}} \right)}
}{
e^{\phi\left( \mathbf{s}_{\hat{\cG}}, \mathbf{s}_{\widetilde{\cG}} \right)} +
{\sum\nolimits_{i=1}^{M}} e^{\phi\left( \mathbf{s}_{\hat{\cG}}, \mathbf{s}_{\cG_i} \right)}
}
\,,
\end{equation}
where, $\hat{\cG}$ denotes the extracted domain-invariant subgraphs and $\widetilde{\cG}$ is the sampled source graphs with same label $Y$. $\left\{ \hat{\cG},\widetilde{\cG} \right\} \sim \mathbb{P}\left(\cG\lvert \cY=Y\right)$ is positive set and $\left\{ \cG \right\} \sim \mathbb{P}\left(\cG\lvert \cY \neq Y\right)$ is negative set.

\smallskip\textbf{Contrast for Domain Adaptation.}
Target domain data provides a fertile ground for model adaptation. These methods contrast different graph augmentations, views, or backbones for contrastive learning.
For contrasting graph augmentations, GraphCL~\cite{you2020graph} and ALEX~\cite{yuan2023alex} explore target data through contrastive learning between original and augmented graphs. GTrans~\cite{jin2023empowering} employs test-time DropEdge augmentation to generate positive samples. The learning objective is:
\begin{equation}
\cL = \sum_{j=1}^{M} \left( 1- \frac{\hat{\mathbf{h}}^{\top}_j \mathbf{h}_j}{\lVert\hat{\mathbf{h}}^{\top}_j\rVert\lVert\mathbf{h}_j\rVert }\right) - 
\sum_{j=1}^{M} \left( 1- \frac{\widetilde{\mathbf{h}}^{\top}_j \mathbf{h}_j}{\lVert\widetilde{\mathbf{h}}^{\top}_j\rVert\lVert\mathbf{h}_j\rVert }\right) \,,
\end{equation}
where $\hat{\mathbf{h}}$ denotes the positive samples generated by augmentation and $\widetilde{\mathbf{h}}$ denotes the negative samples generated from feature shuffling. 
For contrasting different graph views, GCLN~\cite{wu2022attraction} uses contrastive learning with local and global consistency to improve feature representations in the targeted domain.
In addition, some methods contrast different graph backbones to fully exploit information from the target domain. CoCo~\cite{yin2023coco} contrasts GCNs and Graph Kernel Networks, while UDANE~\cite{chen2023universal} contrasts GCNs and MLPs. The different inductive biases of these encoders allow for learning robust features of the target domain through their contrasting representations.

\subsection{Others}

In addition to the aforementioned aspects, GCL also plays a broad and significant role in many other areas of data-efficient graph learning, such as graph noise processing, imbalance learning, out-of-distribution (OOD) issue, adversarial attack, and fairness. Up to this point, there have been ample researches that demonstrate the effectiveness of GCL in these task domains. This section will systematically introduce the main ideas of the GCL-based methods for each task.

\smallskip\textbf{GCL against noise.} Due to issues such as manual labeling or experiment cost, noise is commonly present in graphs, including label noise or structure noise. Therefore, researchers explore GCL-based methods to resist the negative impact of noise, enhancing model robustness. The fundamental paradigm of this task involves learning a model with strong structural learning or category prediction capabilities based on a noisy graph or graph with noisy labels. 

To address the noise issue, GCL is primarily incorporated into model in the form of regularization. CR-GNN~\cite{li2024contrastive} employs a GCL framework that integrates neighbor information to construct the contrastive loss, while also introducing cross-space consistency to reduce the semantic disparity between the contrastive space and classification space. Such a strategy helps mitigate overfitting to noisy labels by leveraging structural information. 
CGNN~\cite{yuan2023learning} incorporates GCL as a form of regularization, which refrains from relying on label information to prevent overfitting to noisy labels, thereby enhancing its robustness and generalization capability. 
OMG~\cite{yin2023omg} combines coupled Mixup with GCL to tackle the problem of noisy labels. Generating positive and negative pairs using convex combination and multiple sample Mixup enhances the model's generalization capability.

\smallskip\textbf{GCL against imbalance.} The imbalance issue in graphs, caused by uneven class distributions, challenges traditional graph learning algorithms, which tend to favor majority classes over minority ones. This bias results in unreliable predictions for underrepresented classes. Researchers seek GCL-based methods to address this issue~\cite{yi2023towards,mao2023rahnet}, aiming to design models that can effectively learn from imbalanced graph data, ensuring accurate and balanced predictions across all classes. 

To handle the imbalance class distribution, there are many promising algorithms proposed for these issues, ImGCL~\cite{zeng2023imgcl} builds a principled GCL framework on imbalanced node classification to automatically and adaptively balance the node representations across all classes with the support of progressively balanced sampling. By leveraging graph and text information, CM-GCL~\cite{qian2022co} proposes inter-modality GCL and intra-modality GCL to automatically generate contrastive pairs and achieve balanced representation across unlabeled data in a co-training paradigm.

\smallskip\textbf{GCL against out-of-distribution (OOD).} The OOD problem in graphs arises when the data distribution encountered during testing differs significantly from that during training. 
Researchers study GCL-based methods to address this issue, aiming to develop models that can detect OOD instances in graphs or provide reliable predictions to the target domain, corresponding to OOD detection and OOD generalization.

In the OOD detection task, with the assistance of GCL, GRADATE~\cite{duan2023graph} proposes a multi-view multi-scale GCL framework that contains node-subgraph, node-node, and subgraph-subgraph contrasts to enhance the robustness of the node embeddings for graph anomaly detection. 
GOOD-D~\cite{liu2023good} employs hierarchical contrastive learning to augment graphs produced via a perturbation-free graph augmentation approach. Its goal is to capture underlying in-distribution (ID) patterns and detect OOD graphs by pinpointing semantic inconsistencies across different levels of granularity, including node-level, graph-level, and group-level distinctions. 
GLADC~\cite{luo2022deep} utilizes the GCL strategy to improve the representations of nodes and graphs in normal graphs using an encoder-decoder architecture. It then assesses the error between these representations and those of the generated reconstruction graph to identify anomalous graphs.
In addition, in the OOD generalization task, 
FLOOD~\cite{liu2023flood} leverages invariant learning and bootstrapped learning with a GCL architecture to acquire invariant representations conducive to risk extrapolation.
CIGA~\cite{chen2022learning} captures the invariance of graphs under various distribution shifts by using GCL to maximize the agreement between the invariant part of graphs with the same label.

\smallskip\textbf{Adversarial attack for GCL.} The field of adversarial attack in graphs involves developing methods to defend against malicious manipulations aimed at disrupting models trained on graphs. In the realm of GCL, the study of adversarial attack serves a crucial purpose. By integrating adversarial attack learning techniques into the GCL framework, researchers aim to enhance the models' robustness against malicious manipulations of graph structures or attributes. 

To build informative contrastive samples and improve the robustness of contrastive learning, many papers have recently considered introducing adversarial attacks into the GCL framework. 
ArieL~\cite{feng2022adversarial} introduces an adversarial graph perspective as a novel data augmentation technique and an information regularizer, aiming at offering high-quality contrastive samples and enhancing the stability of GCL.
CLGA~\cite{zhang2022unsupervised} suggests an unsupervised gradient-based adversarial attack for GCL through flipping the edges with gradient ascent to maximize the contrastive loss. This approach selects the most informative edges, enabling the model to adapt better to various downstream tasks.
GCBA~\cite{zhang2023graph} pays attention to the security of GCL under malicious backdoor adversaries and designs the first backdoor attack including poisoning, crafting, and natural backdoor, which illustrates that specifically designed defenses are needed for further study. 
SP-AGCL~\cite{in2023similarity} proposes a similarity-preserving adversarial GCL that contrasts the clean graph with the node similarity-preserving view and the adversarial view. 
RES~\cite{lin2023certifiably} underscores the certifiable robustness of GCL by introducing a unified criterion for evaluating and certifying its robustness. Additionally, it presents a novel technique to ensure this certifiable robustness, which can be reliably maintained in downstream tasks.

\smallskip\textbf{Fairness for GCL.} Ensuring fairness entails mitigating biases and disparities that may arise in the learning process due to various factors such as degree bias or sampling bias in graphs. By incorporating fairness-aware techniques into GCL, researchers aim to develop models that exhibit equitable behavior. This approach fosters the creation of models capable of making fair predictions and decisions to enhance their applicability in diverse graph-based applications.

Wang et al.~\cite{wang2022uncovering} explore structural fairness and demonstrates theoretically that GCL leads to a clearer community structure, driving low-degree nodes further away from community boundaries. Building upon this theoretical foundation, the paper introduces GRADE, which employs distinct graph augmentation strategies for low- and high-degree nodes.
Kose and Shen~\cite{kose2022fair} introduce innovative fairness-aware graph augmentations using adaptive feature masking and edge deletion to mitigate bias in graph contrastive learning. This study introduces various fairness notions in graphs to inform the design of the developed adaptive augmentation strategies.
To address diverse application scenarios, Graphair~\cite{ling2022learning} devises automated graph data augmentations under the GCL framework. These augmentations aim to avoid sensitive information while retaining other valuable insights, which are themselves derived directly from the input graph.

\section{Real-world Applications}
In this section, we discuss the applications of GCL in various real-world scenarios, encompassing valuable fields such as drug discovery, genomics analysis, recommender systems, social networks, and traffic forecasting, demonstrating its practical utility.

\subsection{Drug Discovery}

Drug discovery~\cite{muttenthaler2021trends,gupta2021artificial} aims to discover medications that require precise drug screening for given targets. Graph representation learning has become increasingly crucial in drug discovery, which provides a generic framework to model complex molecular structures. Therefore, GCL is widely utilized in molecular science with different applications including biomolecular interaction analysis~\cite{rohani2019drug} and molecular property prediction~\cite{li2022geomgcl}. 

\smallskip\textbf{Biomolecular Interaction Analysis.}
As a fundamental problem in drug discovery, biomolecular interaction analysis includes the study of drug-drug interaction, drug-gene interaction and drug-target interaction. Recent machine learning approaches~\cite{wang2024predicting,yao2023semi,tao2023prediction,gao2023similarity} usually utilize graphs to represent both drug molecules and receptor proteins, followed by graph neural networks for downstream prediction tasks. They usually generate graph representations from different views, (e.g., local and global views) and then GCL has been adopted to enhance the agreement across different views for information interaction. For example, CSCo-DTA~\cite{wang2024predicting} generates graph representations from two different views, i.e., the interaction network and biological entities and maximizes the mutual information across these views using GCL. It also adopts select positive views using drug–drug similarities and meta-paths, which enables effective contrastive learning for drug–target binding affinity predictions. SHGCL-DTI~\cite{yao2023semi} introduces GCL for heterogeneous graphs, which involves neighboring and meta-path views for cross-view representation learning. DGCL~\cite{tao2023prediction} constructs an interaction graph and a dynamic hypergraph to provide graph representations from a local and a global. Then a GCL objective is adopted to bridge the complimentary semantics. SMGCL~\cite{gao2023similarity} introduces three different views, i.e., drug similarity, disease-similarity and their associations to effective model local and global topologies, enhanced with GCL for effective drug–disease association prediction. 

\smallskip\textbf{Molecular Property Prediction.} Molecular property prediction is an essential problem in drug discovery and material discovery. As a basic application of graph-level classification, recent approaches leverage GCL to enhance representations of molecules. They~\cite{li2022geomgcl,zheng2023casangcl,zang2023hierarchical,wang2023molecular,ju2023few} usually explore different property augmentation strategies for effective semantic exploration in molecular domains. For example, GeomMPNN~\cite{li2022geomgcl} introduces both 2D graphs and 3D graphs to represent molecules and then learns topological information for graph representations from different views. It then maximizes the mutual information across 2D and 3D views for representation enhancement. CasANGCL~\cite{zheng2023casangcl} first introduces multi-granularity graph perturbation strategies including attribute masking and subgraph extraction to generate different augmented views for GCL. It also introduces supervised contrastive learning by constructing positive pairs using labels. HiMol~\cite{zang2023hierarchical} involves motif structures in the hierarchical molecular representation learning process, which is followed by a self-supervised learning task. iMolCLR~\cite{wang2022improving} decomposes molecules into different substructures and introduces contrastive learning to learn representations of these substructures. These approaches can be also extended to protein function prediction and solubility prediction. For example, HEAL~\cite{gu2023hierarchical} introduces perturbation to hidden space instead of graph samples for regularization, which further enhances protein representations in a self-supervised manner.

\subsection{Genomics Analysis}

Genomics analysis~\cite{erfanian2023deep} is critical for understanding disease mechanisms, evolutionary biology, and benefiting medicine development, which has received extensive attention in the field of machine learning~\cite{lee2020single}. GCL also has extensive applications in different genomics applications including single-cell data imputation~\cite{hou2020systematic}, clustering~\cite{liu2021simultaneous} and multi-omics data integration~\cite{lee2020single}.

\smallskip\textbf{Single-cell Data Imputation and Clustering.} Single-cell RNA-sequencing (scRNA-seq) provides a convenient tool for understanding cell behavior at the single-cell level. Data imputation and clustering are fundamental tasks in single-cell data analysis. Existing GCL approaches~\cite{xiong2023scgcl,lee2023deep,tian2023scgcc,zheng2024subgraph} usually build graphs based on similarity or interaction information and then conduct GCL for discriminative cell representations. For example, scGCL~\cite{xiong2023scgcl} builds a cell graph using the k-nearest neighborhood (kNN) and then infers both local and global information for accurate positive pairs, which would be incorporated into a GCL framework for effective cell representation. An autoencoder framework is adopted to reconstruct the gene expression matrix.  scGPCL~\cite{lee2023deep} constructs a cell-gene graph for representation learning, and then generates augmented views for instance-wise contrastive learning. Prototypical contrastive learning is also adopted to enforce cell representations to approach their corresponding prototypes to enhance clustering results.

\smallskip\textbf{Multi-omics Data Integration.} Recently, multi-omics techniques~\cite{lee2020single} can provide different perspectives for researchers to sufficiently understand complicated biological systems. The complicated multi-omics data can be efficiently modeled by graphs such as co-expression networks. Then, existing approaches~\cite{liu2024muse,rajadhyaksha2023graph,xie2023mtgcl,zong2022const} usually adopt node-level GCL to enhance cell and gene representations for downstream tasks. 
For example, MuSe-GNN~\cite{liu2024muse} adopts a graph Transformer to learn gene representations from co-expression networks and then adopts GCL to maximize the mutual information between similar genes. MOGCL~\cite{rajadhyaksha2023graph} constructs graphs using a pairwise distance matrix, followed by the framework in GCA to learn effective cell representations. MTGCL~\cite{xie2023mtgcl} adopts a multi-task learning framework, which combines semi-supervised node classification with GCL. These effective multi-omics representations are used to determine cancer driver genes for precision medicine. ConST~\cite{zong2022const} further utilizes GCL to learn from spatially resolved transcriptomics (SRT). It contains contrastive learning in three different levels: (1) local-local contrast utilizes attribute masking for augmentation to emphasize important features; (2) local-global contrast maximizes the mutual information between node embeddings and graph embeddings to learn global property; (3) global-context contrast follows the paradigm of prototypical contrastive learning to enhance the clustering performance.

\subsection{Recommender Systems}

Graph-based recommender systems \cite{he2020lightgcn,wu2022graph} serve as novel solutions to modern web platforms to alleviate the information-overload issue and provide personalized recommendation results for users. Learning informative and high-quality node representations becomes the key to promising recommender models. GCL techniques assist the model to fully exploit neighborhood structural information for collaborative filtering \cite{wang2019neural} and other downstream recommendation tasks \cite{wu2019session,fan2019graph}.

\smallskip\textbf{Graph-based Collaborative Filtering.} As a key part of industrial recommendation pipelines, collaborative filtering (CF) models aim at matching users with sets of items based on their interaction history. Since graph-based methods have achieved promising results in CF tasks \cite{shen2021powerful,mao2021ultragcn,guo2023manipulating,qin2024polycf}, later researches have turned their attention towards GCL methods for better performance. As a pioneering work of adopting GCL in CF models, SGL \cite{wu2021self} proposes to generate multiple views of the user-item bipartite graph with node-dropping, edge-dropping, and random-walk. 
Liu et al. \cite{liu2021contrastive} adopt graph perturbation and conduct contrastive learning for debiased recommendation. Inspired by the idea of graph view generation, LightGCL \cite{cai2023lightgcl} proposes an SVD-based contrastive learning scheme, where the global structure of the interaction graph is obtained by a low-rank factorization. HCCF \cite{xia2022hypergraph} extends this idea and leverages a hypergraph structure learning framework as the contrastive view for graph recommender models. Compared with the aforementioned methods that sought augmented graph views for contrastive learning, Yu et al. doubt the necessity of graph augmentations and propose SimGCL \cite{yu2022graph}, a contrastive learning model with random noise-based data augmentation. XSimGCL \cite{yu2023xsimgcl} merges the encoding processes of SimGCL and constructs an extremely simple GCL. DCCF \cite{ren2023disentangled} combines multiple contrastive representations in a disentangled style to exclude the augmentation-induced noise inherent in the graph views. SimpleX \cite{mao2021simplex} constructs the cosine contrastive loss to learn uniform and informative item representations. Apart from traditional user-item interaction graphs, there are efforts on constructing novel graph structures to exploit the collaborative similarity. QRec \cite{yu2021self} turns to hypergraph recommendation with a self-supervised multi-channel contrastive learning and makes personalized recommendations based on social influence. HeCo \cite{wang2021self} proposes a novel co-contrastive learning mechanism for heterogeneous graphs between views of meta-paths. HGCL \cite{chen2023heterogeneous} involves contrastive pairs from heterogeneous graph nodes to achieve knowledge transfer between views. KGCL \cite{yang2022knowledge} introduces knowledge graph-based contrastive learning while suppressing noise within the interaction graphs.

\smallskip\textbf{GNN for Downstream Recommendation.} Apart from collaborative filtering, there are a variety of recommendation tasks aiming at addressing real-world challenges, such as sequential recommendation \cite{qin2024learning} and point-of-interests (POI) recommendation \cite{ju2022kernel,qin2023diffusion}. These specific downstream recommendation tasks require tailor-designed graph learning and contrastive learning schemes. For instance, COTREC \cite{xia2021self} focuses on the session-based recommendation. While item/segment dropout fails to augment session data since it may lead to sparser data, COTREC constructs a global session adjacency to provide and optimize the session representation with contrastive learning. DCRec \cite{yang2023debiased} proposes to augment users' sequential patterns with a global collaborative relationship between items and applies a debiased contrastive learning framework to capture the rich behavior patterns. There are also attempts to leverage contrastive learning to exploit information from multiple data modalities. SDCRec \cite{du2022socially} proposes to contrastively learn better representations for cold-start users with the help of their social connections. MMGCL \cite{yi2022multimodal} uses the modality edge dropout and modality masking techniques to align representations across modalities in the micro-video recommendation. DisenPOI \cite{qin2023disenpoi} tackles the challenges in point-of-interest (POI) recommendation, where the geographical representations are entangled with sequential representations of users. By introducing a contrastive learning-based disentangled loss, DisenPOI learns the separate influence of multiple factors in POI recommendation. RCENR \cite{jiang2023rcenr} generates multiple user/news sub-graphs to enhance news recommendation and incorporates a reinforcement-based strategy for contrastive learning.

\subsection{Social Networks}

Social network analysis \cite{freeman2004development,tabassum2018social} focuses on the behavior of social actors in various types of social networks. Given the widely existing graph structure in social networks, contrastive graph representation learning is widely applied for both social network representation and different network detection tasks.

While there is widespread fraud and fake information in online social media, an important task is to detect possible false information and relative users. RDCL \cite{ma2022towards} adapts contrastive learning objectives to ensure the consistency between perturbed and original social networks. By encouraging the model to maintain resistance to structural perturbations, the RDCL model is able to detect false information in a robust way. CBD \cite{zhou2023detecting} generates augmented views by removing edges between users and proposes a contrastive learning-based pre-train and fine-tune framework for detecting social bots on the fly. CALN \cite{ma-etal-2022-open-topic} constructs contrastive pairs between different topics to generalize the model to unseen open-topic fields, thus improving its ability to detect false information. GACL \cite{sun2022rumor} extends the contrastive generalization by modeling the rumor propagation process and depicting the differences between conversational threads on social networks, achieving promising performance in the rumor detection task.

\subsection{Traffic Forecasting}
Predicting the traffic situation of public transportation systems \cite{li2024survey,zhao2023dynamic} is a popular research field in the application of GNNs. GCL methods are able to fully utilize the spatio-temporal feature of traffic data and provide robust and accurate predictions.

To model the spatio-temporal dynamics of traffic data, GraphST \cite{zhang2023spatial} proposes an adversarial contrastive learning paradigm to aggregate information from multiple views of POI graphs. The collected data from sensors often suffer from noise and incompleteness. To address this issue, UrbanGCL \cite{pan2023spatial} proposes a feature-level and a topology-level data augmentation method to improve model robustness with the spatial and temporal contrastive learning auxiliary tasks. In order to better capture the periodical features in traffic activities, ST-A-PGCL \cite{qu2023st} adaptively conducts contrastive learning between three branches of periodical patterns and achieves promising results. While there are occasional sudden changes that outperform periodicity in real-world traffic flows, GCGAN \cite{xu2023gan} proposes to use a GAN that generates contrastive graph samples to predict sudden changes. To mitigate the data scarcity issue and anomalies, CDAGF \cite{ji2024contrastive} proposes a learnable graph structure to provide the augmented views of location nodes. The generated views are then integrated with a multi-graph fusion convolution layer, optimized by a contrastive fusion objective.
\section{Conclusion and Future Work}
In summary, this survey provides a comprehensive overview of Graph Contrastive Learning (GCL), addressing a critical gap in the literature by elucidating its foundational principles, including augmentation strategies, contrastive modes, and optimization objectives. Additionally, we extend our exploration to cover environments in weakly supervised learning, transfer learning, and other data-efficient scenarios, demonstrating GCL's versatility and impact across various real-world applications such as drug discovery, genomics analysis, recommender systems, social networks, and traffic forecasting. Despite the advancements in GCL, several challenges remain, paving the way for future research to explore and enhance the potential of this promising field.

\smallskip\textbf{More Theoretical Understanding.}
GCL, as an emerging learning paradigm, has not yet built a deep theoretical foundation. Current researches primarily focus on proposing novel training frameworks, which are often intuitively designed and validated through empirical experiments, but lack in-depth understanding of their theoretical properties. In the future, it is necessary to establish a solid theoretical foundation for GCL, around its generalization ability, convergence properties, theoretical bounds of performance, etc., which are beneficial for developing an efficient and reliable framework for graph representation learning. For instance, Yuan et al.~\cite{yuan2024towards} propose a metric to evaluate the generalization capability of GCL and theoretically proved a mutual information upper bound for this metric from an information theory perspective, further guiding the design of a novel GCL framework with more powerful generalization ability.

\smallskip\textbf{More Effective Augmentation Strategies.}
Augmentation strategies play a crucial role in contrastive learning to generate distinguishable and meaningful representations~\cite{zhang2022rethinking}. However, compared to image and text data, graph-structured data inherently possess complex non-Euclidean characteristics, making the design of augmentation strategies challenging and underexplored. It is a promising direction to develop more targeted and efficient augmentation strategies to enhance the performance of GCL. Specifically, further exploration could be conducted on augmentation strategies based on graph topology characteristics, spectral graph theory~\cite{ghose2023spectral}, automatable and adaptive techniques~\cite{zhu2021graph}, dynamically updating approaches, computationally efficient considerations, and augmentation strategies tailored for specific complex graph types such as heterogeneous graphs~\cite{chen2023heterogeneous} and spatio-temporal graphs~\cite{zhang2023automated}.

\smallskip\textbf{More Interpretability.}
Previous works on the interpretability of graph learning models mostly focus on supervised learning settings~\cite{yuan2021explainability,zhang2022protgnn}, while there is still a lack of research on the explainability of graph self-supervised learning. Understanding the intrinsic patterns and structural semantics of the learned representations and potential explanations for predictions made by GCL is crucial for developing more reliable and secure GCL frameworks, enabling their applications to vital fields such as the pharmaceutical industry~\cite{sun2021mocl}. For instance, RGCL~\cite{li2022let} inspired by the concept of invariant rationale discovery (IRD), introduce a rationale generator to automatically learn important subgraphs within molecular graphs for molecular discrimination. It integrates the sufficiency and independence principle of IRD into the GCL framework, enabling the discovery and exploitation of rationales related to molecular properties during the optimization process of GCL, thereby enhancing the framework's interpretability. Exploring explainable GCL from more perspectives is an interesting and practical direction.


\smallskip\textbf{More Integration of Domain Knowledge.}
The application domains of GCL are extensive, including pharmaceutical industry~\cite{sun2021mocl}, bioinformatics~\cite{gu2023hierarchical}, recommender systems~\cite{yu2023xsimgcl}, financial industry~\cite{wang2024assessing}, and traffic prediction~\cite{liu2022contrastive}, etc. These domains often possess their prior knowledge and characteristics, such as chemical bonds and functional groups in molecules, and traffic rules and preferences in transportation networks. Leveraging such domain knowledge and rules to design domain-oriented augmentation strategies, pretext tasks, negative sample generation strategies, or training frameworks significantly aids in enhancing the performance of GCL in specific domains. For instance, KANO~\cite{fang2023knowledge} utilize a chemical element-oriented knowledge graph (ElementKG) to augment molecular graphs by supplementing inter-element relations from ElementKG into the original graphs, thus establishing fundamental connections beyond structural information between atoms and providing richer chemical semantics for molecular representations.

\smallskip\textbf{Improving Robustness.}
Real-world data often comes with various noise and incompleteness, and graph neural networks are susceptible to adversarial attacks~\cite{zugner2018adversarial}, posing significant challenges to the robustness of GCL models. While most current efforts focus on improving GCL model performance in various downstream tasks, research on the robustness of GCL models against structural perturbations, label noise, adversarial attacks, domain shifts, and other interferences remains inadequate. Systematically exploring and developing more robust GCL methods to address various real-world noises or attacks is an important research direction. For instance, Lin et al.~\cite{lin2023certifiably} propose a unified definition of robustness in GCL and introduce the randomized edgedrop smoothing (RES) technique, embedding randomized edgedrop noise into graphs to equip GCL with certified robustness in downstream tasks. By delving deeper into the robustness of GCL, GCL techniques can be better applied to risk-sensitive critical scenarios like financial fraud detection~\cite{wang2019semi}.

\begin{acks}

This paper is partially supported by the National Natural Science Foundation of China (NSFC Grant Numbers 62306014 and 62276002).
\end{acks}

\bibliographystyle{ACM-Reference-Format}
\bibliography{sample-base}


\begin{thebibliography}{235}


\ifx \showCODEN    \undefined \def \showCODEN     #1{\unskip}     \fi
\ifx \showDOI      \undefined \def \showDOI       #1{#1}\fi
\ifx \showISBNx    \undefined \def \showISBNx     #1{\unskip}     \fi
\ifx \showISBNxiii \undefined \def \showISBNxiii  #1{\unskip}     \fi
\ifx \showISSN     \undefined \def \showISSN      #1{\unskip}     \fi
\ifx \showLCCN     \undefined \def \showLCCN      #1{\unskip}     \fi
\ifx \shownote     \undefined \def \shownote      #1{#1}          \fi
\ifx \showarticletitle \undefined \def \showarticletitle #1{#1}   \fi
\ifx \showURL      \undefined \def \showURL       {\relax}        \fi
\providecommand\bibfield[2]{#2}
\providecommand\bibinfo[2]{#2}
\providecommand\natexlab[1]{#1}
\providecommand\showeprint[2][]{arXiv:#2}

\bibitem[Agarwal(2006)]%
        {agarwal2006ranking}
\bibfield{author}{\bibinfo{person}{Shivani Agarwal}.} \bibinfo{year}{2006}\natexlab{}.
\newblock \showarticletitle{Ranking on graph data}. In \bibinfo{booktitle}{\emph{ICML}}. \bibinfo{pages}{25--32}.
\newblock


\bibitem[Ahuja et~al\mbox{.}(2021)]%
        {ahuja2021invariance}
\bibfield{author}{\bibinfo{person}{Kartik Ahuja}, \bibinfo{person}{Ethan Caballero}, \bibinfo{person}{Dinghuai Zhang}, \bibinfo{person}{Jean-Christophe Gagnon-Audet}, \bibinfo{person}{Yoshua Bengio}, \bibinfo{person}{Ioannis Mitliagkas}, {and} \bibinfo{person}{Irina Rish}.} \bibinfo{year}{2021}\natexlab{}.
\newblock \showarticletitle{Invariance principle meets information bottleneck for out-of-distribution generalization}.
\newblock \bibinfo{journal}{\emph{NeurIPS}}  \bibinfo{volume}{34} (\bibinfo{year}{2021}), \bibinfo{pages}{3438--3450}.
\newblock


\bibitem[Alemany et~al\mbox{.}(2022)]%
        {alemany2022review}
\bibfield{author}{\bibinfo{person}{Jos{\'e} Alemany}, \bibinfo{person}{E~Del Val}, {and} \bibinfo{person}{Ana Garc{\'\i}a-Fornes}.} \bibinfo{year}{2022}\natexlab{}.
\newblock \showarticletitle{A review of privacy decision-making mechanisms in online social networks}.
\newblock \bibinfo{journal}{\emph{CSUR}} \bibinfo{volume}{55}, \bibinfo{number}{2} (\bibinfo{year}{2022}), \bibinfo{pages}{1--32}.
\newblock


\bibitem[Bachman et~al\mbox{.}(2019)]%
        {bachman2019learning}
\bibfield{author}{\bibinfo{person}{Philip Bachman}, \bibinfo{person}{R~Devon Hjelm}, {and} \bibinfo{person}{William Buchwalter}.} \bibinfo{year}{2019}\natexlab{}.
\newblock \showarticletitle{Learning Representations by Maximizing Mutual Information Across Views}.
\newblock \bibinfo{journal}{\emph{NeurIPS}}  \bibinfo{volume}{32} (\bibinfo{year}{2019}), \bibinfo{pages}{15509--15519}.
\newblock


\bibitem[Baek et~al\mbox{.}(2020)]%
        {baek2020accurate}
\bibfield{author}{\bibinfo{person}{Jinheon Baek}, \bibinfo{person}{Minki Kang}, {and} \bibinfo{person}{Sung~Ju Hwang}.} \bibinfo{year}{2020}\natexlab{}.
\newblock \showarticletitle{Accurate Learning of Graph Representations with Graph Multiset Pooling}. In \bibinfo{booktitle}{\emph{ICLR}}.
\newblock


\bibitem[Bardes et~al\mbox{.}(2021)]%
        {bardes2021vicreg}
\bibfield{author}{\bibinfo{person}{Adrien Bardes}, \bibinfo{person}{Jean Ponce}, {and} \bibinfo{person}{Yann LeCun}.} \bibinfo{year}{2021}\natexlab{}.
\newblock \showarticletitle{Vicreg: Variance-invariance-covariance regularization for self-supervised learning}.
\newblock \bibinfo{journal}{\emph{arXiv preprint arXiv:2105.04906}} (\bibinfo{year}{2021}).
\newblock


\bibitem[Bielak et~al\mbox{.}(2022)]%
        {bielak2022graph}
\bibfield{author}{\bibinfo{person}{Piotr Bielak}, \bibinfo{person}{Tomasz Kajdanowicz}, {and} \bibinfo{person}{Nitesh~V Chawla}.} \bibinfo{year}{2022}\natexlab{}.
\newblock \showarticletitle{Graph barlow twins: A self-supervised representation learning framework for graphs}.
\newblock \bibinfo{journal}{\emph{Knowledge-Based Systems}}  \bibinfo{volume}{256} (\bibinfo{year}{2022}), \bibinfo{pages}{109631}.
\newblock


\bibitem[Cai et~al\mbox{.}(2023)]%
        {cai2023lightgcl}
\bibfield{author}{\bibinfo{person}{Xuheng Cai}, \bibinfo{person}{Chao Huang}, \bibinfo{person}{Lianghao Xia}, {and} \bibinfo{person}{Xubin Ren}.} \bibinfo{year}{2023}\natexlab{}.
\newblock \showarticletitle{Light{GCL}: Simple Yet Effective Graph Contrastive Learning for Recommendation}. In \bibinfo{booktitle}{\emph{ICLR}}.
\newblock


\bibitem[Cao et~al\mbox{.}(2021)]%
        {cao2021bipartite}
\bibfield{author}{\bibinfo{person}{Jiangxia Cao}, \bibinfo{person}{Xixun Lin}, \bibinfo{person}{Shu Guo}, \bibinfo{person}{Luchen Liu}, \bibinfo{person}{Tingwen Liu}, {and} \bibinfo{person}{Bin Wang}.} \bibinfo{year}{2021}\natexlab{}.
\newblock \showarticletitle{Bipartite graph embedding via mutual information maximization}. In \bibinfo{booktitle}{\emph{WSDM}}. \bibinfo{pages}{635--643}.
\newblock


\bibitem[Caron et~al\mbox{.}(2020)]%
        {caron2020unsupervised}
\bibfield{author}{\bibinfo{person}{Mathilde Caron}, \bibinfo{person}{Ishan Misra}, \bibinfo{person}{Julien Mairal}, \bibinfo{person}{Priya Goyal}, \bibinfo{person}{Piotr Bojanowski}, {and} \bibinfo{person}{Armand Joulin}.} \bibinfo{year}{2020}\natexlab{}.
\newblock \showarticletitle{Unsupervised learning of visual features by contrasting cluster assignments}.
\newblock \bibinfo{journal}{\emph{NeurIPS}}  \bibinfo{volume}{33} (\bibinfo{year}{2020}), \bibinfo{pages}{9912--9924}.
\newblock


\bibitem[Chen et~al\mbox{.}(2023a)]%
        {chen2023universal}
\bibfield{author}{\bibinfo{person}{Jushuo Chen}, \bibinfo{person}{Feifei Dai}, \bibinfo{person}{Xiaoyan Gu}, \bibinfo{person}{Jiang Zhou}, \bibinfo{person}{Bo Li}, {and} \bibinfo{person}{Weipinng Wang}.} \bibinfo{year}{2023}\natexlab{a}.
\newblock \showarticletitle{Universal domain adaptive network embedding for node classification}. In \bibinfo{booktitle}{\emph{ACMMM}}. \bibinfo{pages}{4022--4030}.
\newblock


\bibitem[Chen et~al\mbox{.}(2023b)]%
        {chen2023zero}
\bibfield{author}{\bibinfo{person}{Jiaoyan Chen}, \bibinfo{person}{Yuxia Geng}, \bibinfo{person}{Zhuo Chen}, \bibinfo{person}{Jeff~Z Pan}, \bibinfo{person}{Yuan He}, \bibinfo{person}{Wen Zhang}, \bibinfo{person}{Ian Horrocks}, {and} \bibinfo{person}{Huajun Chen}.} \bibinfo{year}{2023}\natexlab{b}.
\newblock \showarticletitle{Zero-shot and few-shot learning with knowledge graphs: A comprehensive survey}.
\newblock \bibinfo{journal}{\emph{Proc. IEEE}} (\bibinfo{year}{2023}).
\newblock


\bibitem[Chen and Kou(2023)]%
        {chen2023attribute}
\bibfield{author}{\bibinfo{person}{Jialu Chen} {and} \bibinfo{person}{Gang Kou}.} \bibinfo{year}{2023}\natexlab{}.
\newblock \showarticletitle{Attribute and structure preserving graph contrastive learning}. In \bibinfo{booktitle}{\emph{AAAI}}, Vol.~\bibinfo{volume}{37}. \bibinfo{pages}{7024--7032}.
\newblock


\bibitem[Chen et~al\mbox{.}(2023c)]%
        {chen2023heterogeneous}
\bibfield{author}{\bibinfo{person}{Mengru Chen}, \bibinfo{person}{Chao Huang}, \bibinfo{person}{Lianghao Xia}, \bibinfo{person}{Wei Wei}, \bibinfo{person}{Yong Xu}, {and} \bibinfo{person}{Ronghua Luo}.} \bibinfo{year}{2023}\natexlab{c}.
\newblock \showarticletitle{Heterogeneous graph contrastive learning for recommendation}. In \bibinfo{booktitle}{\emph{WSDM}}. \bibinfo{pages}{544--552}.
\newblock


\bibitem[Chen et~al\mbox{.}(2020)]%
        {chen2020simple}
\bibfield{author}{\bibinfo{person}{Ting Chen}, \bibinfo{person}{Simon Kornblith}, \bibinfo{person}{Mohammad Norouzi}, {and} \bibinfo{person}{Geoffrey Hinton}.} \bibinfo{year}{2020}\natexlab{}.
\newblock \showarticletitle{A simple framework for contrastive learning of visual representations}. In \bibinfo{booktitle}{\emph{ICML}}. \bibinfo{pages}{1597--1607}.
\newblock


\bibitem[Chen et~al\mbox{.}(2022)]%
        {chen2022learning}
\bibfield{author}{\bibinfo{person}{Yongqiang Chen}, \bibinfo{person}{Yonggang Zhang}, \bibinfo{person}{Yatao Bian}, \bibinfo{person}{Han Yang}, \bibinfo{person}{MA Kaili}, \bibinfo{person}{Binghui Xie}, \bibinfo{person}{Tongliang Liu}, \bibinfo{person}{Bo Han}, {and} \bibinfo{person}{James Cheng}.} \bibinfo{year}{2022}\natexlab{}.
\newblock \showarticletitle{Learning causally invariant representations for out-of-distribution generalization on graphs}.
\newblock \bibinfo{journal}{\emph{NeurIPS}}  \bibinfo{volume}{35} (\bibinfo{year}{2022}), \bibinfo{pages}{22131--22148}.
\newblock


\bibitem[Devlin et~al\mbox{.}(2018)]%
        {devlin2018bert}
\bibfield{author}{\bibinfo{person}{Jacob Devlin}, \bibinfo{person}{Ming-Wei Chang}, \bibinfo{person}{Kenton Lee}, {and} \bibinfo{person}{Kristina Toutanova}.} \bibinfo{year}{2018}\natexlab{}.
\newblock \showarticletitle{Bert: Pre-training of deep bidirectional transformers for language understanding}.
\newblock \bibinfo{journal}{\emph{arXiv preprint arXiv:1810.04805}} (\bibinfo{year}{2018}).
\newblock


\bibitem[Ding et~al\mbox{.}(2021)]%
        {ding2021few}
\bibfield{author}{\bibinfo{person}{Kaize Ding}, \bibinfo{person}{Qinghai Zhou}, \bibinfo{person}{Hanghang Tong}, {and} \bibinfo{person}{Huan Liu}.} \bibinfo{year}{2021}\natexlab{}.
\newblock \showarticletitle{Few-shot network anomaly detection via cross-network meta-learning}. In \bibinfo{booktitle}{\emph{WWW}}. \bibinfo{pages}{2448--2456}.
\newblock


\bibitem[Du et~al\mbox{.}(2022)]%
        {du2022socially}
\bibfield{author}{\bibinfo{person}{Jing Du}, \bibinfo{person}{Zesheng Ye}, \bibinfo{person}{Lina Yao}, \bibinfo{person}{Bin Guo}, {and} \bibinfo{person}{Zhiwen Yu}.} \bibinfo{year}{2022}\natexlab{}.
\newblock \showarticletitle{Socially-aware dual contrastive learning for cold-start recommendation}. In \bibinfo{booktitle}{\emph{SIGIR}}. \bibinfo{pages}{1927--1932}.
\newblock


\bibitem[Duan et~al\mbox{.}(2023)]%
        {duan2023graph}
\bibfield{author}{\bibinfo{person}{Jingcan Duan}, \bibinfo{person}{Siwei Wang}, \bibinfo{person}{Pei Zhang}, \bibinfo{person}{En Zhu}, \bibinfo{person}{Jingtao Hu}, \bibinfo{person}{Hu Jin}, \bibinfo{person}{Yue Liu}, {and} \bibinfo{person}{Zhibin Dong}.} \bibinfo{year}{2023}\natexlab{}.
\newblock \showarticletitle{Graph anomaly detection via multi-scale contrastive learning networks with augmented view}. In \bibinfo{booktitle}{\emph{AAAI}}. \bibinfo{pages}{7459--7467}.
\newblock


\bibitem[Erfanian et~al\mbox{.}(2023)]%
        {erfanian2023deep}
\bibfield{author}{\bibinfo{person}{Nafiseh Erfanian}, \bibinfo{person}{A~Ali Heydari}, \bibinfo{person}{Adib~Miraki Feriz}, \bibinfo{person}{Pablo Ia{\~n}ez}, \bibinfo{person}{Afshin Derakhshani}, \bibinfo{person}{Mohammad Ghasemigol}, \bibinfo{person}{Mohsen Farahpour}, \bibinfo{person}{Seyyed~Mohammad Razavi}, \bibinfo{person}{Saeed Nasseri}, \bibinfo{person}{Hossein Safarpour}, {et~al\mbox{.}}} \bibinfo{year}{2023}\natexlab{}.
\newblock \showarticletitle{Deep learning applications in single-cell genomics and transcriptomics data analysis}.
\newblock \bibinfo{journal}{\emph{Biomedicine \& Pharmacotherapy}}  \bibinfo{volume}{165} (\bibinfo{year}{2023}), \bibinfo{pages}{115077}.
\newblock


\bibitem[Fan et~al\mbox{.}(2019)]%
        {fan2019graph}
\bibfield{author}{\bibinfo{person}{Wenqi Fan}, \bibinfo{person}{Yao Ma}, \bibinfo{person}{Qing Li}, \bibinfo{person}{Yuan He}, \bibinfo{person}{Eric Zhao}, \bibinfo{person}{Jiliang Tang}, {and} \bibinfo{person}{Dawei Yin}.} \bibinfo{year}{2019}\natexlab{}.
\newblock \showarticletitle{Graph neural networks for social recommendation}. In \bibinfo{booktitle}{\emph{WWW}}. \bibinfo{pages}{417--426}.
\newblock


\bibitem[Fang et~al\mbox{.}(2023)]%
        {fang2023knowledge}
\bibfield{author}{\bibinfo{person}{Yin Fang}, \bibinfo{person}{Qiang Zhang}, \bibinfo{person}{Ningyu Zhang}, \bibinfo{person}{Zhuo Chen}, \bibinfo{person}{Xiang Zhuang}, \bibinfo{person}{Xin Shao}, \bibinfo{person}{Xiaohui Fan}, {and} \bibinfo{person}{Huajun Chen}.} \bibinfo{year}{2023}\natexlab{}.
\newblock \showarticletitle{Knowledge graph-enhanced molecular contrastive learning with functional prompt}.
\newblock \bibinfo{journal}{\emph{Nature Machine Intelligence}} \bibinfo{volume}{5}, \bibinfo{number}{5} (\bibinfo{year}{2023}), \bibinfo{pages}{542--553}.
\newblock


\bibitem[Feng et~al\mbox{.}(2022)]%
        {feng2022adversarial}
\bibfield{author}{\bibinfo{person}{Shengyu Feng}, \bibinfo{person}{Baoyu Jing}, \bibinfo{person}{Yada Zhu}, {and} \bibinfo{person}{Hanghang Tong}.} \bibinfo{year}{2022}\natexlab{}.
\newblock \showarticletitle{Adversarial graph contrastive learning with information regularization}. In \bibinfo{booktitle}{\emph{WWW}}. \bibinfo{pages}{1362--1371}.
\newblock


\bibitem[Fey et~al\mbox{.}(2019)]%
        {fey2019deep}
\bibfield{author}{\bibinfo{person}{Matthias Fey}, \bibinfo{person}{Jan~E Lenssen}, \bibinfo{person}{Christopher Morris}, \bibinfo{person}{Jonathan Masci}, {and} \bibinfo{person}{Nils~M Kriege}.} \bibinfo{year}{2019}\natexlab{}.
\newblock \showarticletitle{Deep Graph Matching Consensus}. In \bibinfo{booktitle}{\emph{ICLR}}.
\newblock


\bibitem[Freeman(2004)]%
        {freeman2004development}
\bibfield{author}{\bibinfo{person}{Linton Freeman}.} \bibinfo{year}{2004}\natexlab{}.
\newblock \showarticletitle{The development of social network analysis}.
\newblock \bibinfo{journal}{\emph{A Study in the Sociology of Science}} \bibinfo{volume}{1}, \bibinfo{number}{687} (\bibinfo{year}{2004}), \bibinfo{pages}{159--167}.
\newblock


\bibitem[Gao et~al\mbox{.}(2023)]%
        {gao2023similarity}
\bibfield{author}{\bibinfo{person}{Zihao Gao}, \bibinfo{person}{Huifang Ma}, \bibinfo{person}{Xiaohui Zhang}, \bibinfo{person}{Yike Wang}, {and} \bibinfo{person}{Zheyu Wu}.} \bibinfo{year}{2023}\natexlab{}.
\newblock \showarticletitle{Similarity measures-based graph co-contrastive learning for drug--disease association prediction}.
\newblock \bibinfo{journal}{\emph{Bioinformatics}} \bibinfo{volume}{39}, \bibinfo{number}{6} (\bibinfo{year}{2023}), \bibinfo{pages}{btad357}.
\newblock


\bibitem[Gasteiger et~al\mbox{.}(2019)]%
        {gasteiger2019diffusion}
\bibfield{author}{\bibinfo{person}{Johannes Gasteiger}, \bibinfo{person}{Stefan Wei{\ss}enberger}, {and} \bibinfo{person}{Stephan G{\"u}nnemann}.} \bibinfo{year}{2019}\natexlab{}.
\newblock \showarticletitle{Diffusion improves graph learning}.
\newblock \bibinfo{journal}{\emph{NeurIPS}}  \bibinfo{volume}{32} (\bibinfo{year}{2019}), \bibinfo{pages}{13333--13345}.
\newblock


\bibitem[Ghose et~al\mbox{.}(2023)]%
        {ghose2023spectral}
\bibfield{author}{\bibinfo{person}{Amur Ghose}, \bibinfo{person}{Yingxue Zhang}, \bibinfo{person}{Jianye Hao}, {and} \bibinfo{person}{Mark Coates}.} \bibinfo{year}{2023}\natexlab{}.
\newblock \showarticletitle{Spectral augmentations for graph contrastive learning}. In \bibinfo{booktitle}{\emph{AISTATS}}. \bibinfo{pages}{11213--11266}.
\newblock


\bibitem[Gilmer et~al\mbox{.}(2017)]%
        {gilmer2017neural}
\bibfield{author}{\bibinfo{person}{Justin Gilmer}, \bibinfo{person}{Samuel~S Schoenholz}, \bibinfo{person}{Patrick~F Riley}, \bibinfo{person}{Oriol Vinyals}, {and} \bibinfo{person}{George~E Dahl}.} \bibinfo{year}{2017}\natexlab{}.
\newblock \showarticletitle{Neural message passing for quantum chemistry}. In \bibinfo{booktitle}{\emph{ICML}}. \bibinfo{pages}{1263--1272}.
\newblock


\bibitem[Gong et~al\mbox{.}(2023)]%
        {gong2023ma}
\bibfield{author}{\bibinfo{person}{Xumeng Gong}, \bibinfo{person}{Cheng Yang}, {and} \bibinfo{person}{Chuan Shi}.} \bibinfo{year}{2023}\natexlab{}.
\newblock \showarticletitle{Ma-gcl: Model augmentation tricks for graph contrastive learning}. In \bibinfo{booktitle}{\emph{AAAI}}, Vol.~\bibinfo{volume}{37}. \bibinfo{pages}{4284--4292}.
\newblock


\bibitem[Grill et~al\mbox{.}(2020)]%
        {grill2020bootstrap}
\bibfield{author}{\bibinfo{person}{Jean-Bastien Grill}, \bibinfo{person}{Florian Strub}, \bibinfo{person}{Florent Altch{\'e}}, \bibinfo{person}{Corentin Tallec}, \bibinfo{person}{Pierre Richemond}, \bibinfo{person}{Elena Buchatskaya}, \bibinfo{person}{Carl Doersch}, \bibinfo{person}{Bernardo Avila~Pires}, \bibinfo{person}{Zhaohan Guo}, \bibinfo{person}{Mohammad Gheshlaghi~Azar}, {et~al\mbox{.}}} \bibinfo{year}{2020}\natexlab{}.
\newblock \showarticletitle{Bootstrap your own latent-a new approach to self-supervised learning}.
\newblock \bibinfo{journal}{\emph{NeurIPS}}  \bibinfo{volume}{33} (\bibinfo{year}{2020}), \bibinfo{pages}{21271--21284}.
\newblock


\bibitem[Gu et~al\mbox{.}(2023)]%
        {gu2023hierarchical}
\bibfield{author}{\bibinfo{person}{Zhonghui Gu}, \bibinfo{person}{Xiao Luo}, \bibinfo{person}{Jiaxiao Chen}, \bibinfo{person}{Minghua Deng}, {and} \bibinfo{person}{Luhua Lai}.} \bibinfo{year}{2023}\natexlab{}.
\newblock \showarticletitle{Hierarchical graph transformer with contrastive learning for protein function prediction}.
\newblock \bibinfo{journal}{\emph{Bioinformatics}} \bibinfo{volume}{39}, \bibinfo{number}{7} (\bibinfo{year}{2023}), \bibinfo{pages}{btad410}.
\newblock


\bibitem[Guo et~al\mbox{.}(2023)]%
        {guo2023manipulating}
\bibfield{author}{\bibinfo{person}{Jiayan Guo}, \bibinfo{person}{Lun Du}, \bibinfo{person}{Xu Chen}, \bibinfo{person}{Xiaojun Ma}, \bibinfo{person}{Qiang Fu}, \bibinfo{person}{Shi Han}, \bibinfo{person}{Dongmei Zhang}, {and} \bibinfo{person}{Yan Zhang}.} \bibinfo{year}{2023}\natexlab{}.
\newblock \showarticletitle{On Manipulating Signals of User-Item Graph: A Jacobi Polynomial-based Graph Collaborative Filtering}. In \bibinfo{booktitle}{\emph{KDD}}. \bibinfo{pages}{602--613}.
\newblock


\bibitem[Gupta et~al\mbox{.}(2021)]%
        {gupta2021artificial}
\bibfield{author}{\bibinfo{person}{Rohan Gupta}, \bibinfo{person}{Devesh Srivastava}, \bibinfo{person}{Mehar Sahu}, \bibinfo{person}{Swati Tiwari}, \bibinfo{person}{Rashmi~K Ambasta}, {and} \bibinfo{person}{Pravir Kumar}.} \bibinfo{year}{2021}\natexlab{}.
\newblock \showarticletitle{Artificial intelligence to deep learning: machine intelligence approach for drug discovery}.
\newblock \bibinfo{journal}{\emph{Molecular diversity}}  \bibinfo{volume}{25} (\bibinfo{year}{2021}), \bibinfo{pages}{1315--1360}.
\newblock


\bibitem[Gutmann and Hyv{\"a}rinen(2010)]%
        {gutmann2010noise}
\bibfield{author}{\bibinfo{person}{Michael Gutmann} {and} \bibinfo{person}{Aapo Hyv{\"a}rinen}.} \bibinfo{year}{2010}\natexlab{}.
\newblock \showarticletitle{Noise-contrastive estimation: A new estimation principle for unnormalized statistical models}. In \bibinfo{booktitle}{\emph{AISTATS}}. \bibinfo{pages}{297--304}.
\newblock


\bibitem[Hamilton et~al\mbox{.}(2017)]%
        {hamilton2017inductive}
\bibfield{author}{\bibinfo{person}{Will Hamilton}, \bibinfo{person}{Zhitao Ying}, {and} \bibinfo{person}{Jure Leskovec}.} \bibinfo{year}{2017}\natexlab{}.
\newblock \showarticletitle{Inductive representation learning on large graphs}.
\newblock \bibinfo{journal}{\emph{NeurIPS}}  \bibinfo{volume}{30} (\bibinfo{year}{2017}).
\newblock


\bibitem[Han et~al\mbox{.}(2022)]%
        {han2022generative}
\bibfield{author}{\bibinfo{person}{Yuehui Han}, \bibinfo{person}{Le Hui}, \bibinfo{person}{Haobo Jiang}, \bibinfo{person}{Jianjun Qian}, {and} \bibinfo{person}{Jin Xie}.} \bibinfo{year}{2022}\natexlab{}.
\newblock \showarticletitle{Generative Subgraph Contrast for Self-Supervised Graph Representation Learning}. In \bibinfo{booktitle}{\emph{ECCV}}. \bibinfo{pages}{91--107}.
\newblock


\bibitem[Hao et~al\mbox{.}(2020)]%
        {hao2020asgn}
\bibfield{author}{\bibinfo{person}{Zhongkai Hao}, \bibinfo{person}{Chengqiang Lu}, \bibinfo{person}{Zhenya Huang}, \bibinfo{person}{Hao Wang}, \bibinfo{person}{Zheyuan Hu}, \bibinfo{person}{Qi Liu}, \bibinfo{person}{Enhong Chen}, {and} \bibinfo{person}{Cheekong Lee}.} \bibinfo{year}{2020}\natexlab{}.
\newblock \showarticletitle{ASGN: An active semi-supervised graph neural network for molecular property prediction}. In \bibinfo{booktitle}{\emph{KDD}}. \bibinfo{pages}{731--752}.
\newblock


\bibitem[Hassani and Khasahmadi(2020)]%
        {hassani2020contrastive}
\bibfield{author}{\bibinfo{person}{Kaveh Hassani} {and} \bibinfo{person}{Amir~Hosein Khasahmadi}.} \bibinfo{year}{2020}\natexlab{}.
\newblock \showarticletitle{Contrastive multi-view representation learning on graphs}. In \bibinfo{booktitle}{\emph{ICML}}. \bibinfo{pages}{4116--4126}.
\newblock


\bibitem[He et~al\mbox{.}(2024)]%
        {he2024new}
\bibfield{author}{\bibinfo{person}{Dongxiao He}, \bibinfo{person}{Jitao Zhao}, \bibinfo{person}{Cuiying Huo}, \bibinfo{person}{Yongqi Huang}, \bibinfo{person}{Yuxiao Huang}, {and} \bibinfo{person}{Zhiyong Feng}.} \bibinfo{year}{2024}\natexlab{}.
\newblock \showarticletitle{A New Mechanism for Eliminating Implicit Conflict in Graph Contrastive Learning}. In \bibinfo{booktitle}{\emph{AAAI}}, Vol.~\bibinfo{volume}{38}. \bibinfo{pages}{12340--12348}.
\newblock


\bibitem[He et~al\mbox{.}(2020b)]%
        {he2020momentum}
\bibfield{author}{\bibinfo{person}{Kaiming He}, \bibinfo{person}{Haoqi Fan}, \bibinfo{person}{Yuxin Wu}, \bibinfo{person}{Saining Xie}, {and} \bibinfo{person}{Ross Girshick}.} \bibinfo{year}{2020}\natexlab{b}.
\newblock \showarticletitle{Momentum contrast for unsupervised visual representation learning}. In \bibinfo{booktitle}{\emph{CVPR}}. \bibinfo{pages}{9729--9738}.
\newblock


\bibitem[He et~al\mbox{.}(2020a)]%
        {he2020lightgcn}
\bibfield{author}{\bibinfo{person}{Xiangnan He}, \bibinfo{person}{Kuan Deng}, \bibinfo{person}{Xiang Wang}, \bibinfo{person}{Yan Li}, \bibinfo{person}{Yongdong Zhang}, {and} \bibinfo{person}{Meng Wang}.} \bibinfo{year}{2020}\natexlab{a}.
\newblock \showarticletitle{Lightgcn: Simplifying and powering graph convolution network for recommendation}. In \bibinfo{booktitle}{\emph{SIGIR}}. \bibinfo{pages}{639--648}.
\newblock


\bibitem[Hjelm et~al\mbox{.}(2018)]%
        {hjelm2018learning}
\bibfield{author}{\bibinfo{person}{R~Devon Hjelm}, \bibinfo{person}{Alex Fedorov}, \bibinfo{person}{Samuel Lavoie-Marchildon}, \bibinfo{person}{Karan Grewal}, \bibinfo{person}{Phil Bachman}, \bibinfo{person}{Adam Trischler}, {and} \bibinfo{person}{Yoshua Bengio}.} \bibinfo{year}{2018}\natexlab{}.
\newblock \showarticletitle{Learning Deep Representations by Mutual Information Estimation and Maximization}. In \bibinfo{booktitle}{\emph{ICLR}}.
\newblock


\bibitem[Hou et~al\mbox{.}(2020)]%
        {hou2020systematic}
\bibfield{author}{\bibinfo{person}{Wenpin Hou}, \bibinfo{person}{Zhicheng Ji}, \bibinfo{person}{Hongkai Ji}, {and} \bibinfo{person}{Stephanie~C Hicks}.} \bibinfo{year}{2020}\natexlab{}.
\newblock \showarticletitle{A systematic evaluation of single-cell RNA-sequencing imputation methods}.
\newblock \bibinfo{journal}{\emph{Genome biology}}  \bibinfo{volume}{21} (\bibinfo{year}{2020}), \bibinfo{pages}{1--30}.
\newblock


\bibitem[Hu et~al\mbox{.}(2019)]%
        {hu2019strategies}
\bibfield{author}{\bibinfo{person}{Weihua Hu}, \bibinfo{person}{Bowen Liu}, \bibinfo{person}{Joseph Gomes}, \bibinfo{person}{Marinka Zitnik}, \bibinfo{person}{Percy Liang}, \bibinfo{person}{Vijay Pande}, {and} \bibinfo{person}{Jure Leskovec}.} \bibinfo{year}{2019}\natexlab{}.
\newblock \showarticletitle{Strategies for pre-training graph neural networks}.
\newblock \bibinfo{journal}{\emph{arXiv preprint arXiv:1905.12265}} (\bibinfo{year}{2019}).
\newblock


\bibitem[Hu et~al\mbox{.}(2020)]%
        {hu2020gpt}
\bibfield{author}{\bibinfo{person}{Ziniu Hu}, \bibinfo{person}{Yuxiao Dong}, \bibinfo{person}{Kuansan Wang}, \bibinfo{person}{Kai-Wei Chang}, {and} \bibinfo{person}{Yizhou Sun}.} \bibinfo{year}{2020}\natexlab{}.
\newblock \showarticletitle{Gpt-gnn: Generative pre-training of graph neural networks}. In \bibinfo{booktitle}{\emph{KDD}}. \bibinfo{pages}{1857--1867}.
\newblock


\bibitem[In et~al\mbox{.}(2023)]%
        {in2023similarity}
\bibfield{author}{\bibinfo{person}{Yeonjun In}, \bibinfo{person}{Kanghoon Yoon}, {and} \bibinfo{person}{Chanyoung Park}.} \bibinfo{year}{2023}\natexlab{}.
\newblock \showarticletitle{Similarity preserving adversarial graph contrastive learning}. In \bibinfo{booktitle}{\emph{KDD}}. \bibinfo{pages}{867--878}.
\newblock


\bibitem[Ji et~al\mbox{.}(2024)]%
        {ji2024contrastive}
\bibfield{author}{\bibinfo{person}{Changtao Ji}, \bibinfo{person}{Yan Xu}, \bibinfo{person}{Yu Lu}, \bibinfo{person}{Xiaoyu Huang}, {and} \bibinfo{person}{Yuzhe Zhu}.} \bibinfo{year}{2024}\natexlab{}.
\newblock \showarticletitle{Contrastive Learning-Based Adaptive Graph Fusion Convolution Network With Residual-Enhanced Decomposition Strategy for Traffic Flow Forecasting}.
\newblock \bibinfo{journal}{\emph{IoT}} (\bibinfo{year}{2024}).
\newblock


\bibitem[Jiang et~al\mbox{.}(2023)]%
        {jiang2023rcenr}
\bibfield{author}{\bibinfo{person}{Hao Jiang}, \bibinfo{person}{Chuanzhen Li}, \bibinfo{person}{Juanjuan Cai}, {and} \bibinfo{person}{Jingling Wang}.} \bibinfo{year}{2023}\natexlab{}.
\newblock \showarticletitle{RCENR: A Reinforced and Contrastive Heterogeneous Network Reasoning Model for Explainable News Recommendation}. In \bibinfo{booktitle}{\emph{SIGIR}}. \bibinfo{pages}{1710--1720}.
\newblock


\bibitem[Jiao et~al\mbox{.}(2020)]%
        {jiao2020sub}
\bibfield{author}{\bibinfo{person}{Yizhu Jiao}, \bibinfo{person}{Yun Xiong}, \bibinfo{person}{Jiawei Zhang}, \bibinfo{person}{Yao Zhang}, \bibinfo{person}{Tianqi Zhang}, {and} \bibinfo{person}{Yangyong Zhu}.} \bibinfo{year}{2020}\natexlab{}.
\newblock \showarticletitle{Sub-graph contrast for scalable self-supervised graph representation learning}. In \bibinfo{booktitle}{\emph{ICDM}}. \bibinfo{pages}{222--231}.
\newblock


\bibitem[Jin et~al\mbox{.}(2021b)]%
        {jin2021multi}
\bibfield{author}{\bibinfo{person}{Ming Jin}, \bibinfo{person}{Yizhen Zheng}, \bibinfo{person}{Yuan-Fang Li}, \bibinfo{person}{Chen Gong}, \bibinfo{person}{Chuan Zhou}, {and} \bibinfo{person}{Shirui Pan}.} \bibinfo{year}{2021}\natexlab{b}.
\newblock \showarticletitle{Multi-scale contrastive siamese networks for self-supervised graph representation learning}. In \bibinfo{booktitle}{\emph{IJCAI}}. \bibinfo{pages}{1477--1483}.
\newblock


\bibitem[Jin et~al\mbox{.}(2020)]%
        {jin2020self}
\bibfield{author}{\bibinfo{person}{Wei Jin}, \bibinfo{person}{Tyler Derr}, \bibinfo{person}{Haochen Liu}, \bibinfo{person}{Yiqi Wang}, \bibinfo{person}{Suhang Wang}, \bibinfo{person}{Zitao Liu}, {and} \bibinfo{person}{Jiliang Tang}.} \bibinfo{year}{2020}\natexlab{}.
\newblock \showarticletitle{Self-supervised learning on graphs: Deep insights and new direction}.
\newblock \bibinfo{journal}{\emph{arXiv preprint arXiv:2006.10141}} (\bibinfo{year}{2020}).
\newblock


\bibitem[Jin et~al\mbox{.}(2021a)]%
        {jin2021node}
\bibfield{author}{\bibinfo{person}{Wei Jin}, \bibinfo{person}{Tyler Derr}, \bibinfo{person}{Yiqi Wang}, \bibinfo{person}{Yao Ma}, \bibinfo{person}{Zitao Liu}, {and} \bibinfo{person}{Jiliang Tang}.} \bibinfo{year}{2021}\natexlab{a}.
\newblock \showarticletitle{Node similarity preserving graph convolutional networks}. In \bibinfo{booktitle}{\emph{WSDM}}. \bibinfo{pages}{148--156}.
\newblock


\bibitem[Jin et~al\mbox{.}(2023)]%
        {jin2023empowering}
\bibfield{author}{\bibinfo{person}{Wei Jin}, \bibinfo{person}{Tong Zhao}, \bibinfo{person}{Jiayuan Ding}, \bibinfo{person}{Yozen Liu}, \bibinfo{person}{Jiliang Tang}, {and} \bibinfo{person}{Neil Shah}.} \bibinfo{year}{2023}\natexlab{}.
\newblock \showarticletitle{Empowering graph representation learning with test-time graph transformation}. In \bibinfo{booktitle}{\emph{ICLR}}.
\newblock


\bibitem[Jing and Tian(2020)]%
        {jing2020self}
\bibfield{author}{\bibinfo{person}{Longlong Jing} {and} \bibinfo{person}{Yingli Tian}.} \bibinfo{year}{2020}\natexlab{}.
\newblock \showarticletitle{Self-supervised visual feature learning with deep neural networks: A survey}.
\newblock \bibinfo{journal}{\emph{TPAMI}} \bibinfo{volume}{43}, \bibinfo{number}{11} (\bibinfo{year}{2020}), \bibinfo{pages}{4037--4058}.
\newblock


\bibitem[Ju et~al\mbox{.}(2024a)]%
        {ju2024comprehensive}
\bibfield{author}{\bibinfo{person}{Wei Ju}, \bibinfo{person}{Zheng Fang}, \bibinfo{person}{Yiyang Gu}, \bibinfo{person}{Zequn Liu}, \bibinfo{person}{Qingqing Long}, \bibinfo{person}{Ziyue Qiao}, \bibinfo{person}{Yifang Qin}, \bibinfo{person}{Jianhao Shen}, \bibinfo{person}{Fang Sun}, \bibinfo{person}{Zhiping Xiao}, {et~al\mbox{.}}} \bibinfo{year}{2024}\natexlab{a}.
\newblock \showarticletitle{A comprehensive survey on deep graph representation learning}.
\newblock \bibinfo{journal}{\emph{Neural Networks}} (\bibinfo{year}{2024}), \bibinfo{pages}{106207}.
\newblock


\bibitem[Ju et~al\mbox{.}(2023a)]%
        {ju2023glcc}
\bibfield{author}{\bibinfo{person}{Wei Ju}, \bibinfo{person}{Yiyang Gu}, \bibinfo{person}{Binqi Chen}, \bibinfo{person}{Gongbo Sun}, \bibinfo{person}{Yifang Qin}, \bibinfo{person}{Xingyuming Liu}, \bibinfo{person}{Xiao Luo}, {and} \bibinfo{person}{Ming Zhang}.} \bibinfo{year}{2023}\natexlab{a}.
\newblock \showarticletitle{Glcc: A general framework for graph-level clustering}. In \bibinfo{booktitle}{\emph{AAAI}}, Vol.~\bibinfo{volume}{37}. \bibinfo{pages}{4391--4399}.
\newblock


\bibitem[Ju et~al\mbox{.}(2023b)]%
        {ju2023unsupervised}
\bibfield{author}{\bibinfo{person}{Wei Ju}, \bibinfo{person}{Yiyang Gu}, \bibinfo{person}{Xiao Luo}, \bibinfo{person}{Yifan Wang}, \bibinfo{person}{Haochen Yuan}, \bibinfo{person}{Huasong Zhong}, {and} \bibinfo{person}{Ming Zhang}.} \bibinfo{year}{2023}\natexlab{b}.
\newblock \showarticletitle{Unsupervised graph-level representation learning with hierarchical contrasts}.
\newblock \bibinfo{journal}{\emph{Neural Networks}}  \bibinfo{volume}{158} (\bibinfo{year}{2023}), \bibinfo{pages}{359--368}.
\newblock


\bibitem[Ju et~al\mbox{.}(2023c)]%
        {ju2023few}
\bibfield{author}{\bibinfo{person}{Wei Ju}, \bibinfo{person}{Zequn Liu}, \bibinfo{person}{Yifang Qin}, \bibinfo{person}{Bin Feng}, \bibinfo{person}{Chen Wang}, \bibinfo{person}{Zhihui Guo}, \bibinfo{person}{Xiao Luo}, {and} \bibinfo{person}{Ming Zhang}.} \bibinfo{year}{2023}\natexlab{c}.
\newblock \showarticletitle{Few-shot molecular property prediction via Hierarchically Structured Learning on Relation Graphs}.
\newblock \bibinfo{journal}{\emph{Neural Networks}}  \bibinfo{volume}{163} (\bibinfo{year}{2023}), \bibinfo{pages}{122--131}.
\newblock


\bibitem[Ju et~al\mbox{.}(2024b)]%
        {ju2024hypergraph}
\bibfield{author}{\bibinfo{person}{Wei Ju}, \bibinfo{person}{Zhengyang Mao}, \bibinfo{person}{Siyu Yi}, \bibinfo{person}{Yifang Qin}, \bibinfo{person}{Yiyang Gu}, \bibinfo{person}{Zhiping Xiao}, \bibinfo{person}{Yifan Wang}, \bibinfo{person}{Xiao Luo}, {and} \bibinfo{person}{Ming Zhang}.} \bibinfo{year}{2024}\natexlab{b}.
\newblock \showarticletitle{Hypergraph-enhanced Dual Semi-supervised Graph Classification}.
\newblock \bibinfo{journal}{\emph{arXiv preprint arXiv:2405.04773}} (\bibinfo{year}{2024}).
\newblock


\bibitem[Ju et~al\mbox{.}(2022a)]%
        {ju2022kernel}
\bibfield{author}{\bibinfo{person}{Wei Ju}, \bibinfo{person}{Yifang Qin}, \bibinfo{person}{Ziyue Qiao}, \bibinfo{person}{Xiao Luo}, \bibinfo{person}{Yifan Wang}, \bibinfo{person}{Yanjie Fu}, {and} \bibinfo{person}{Ming Zhang}.} \bibinfo{year}{2022}\natexlab{a}.
\newblock \showarticletitle{Kernel-based substructure exploration for next POI recommendation}. In \bibinfo{booktitle}{\emph{ICDM}}. \bibinfo{pages}{221--230}.
\newblock


\bibitem[Ju et~al\mbox{.}(2022b)]%
        {ju2022kgnn}
\bibfield{author}{\bibinfo{person}{Wei Ju}, \bibinfo{person}{Junwei Yang}, \bibinfo{person}{Meng Qu}, \bibinfo{person}{Weiping Song}, \bibinfo{person}{Jianhao Shen}, {and} \bibinfo{person}{Ming Zhang}.} \bibinfo{year}{2022}\natexlab{b}.
\newblock \showarticletitle{Kgnn: Harnessing kernel-based networks for semi-supervised graph classification}. In \bibinfo{booktitle}{\emph{WSDM}}. \bibinfo{pages}{421--429}.
\newblock


\bibitem[Ju et~al\mbox{.}(2024c)]%
        {ju2024survey}
\bibfield{author}{\bibinfo{person}{Wei Ju}, \bibinfo{person}{Siyu Yi}, \bibinfo{person}{Yifan Wang}, \bibinfo{person}{Qingqing Long}, \bibinfo{person}{Junyu Luo}, \bibinfo{person}{Zhiping Xiao}, {and} \bibinfo{person}{Ming Zhang}.} \bibinfo{year}{2024}\natexlab{c}.
\newblock \showarticletitle{A survey of data-efficient graph learning}.
\newblock \bibinfo{journal}{\emph{IJCAI}} (\bibinfo{year}{2024}).
\newblock


\bibitem[Ju et~al\mbox{.}(2024d)]%
        {ju2024survey_realworld}
\bibfield{author}{\bibinfo{person}{Wei Ju}, \bibinfo{person}{Siyu Yi}, \bibinfo{person}{Yifan Wang}, \bibinfo{person}{Zhiping Xiao}, \bibinfo{person}{Zhengyang Mao}, \bibinfo{person}{Hourun Li}, \bibinfo{person}{Yiyang Gu}, \bibinfo{person}{Yifang Qin}, \bibinfo{person}{Nan Yin}, \bibinfo{person}{Senzhang Wang}, {et~al\mbox{.}}} \bibinfo{year}{2024}\natexlab{d}.
\newblock \showarticletitle{A Survey of Graph Neural Networks in Real world: Imbalance, Noise, Privacy and OOD Challenges}.
\newblock \bibinfo{journal}{\emph{arXiv preprint arXiv:2403.04468}} (\bibinfo{year}{2024}).
\newblock


\bibitem[Khosla et~al\mbox{.}(2020)]%
        {khosla2020supervised}
\bibfield{author}{\bibinfo{person}{Prannay Khosla}, \bibinfo{person}{Piotr Teterwak}, \bibinfo{person}{Chen Wang}, \bibinfo{person}{Aaron Sarna}, \bibinfo{person}{Yonglong Tian}, \bibinfo{person}{Phillip Isola}, \bibinfo{person}{Aaron Maschinot}, \bibinfo{person}{Ce Liu}, {and} \bibinfo{person}{Dilip Krishnan}.} \bibinfo{year}{2020}\natexlab{}.
\newblock \showarticletitle{Supervised contrastive learning}.
\newblock \bibinfo{journal}{\emph{NeurIPS}}  \bibinfo{volume}{33} (\bibinfo{year}{2020}), \bibinfo{pages}{18661--18673}.
\newblock


\bibitem[Kipf and Welling(2016)]%
        {kipf2016semi}
\bibfield{author}{\bibinfo{person}{Thomas~N Kipf} {and} \bibinfo{person}{Max Welling}.} \bibinfo{year}{2016}\natexlab{}.
\newblock \showarticletitle{Semi-supervised classification with graph convolutional networks}.
\newblock \bibinfo{journal}{\emph{arXiv preprint arXiv:1609.02907}} (\bibinfo{year}{2016}).
\newblock


\bibitem[Kipf and Welling(2017)]%
        {kipf2017semi}
\bibfield{author}{\bibinfo{person}{Thomas~N Kipf} {and} \bibinfo{person}{Max Welling}.} \bibinfo{year}{2017}\natexlab{}.
\newblock \showarticletitle{Semi-supervised classification with graph convolutional networks}.
\newblock In \bibinfo{booktitle}{\emph{ICLR}}.
\newblock


\bibitem[Kondor and Lafferty(2002)]%
        {kondor2002diffusion}
\bibfield{author}{\bibinfo{person}{Risi~Imre Kondor} {and} \bibinfo{person}{John Lafferty}.} \bibinfo{year}{2002}\natexlab{}.
\newblock \showarticletitle{Diffusion kernels on graphs and other discrete structures}. In \bibinfo{booktitle}{\emph{ICML}}, Vol.~\bibinfo{volume}{2002}. \bibinfo{pages}{315--322}.
\newblock


\bibitem[Kose and Shen(2022)]%
        {kose2022fair}
\bibfield{author}{\bibinfo{person}{Oyku~Deniz Kose} {and} \bibinfo{person}{Yanning Shen}.} \bibinfo{year}{2022}\natexlab{}.
\newblock \showarticletitle{Fair contrastive learning on graphs}.
\newblock \bibinfo{journal}{\emph{TSIPN}}  \bibinfo{volume}{8} (\bibinfo{year}{2022}), \bibinfo{pages}{475--488}.
\newblock


\bibitem[Lee et~al\mbox{.}(2020)]%
        {lee2020single}
\bibfield{author}{\bibinfo{person}{Jeongwoo Lee}, \bibinfo{person}{Do~Young Hyeon}, {and} \bibinfo{person}{Daehee Hwang}.} \bibinfo{year}{2020}\natexlab{}.
\newblock \showarticletitle{Single-cell multiomics: technologies and data analysis methods}.
\newblock \bibinfo{journal}{\emph{Experimental \& Molecular Medicine}} \bibinfo{volume}{52}, \bibinfo{number}{9} (\bibinfo{year}{2020}), \bibinfo{pages}{1428--1442}.
\newblock


\bibitem[Lee et~al\mbox{.}(2023)]%
        {lee2023deep}
\bibfield{author}{\bibinfo{person}{Junseok Lee}, \bibinfo{person}{Sungwon Kim}, \bibinfo{person}{Dongmin Hyun}, \bibinfo{person}{Namkyeong Lee}, \bibinfo{person}{Yejin Kim}, {and} \bibinfo{person}{Chanyoung Park}.} \bibinfo{year}{2023}\natexlab{}.
\newblock \showarticletitle{Deep single-cell RNA-seq data clustering with graph prototypical contrastive learning}.
\newblock \bibinfo{journal}{\emph{Bioinformatics}} \bibinfo{volume}{39}, \bibinfo{number}{6} (\bibinfo{year}{2023}), \bibinfo{pages}{btad342}.
\newblock


\bibitem[Li et~al\mbox{.}(2022b)]%
        {li2022graph}
\bibfield{author}{\bibinfo{person}{Bolian Li}, \bibinfo{person}{Baoyu Jing}, {and} \bibinfo{person}{Hanghang Tong}.} \bibinfo{year}{2022}\natexlab{b}.
\newblock \showarticletitle{Graph communal contrastive learning}. In \bibinfo{booktitle}{\emph{WWW}}. \bibinfo{pages}{1203--1213}.
\newblock


\bibitem[Li et~al\mbox{.}(2024b)]%
        {li2024survey}
\bibfield{author}{\bibinfo{person}{Hourun Li}, \bibinfo{person}{Yusheng Zhao}, \bibinfo{person}{Zhengyang Mao}, \bibinfo{person}{Yifang Qin}, \bibinfo{person}{Zhiping Xiao}, \bibinfo{person}{Jiaqi Feng}, \bibinfo{person}{Yiyang Gu}, \bibinfo{person}{Wei Ju}, \bibinfo{person}{Xiao Luo}, {and} \bibinfo{person}{Ming Zhang}.} \bibinfo{year}{2024}\natexlab{b}.
\newblock \showarticletitle{A survey on graph neural networks in intelligent transportation systems}.
\newblock \bibinfo{journal}{\emph{arXiv preprint arXiv:2401.00713}} (\bibinfo{year}{2024}).
\newblock


\bibitem[Li et~al\mbox{.}(2021)]%
        {li2021comatch}
\bibfield{author}{\bibinfo{person}{Junnan Li}, \bibinfo{person}{Caiming Xiong}, {and} \bibinfo{person}{Steven~CH Hoi}.} \bibinfo{year}{2021}\natexlab{}.
\newblock \showarticletitle{Comatch: Semi-supervised learning with contrastive graph regularization}. In \bibinfo{booktitle}{\emph{ICCV}}. \bibinfo{pages}{9475--9484}.
\newblock


\bibitem[Li et~al\mbox{.}(2022a)]%
        {li2022graph_healthcare}
\bibfield{author}{\bibinfo{person}{Michelle~M Li}, \bibinfo{person}{Kexin Huang}, {and} \bibinfo{person}{Marinka Zitnik}.} \bibinfo{year}{2022}\natexlab{a}.
\newblock \showarticletitle{Graph representation learning in biomedicine and healthcare}.
\newblock \bibinfo{journal}{\emph{Nature Biomedical Engineering}} \bibinfo{volume}{6}, \bibinfo{number}{12} (\bibinfo{year}{2022}), \bibinfo{pages}{1353--1369}.
\newblock


\bibitem[Li et~al\mbox{.}(2022c)]%
        {li2022let}
\bibfield{author}{\bibinfo{person}{Sihang Li}, \bibinfo{person}{Xiang Wang}, \bibinfo{person}{An Zhang}, \bibinfo{person}{Yingxin Wu}, \bibinfo{person}{Xiangnan He}, {and} \bibinfo{person}{Tat-Seng Chua}.} \bibinfo{year}{2022}\natexlab{c}.
\newblock \showarticletitle{Let invariant rationale discovery inspire graph contrastive learning}. In \bibinfo{booktitle}{\emph{ICML}}. \bibinfo{pages}{13052--13065}.
\newblock


\bibitem[Li et~al\mbox{.}(2022d)]%
        {li2022geomgcl}
\bibfield{author}{\bibinfo{person}{Shuangli Li}, \bibinfo{person}{Jingbo Zhou}, \bibinfo{person}{Tong Xu}, \bibinfo{person}{Dejing Dou}, {and} \bibinfo{person}{Hui Xiong}.} \bibinfo{year}{2022}\natexlab{d}.
\newblock \showarticletitle{Geomgcl: Geometric graph contrastive learning for molecular property prediction}. In \bibinfo{booktitle}{\emph{AAAI}}, Vol.~\bibinfo{volume}{36}. \bibinfo{pages}{4541--4549}.
\newblock


\bibitem[Li et~al\mbox{.}(2024a)]%
        {li2024contrastive}
\bibfield{author}{\bibinfo{person}{Xianxian Li}, \bibinfo{person}{Qiyu Li}, \bibinfo{person}{Haodong Qian}, \bibinfo{person}{Jinyan Wang}, {et~al\mbox{.}}} \bibinfo{year}{2024}\natexlab{a}.
\newblock \showarticletitle{Contrastive learning of graphs under label noise}.
\newblock \bibinfo{journal}{\emph{Neural Networks}} (\bibinfo{year}{2024}), \bibinfo{pages}{106113}.
\newblock


\bibitem[Lin et~al\mbox{.}(2023)]%
        {lin2023certifiably}
\bibfield{author}{\bibinfo{person}{Minhua Lin}, \bibinfo{person}{Teng Xiao}, \bibinfo{person}{Enyan Dai}, \bibinfo{person}{Xiang Zhang}, {and} \bibinfo{person}{Suhang Wang}.} \bibinfo{year}{2023}\natexlab{}.
\newblock \showarticletitle{Certifiably Robust Graph Contrastive Learning}.
\newblock \bibinfo{journal}{\emph{NeurIPS}}  \bibinfo{volume}{36} (\bibinfo{year}{2023}).
\newblock


\bibitem[Ling et~al\mbox{.}(2022)]%
        {ling2022learning}
\bibfield{author}{\bibinfo{person}{Hongyi Ling}, \bibinfo{person}{Zhimeng Jiang}, \bibinfo{person}{Youzhi Luo}, \bibinfo{person}{Shuiwang Ji}, {and} \bibinfo{person}{Na Zou}.} \bibinfo{year}{2022}\natexlab{}.
\newblock \showarticletitle{Learning fair graph representations via automated data augmentations}. In \bibinfo{booktitle}{\emph{ICLR}}.
\newblock


\bibitem[Linsker(1988)]%
        {linsker1988self}
\bibfield{author}{\bibinfo{person}{Ralph Linsker}.} \bibinfo{year}{1988}\natexlab{}.
\newblock \showarticletitle{Self-organization in a perceptual network}.
\newblock \bibinfo{journal}{\emph{Computer}} \bibinfo{volume}{21}, \bibinfo{number}{3} (\bibinfo{year}{1988}), \bibinfo{pages}{105--117}.
\newblock


\bibitem[Liu et~al\mbox{.}(2021c)]%
        {liu2021noise}
\bibfield{author}{\bibinfo{person}{Chang Liu}, \bibinfo{person}{Han Yu}, \bibinfo{person}{Boyang Li}, \bibinfo{person}{Zhiqi Shen}, \bibinfo{person}{Zhanning Gao}, \bibinfo{person}{Peiran Ren}, \bibinfo{person}{Xuansong Xie}, \bibinfo{person}{Lizhen Cui}, {and} \bibinfo{person}{Chunyan Miao}.} \bibinfo{year}{2021}\natexlab{c}.
\newblock \showarticletitle{Noise-resistant deep metric learning with ranking-based instance selection}. In \bibinfo{booktitle}{\emph{CVPR}}. \bibinfo{pages}{6811--6820}.
\newblock


\bibitem[Liu et~al\mbox{.}(2023c)]%
        {liu2023b2}
\bibfield{author}{\bibinfo{person}{Mengyue Liu}, \bibinfo{person}{Yun Lin}, \bibinfo{person}{Jun Liu}, \bibinfo{person}{Bohao Liu}, \bibinfo{person}{Qinghua Zheng}, {and} \bibinfo{person}{Jin~Song Dong}.} \bibinfo{year}{2023}\natexlab{c}.
\newblock \showarticletitle{B2-Sampling: Fusing Balanced and Biased Sampling for Graph Contrastive Learning}. In \bibinfo{booktitle}{\emph{KDD}}. \bibinfo{pages}{1489--1500}.
\newblock


\bibitem[Liu et~al\mbox{.}(2021a)]%
        {liu2021simultaneous}
\bibfield{author}{\bibinfo{person}{Qiao Liu}, \bibinfo{person}{Shengquan Chen}, \bibinfo{person}{Rui Jiang}, {and} \bibinfo{person}{Wing~Hung Wong}.} \bibinfo{year}{2021}\natexlab{a}.
\newblock \showarticletitle{Simultaneous deep generative modelling and clustering of single-cell genomic data}.
\newblock \bibinfo{journal}{\emph{Nature machine intelligence}} \bibinfo{volume}{3}, \bibinfo{number}{6} (\bibinfo{year}{2021}), \bibinfo{pages}{536--544}.
\newblock


\bibitem[Liu et~al\mbox{.}(2024b)]%
        {liu2024muse}
\bibfield{author}{\bibinfo{person}{Tianyu Liu}, \bibinfo{person}{Yuge Wang}, \bibinfo{person}{Rex Ying}, {and} \bibinfo{person}{Hongyu Zhao}.} \bibinfo{year}{2024}\natexlab{b}.
\newblock \showarticletitle{MuSe-GNN: Learning Unified Gene Representation From Multimodal Biological Graph Data}.
\newblock \bibinfo{journal}{\emph{NeurIPS}}  \bibinfo{volume}{36} (\bibinfo{year}{2024}).
\newblock


\bibitem[Liu et~al\mbox{.}(2022b)]%
        {liu2022contrastive}
\bibfield{author}{\bibinfo{person}{Xu Liu}, \bibinfo{person}{Yuxuan Liang}, \bibinfo{person}{Chao Huang}, \bibinfo{person}{Yu Zheng}, \bibinfo{person}{Bryan Hooi}, {and} \bibinfo{person}{Roger Zimmermann}.} \bibinfo{year}{2022}\natexlab{b}.
\newblock \showarticletitle{When do contrastive learning signals help spatio-temporal graph forecasting?}. In \bibinfo{booktitle}{\emph{SIGSPATIAL}}. \bibinfo{pages}{1--12}.
\newblock


\bibitem[Liu et~al\mbox{.}(2021d)]%
        {liu2021self}
\bibfield{author}{\bibinfo{person}{Xiao Liu}, \bibinfo{person}{Fanjin Zhang}, \bibinfo{person}{Zhenyu Hou}, \bibinfo{person}{Li Mian}, \bibinfo{person}{Zhaoyu Wang}, \bibinfo{person}{Jing Zhang}, {and} \bibinfo{person}{Jie Tang}.} \bibinfo{year}{2021}\natexlab{d}.
\newblock \showarticletitle{Self-supervised learning: Generative or contrastive}.
\newblock \bibinfo{journal}{\emph{TKDE}} \bibinfo{volume}{35}, \bibinfo{number}{1} (\bibinfo{year}{2021}), \bibinfo{pages}{857--876}.
\newblock


\bibitem[Liu et~al\mbox{.}(2023a)]%
        {liu2023flood}
\bibfield{author}{\bibinfo{person}{Yang Liu}, \bibinfo{person}{Xiang Ao}, \bibinfo{person}{Fuli Feng}, \bibinfo{person}{Yunshan Ma}, \bibinfo{person}{Kuan Li}, \bibinfo{person}{Tat-Seng Chua}, {and} \bibinfo{person}{Qing He}.} \bibinfo{year}{2023}\natexlab{a}.
\newblock \showarticletitle{FLOOD: A flexible invariant learning framework for out-of-distribution generalization on graphs}. In \bibinfo{booktitle}{\emph{KDD}}. \bibinfo{pages}{1548--1558}.
\newblock


\bibitem[Liu et~al\mbox{.}(2023b)]%
        {liu2023good}
\bibfield{author}{\bibinfo{person}{Yixin Liu}, \bibinfo{person}{Kaize Ding}, \bibinfo{person}{Huan Liu}, {and} \bibinfo{person}{Shirui Pan}.} \bibinfo{year}{2023}\natexlab{b}.
\newblock \showarticletitle{{GOOD-D}: On unsupervised graph out-of-distribution detection}. In \bibinfo{booktitle}{\emph{WSDM}}. \bibinfo{pages}{339--347}.
\newblock


\bibitem[Liu et~al\mbox{.}(2024a)]%
        {liu2024towards}
\bibfield{author}{\bibinfo{person}{Yixin Liu}, \bibinfo{person}{Kaize Ding}, \bibinfo{person}{Qinghua Lu}, \bibinfo{person}{Fuyi Li}, \bibinfo{person}{Leo~Yu Zhang}, {and} \bibinfo{person}{Shirui Pan}.} \bibinfo{year}{2024}\natexlab{a}.
\newblock \showarticletitle{Towards self-interpretable graph-level anomaly detection}.
\newblock \bibinfo{journal}{\emph{NeurIPS}}  \bibinfo{volume}{36} (\bibinfo{year}{2024}).
\newblock


\bibitem[Liu et~al\mbox{.}(2022a)]%
        {liu2022graph}
\bibfield{author}{\bibinfo{person}{Yixin Liu}, \bibinfo{person}{Ming Jin}, \bibinfo{person}{Shirui Pan}, \bibinfo{person}{Chuan Zhou}, \bibinfo{person}{Yu Zheng}, \bibinfo{person}{Feng Xia}, {and} \bibinfo{person}{S~Yu Philip}.} \bibinfo{year}{2022}\natexlab{a}.
\newblock \showarticletitle{Graph self-supervised learning: A survey}.
\newblock \bibinfo{journal}{\emph{TKDE}} \bibinfo{volume}{35}, \bibinfo{number}{6} (\bibinfo{year}{2022}), \bibinfo{pages}{5879--5900}.
\newblock


\bibitem[Liu et~al\mbox{.}(2023d)]%
        {liu2023multi}
\bibfield{author}{\bibinfo{person}{Yanbei Liu}, \bibinfo{person}{Yu Zhao}, \bibinfo{person}{Xiao Wang}, \bibinfo{person}{Lei Geng}, {and} \bibinfo{person}{Zhitao Xiao}.} \bibinfo{year}{2023}\natexlab{d}.
\newblock \showarticletitle{Multi-scale subgraph contrastive learning}. In \bibinfo{booktitle}{\emph{IJCAI}}. \bibinfo{pages}{2215--2223}.
\newblock


\bibitem[Liu et~al\mbox{.}(2022c)]%
        {liu2022towards}
\bibfield{author}{\bibinfo{person}{Yixin Liu}, \bibinfo{person}{Yu Zheng}, \bibinfo{person}{Daokun Zhang}, \bibinfo{person}{Hongxu Chen}, \bibinfo{person}{Hao Peng}, {and} \bibinfo{person}{Shirui Pan}.} \bibinfo{year}{2022}\natexlab{c}.
\newblock \showarticletitle{Towards unsupervised deep graph structure learning}. In \bibinfo{booktitle}{\emph{WWW}}. \bibinfo{pages}{1392--1403}.
\newblock


\bibitem[Liu et~al\mbox{.}(2021b)]%
        {liu2021contrastive}
\bibfield{author}{\bibinfo{person}{Zhuang Liu}, \bibinfo{person}{Yunpu Ma}, \bibinfo{person}{Yuanxin Ouyang}, {and} \bibinfo{person}{Zhang Xiong}.} \bibinfo{year}{2021}\natexlab{b}.
\newblock \showarticletitle{Contrastive learning for recommender system}.
\newblock \bibinfo{journal}{\emph{arXiv preprint arXiv:2101.01317}} (\bibinfo{year}{2021}).
\newblock


\bibitem[Long et~al\mbox{.}(2021a)]%
        {long2021theoretically}
\bibfield{author}{\bibinfo{person}{Qingqing Long}, \bibinfo{person}{Yilun Jin}, \bibinfo{person}{Yi Wu}, {and} \bibinfo{person}{Guojie Song}.} \bibinfo{year}{2021}\natexlab{a}.
\newblock \showarticletitle{Theoretically improving graph neural networks via anonymous walk graph kernels}. In \bibinfo{booktitle}{\emph{WWW}}. \bibinfo{pages}{1204--1214}.
\newblock


\bibitem[Long et~al\mbox{.}(2021b)]%
        {long2021hgk}
\bibfield{author}{\bibinfo{person}{Qingqing Long}, \bibinfo{person}{Lingjun Xu}, \bibinfo{person}{Zheng Fang}, {and} \bibinfo{person}{Guojie Song}.} \bibinfo{year}{2021}\natexlab{b}.
\newblock \showarticletitle{HGK-GNN: Heterogeneous Graph Kernel based Graph Neural Networks}. In \bibinfo{booktitle}{\emph{KDD}}. \bibinfo{pages}{1129--1138}.
\newblock


\bibitem[Lu et~al\mbox{.}(2023)]%
        {lu2023pseudo}
\bibfield{author}{\bibinfo{person}{Weigang Lu}, \bibinfo{person}{Ziyu Guan}, \bibinfo{person}{Wei Zhao}, \bibinfo{person}{Yaming Yang}, \bibinfo{person}{Yuanhai Lv}, \bibinfo{person}{Lining Xing}, \bibinfo{person}{Baosheng Yu}, {and} \bibinfo{person}{Dacheng Tao}.} \bibinfo{year}{2023}\natexlab{}.
\newblock \showarticletitle{Pseudo contrastive learning for graph-based semi-supervised learning}.
\newblock \bibinfo{journal}{\emph{arXiv preprint arXiv:2302.09532}} (\bibinfo{year}{2023}).
\newblock


\bibitem[Luo et~al\mbox{.}(2023a)]%
        {luo2023self}
\bibfield{author}{\bibinfo{person}{Xiao Luo}, \bibinfo{person}{Wei Ju}, \bibinfo{person}{Yiyang Gu}, \bibinfo{person}{Zhengyang Mao}, \bibinfo{person}{Luchen Liu}, \bibinfo{person}{Yuhui Yuan}, {and} \bibinfo{person}{Ming Zhang}.} \bibinfo{year}{2023}\natexlab{a}.
\newblock \showarticletitle{Self-supervised graph-level representation learning with adversarial contrastive learning}.
\newblock \bibinfo{journal}{\emph{TKDD}} \bibinfo{volume}{18}, \bibinfo{number}{2} (\bibinfo{year}{2023}), \bibinfo{pages}{1--23}.
\newblock


\bibitem[Luo et~al\mbox{.}(2023b)]%
        {luo2023towards_node}
\bibfield{author}{\bibinfo{person}{Xiao Luo}, \bibinfo{person}{Wei Ju}, \bibinfo{person}{Yiyang Gu}, \bibinfo{person}{Yifang Qin}, \bibinfo{person}{Siyu Yi}, \bibinfo{person}{Daqing Wu}, \bibinfo{person}{Luchen Liu}, {and} \bibinfo{person}{Ming Zhang}.} \bibinfo{year}{2023}\natexlab{b}.
\newblock \showarticletitle{Towards Effective Semi-supervised Node Classification with Hybrid Curriculum Pseudo-labeling}.
\newblock \bibinfo{journal}{\emph{TOMM}} (\bibinfo{year}{2023}).
\newblock


\bibitem[Luo et~al\mbox{.}(2022a)]%
        {luo2022dualgraph}
\bibfield{author}{\bibinfo{person}{Xiao Luo}, \bibinfo{person}{Wei Ju}, \bibinfo{person}{Meng Qu}, \bibinfo{person}{Chong Chen}, \bibinfo{person}{Minghua Deng}, \bibinfo{person}{Xian-Sheng Hua}, {and} \bibinfo{person}{Ming Zhang}.} \bibinfo{year}{2022}\natexlab{a}.
\newblock \showarticletitle{Dualgraph: Improving semi-supervised graph classification via dual contrastive learning}. In \bibinfo{booktitle}{\emph{ICDE}}. \bibinfo{pages}{699--712}.
\newblock


\bibitem[Luo et~al\mbox{.}(2022b)]%
        {luo2022clear}
\bibfield{author}{\bibinfo{person}{Xiao Luo}, \bibinfo{person}{Wei Ju}, \bibinfo{person}{Meng Qu}, \bibinfo{person}{Yiyang Gu}, \bibinfo{person}{Chong Chen}, \bibinfo{person}{Minghua Deng}, \bibinfo{person}{Xian-Sheng Hua}, {and} \bibinfo{person}{Ming Zhang}.} \bibinfo{year}{2022}\natexlab{b}.
\newblock \showarticletitle{Clear: Cluster-enhanced contrast for self-supervised graph representation learning}.
\newblock \bibinfo{journal}{\emph{TNNLS}} (\bibinfo{year}{2022}).
\newblock


\bibitem[Luo et~al\mbox{.}(2022c)]%
        {luo2022deep}
\bibfield{author}{\bibinfo{person}{Xuexiong Luo}, \bibinfo{person}{Jia Wu}, \bibinfo{person}{Jian Yang}, \bibinfo{person}{Shan Xue}, \bibinfo{person}{Hao Peng}, \bibinfo{person}{Chuan Zhou}, \bibinfo{person}{Hongyang Chen}, \bibinfo{person}{Zhao Li}, {and} \bibinfo{person}{Quan~Z Sheng}.} \bibinfo{year}{2022}\natexlab{c}.
\newblock \showarticletitle{Deep graph level anomaly detection with contrastive learning}.
\newblock \bibinfo{journal}{\emph{Scientific Reports}} \bibinfo{volume}{12}, \bibinfo{number}{1} (\bibinfo{year}{2022}), \bibinfo{pages}{19867}.
\newblock


\bibitem[Luo et~al\mbox{.}(2023c)]%
        {luo2023rignn}
\bibfield{author}{\bibinfo{person}{Xiao Luo}, \bibinfo{person}{Yusheng Zhao}, \bibinfo{person}{Zhengyang Mao}, \bibinfo{person}{Yifang Qin}, \bibinfo{person}{Wei Ju}, \bibinfo{person}{Ming Zhang}, {and} \bibinfo{person}{Yizhou Sun}.} \bibinfo{year}{2023}\natexlab{c}.
\newblock \showarticletitle{RIGNN: A Rationale Perspective for Semi-supervised Open-world Graph Classification}.
\newblock \bibinfo{journal}{\emph{TMLR}} (\bibinfo{year}{2023}).
\newblock


\bibitem[Luo et~al\mbox{.}(2023d)]%
        {luo2023towards}
\bibfield{author}{\bibinfo{person}{Xiao Luo}, \bibinfo{person}{Yusheng Zhao}, \bibinfo{person}{Yifang Qin}, \bibinfo{person}{Wei Ju}, {and} \bibinfo{person}{Ming Zhang}.} \bibinfo{year}{2023}\natexlab{d}.
\newblock \showarticletitle{Towards Semi-supervised Universal Graph Classification}.
\newblock \bibinfo{journal}{\emph{TKDE}} (\bibinfo{year}{2023}).
\newblock


\bibitem[Ma et~al\mbox{.}(2022b)]%
        {ma2022towards}
\bibfield{author}{\bibinfo{person}{Guanghui Ma}, \bibinfo{person}{Chunming Hu}, \bibinfo{person}{Ling Ge}, \bibinfo{person}{Junfan Chen}, \bibinfo{person}{Hong Zhang}, {and} \bibinfo{person}{Richong Zhang}.} \bibinfo{year}{2022}\natexlab{b}.
\newblock \showarticletitle{Towards Robust False Information Detection on Social Networks with Contrastive Learning}. In \bibinfo{booktitle}{\emph{CIKM}}. \bibinfo{pages}{1441--1450}.
\newblock


\bibitem[Ma et~al\mbox{.}(2022a)]%
        {ma-etal-2022-open-topic}
\bibfield{author}{\bibinfo{person}{Guanghui Ma}, \bibinfo{person}{Chunming Hu}, \bibinfo{person}{Ling Ge}, {and} \bibinfo{person}{Hong Zhang}.} \bibinfo{year}{2022}\natexlab{a}.
\newblock \showarticletitle{Open-topic false information detection on social networks with contrastive adversarial learning}. In \bibinfo{booktitle}{\emph{EMNLP}}. \bibinfo{pages}{2911--2923}.
\newblock


\bibitem[Manning and Schutze(1999)]%
        {manning1999foundations}
\bibfield{author}{\bibinfo{person}{Christopher Manning} {and} \bibinfo{person}{Hinrich Schutze}.} \bibinfo{year}{1999}\natexlab{}.
\newblock \bibinfo{booktitle}{\emph{Foundations of statistical natural language processing}}.
\newblock \bibinfo{publisher}{MIT press}.
\newblock


\bibitem[Mao et~al\mbox{.}(2021a)]%
        {mao2021simplex}
\bibfield{author}{\bibinfo{person}{Kelong Mao}, \bibinfo{person}{Jieming Zhu}, \bibinfo{person}{Jinpeng Wang}, \bibinfo{person}{Quanyu Dai}, \bibinfo{person}{Zhenhua Dong}, \bibinfo{person}{Xi Xiao}, {and} \bibinfo{person}{Xiuqiang He}.} \bibinfo{year}{2021}\natexlab{a}.
\newblock \showarticletitle{SimpleX: A simple and strong baseline for collaborative filtering}. In \bibinfo{booktitle}{\emph{CIKM}}. \bibinfo{pages}{1243--1252}.
\newblock


\bibitem[Mao et~al\mbox{.}(2021b)]%
        {mao2021ultragcn}
\bibfield{author}{\bibinfo{person}{Kelong Mao}, \bibinfo{person}{Jieming Zhu}, \bibinfo{person}{Xi Xiao}, \bibinfo{person}{Biao Lu}, \bibinfo{person}{Zhaowei Wang}, {and} \bibinfo{person}{Xiuqiang He}.} \bibinfo{year}{2021}\natexlab{b}.
\newblock \showarticletitle{UltraGCN: ultra simplification of graph convolutional networks for recommendation}. In \bibinfo{booktitle}{\emph{CIKM}}. \bibinfo{pages}{1253--1262}.
\newblock


\bibitem[Mao et~al\mbox{.}(2023)]%
        {mao2023rahnet}
\bibfield{author}{\bibinfo{person}{Zhengyang Mao}, \bibinfo{person}{Wei Ju}, \bibinfo{person}{Yifang Qin}, \bibinfo{person}{Xiao Luo}, {and} \bibinfo{person}{Ming Zhang}.} \bibinfo{year}{2023}\natexlab{}.
\newblock \showarticletitle{RAHNet: Retrieval Augmented Hybrid Network for Long-tailed Graph Classification}. In \bibinfo{booktitle}{\emph{ACMMM}}. \bibinfo{pages}{3817--3826}.
\newblock


\bibitem[Mavromatis and Karypis(2020)]%
        {mavromatis2020graph}
\bibfield{author}{\bibinfo{person}{Costas Mavromatis} {and} \bibinfo{person}{George Karypis}.} \bibinfo{year}{2020}\natexlab{}.
\newblock \showarticletitle{Graph infoclust: Leveraging cluster-level node information for unsupervised graph representation learning}.
\newblock \bibinfo{journal}{\emph{arXiv preprint arXiv:2009.06946}} (\bibinfo{year}{2020}).
\newblock


\bibitem[Muttenthaler et~al\mbox{.}(2021)]%
        {muttenthaler2021trends}
\bibfield{author}{\bibinfo{person}{Markus Muttenthaler}, \bibinfo{person}{Glenn~F King}, \bibinfo{person}{David~J Adams}, {and} \bibinfo{person}{Paul~F Alewood}.} \bibinfo{year}{2021}\natexlab{}.
\newblock \showarticletitle{Trends in peptide drug discovery}.
\newblock \bibinfo{journal}{\emph{Nature reviews Drug discovery}} \bibinfo{volume}{20}, \bibinfo{number}{4} (\bibinfo{year}{2021}), \bibinfo{pages}{309--325}.
\newblock


\bibitem[Oord et~al\mbox{.}(2018)]%
        {oord2018representation}
\bibfield{author}{\bibinfo{person}{Aaron van~den Oord}, \bibinfo{person}{Yazhe Li}, {and} \bibinfo{person}{Oriol Vinyals}.} \bibinfo{year}{2018}\natexlab{}.
\newblock \showarticletitle{Representation learning with contrastive predictive coding}.
\newblock \bibinfo{journal}{\emph{arXiv preprint arXiv:1807.03748}} (\bibinfo{year}{2018}).
\newblock


\bibitem[Page et~al\mbox{.}(1998)]%
        {page1998pagerank}
\bibfield{author}{\bibinfo{person}{Lawrence Page}, \bibinfo{person}{Sergey Brin}, \bibinfo{person}{Rajeev Motwani}, {and} \bibinfo{person}{Terry Winograd}.} \bibinfo{year}{1998}\natexlab{}.
\newblock \bibinfo{booktitle}{\emph{The pagerank citation ranking: Bring order to the web}}.
\newblock \bibinfo{type}{{T}echnical {R}eport}. \bibinfo{institution}{Technical report, stanford University}.
\newblock


\bibitem[Pan et~al\mbox{.}(2023)]%
        {pan2023spatial}
\bibfield{author}{\bibinfo{person}{Lin Pan}, \bibinfo{person}{Qianqian Ren}, {and} \bibinfo{person}{Jinbao Li}.} \bibinfo{year}{2023}\natexlab{}.
\newblock \showarticletitle{Spatial-temporal graph contrastive learning for urban traffic flow forecasting}.
\newblock \bibinfo{journal}{\emph{Authorea Preprints}} (\bibinfo{year}{2023}).
\newblock


\bibitem[Park et~al\mbox{.}(2020)]%
        {park2020unsupervised}
\bibfield{author}{\bibinfo{person}{Chanyoung Park}, \bibinfo{person}{Donghyun Kim}, \bibinfo{person}{Jiawei Han}, {and} \bibinfo{person}{Hwanjo Yu}.} \bibinfo{year}{2020}\natexlab{}.
\newblock \showarticletitle{Unsupervised attributed multiplex network embedding}. In \bibinfo{booktitle}{\emph{AAAI}}, Vol.~\bibinfo{volume}{34}. \bibinfo{pages}{5371--5378}.
\newblock


\bibitem[Peng et~al\mbox{.}(2020)]%
        {peng2020self}
\bibfield{author}{\bibinfo{person}{Zhen Peng}, \bibinfo{person}{Yixiang Dong}, \bibinfo{person}{Minnan Luo}, \bibinfo{person}{Xiao-Ming Wu}, {and} \bibinfo{person}{Qinghua Zheng}.} \bibinfo{year}{2020}\natexlab{}.
\newblock \showarticletitle{Self-supervised graph representation learning via global context prediction}.
\newblock \bibinfo{journal}{\emph{arXiv preprint arXiv:2003.01604}} (\bibinfo{year}{2020}).
\newblock


\bibitem[Poole et~al\mbox{.}(2019)]%
        {poole2019variational}
\bibfield{author}{\bibinfo{person}{Ben Poole}, \bibinfo{person}{Sherjil Ozair}, \bibinfo{person}{Aaron Van Den~Oord}, \bibinfo{person}{Alex Alemi}, {and} \bibinfo{person}{George Tucker}.} \bibinfo{year}{2019}\natexlab{}.
\newblock \showarticletitle{On variational bounds of mutual information}. In \bibinfo{booktitle}{\emph{ICML}}. \bibinfo{pages}{5171--5180}.
\newblock


\bibitem[Qian et~al\mbox{.}(2022)]%
        {qian2022co}
\bibfield{author}{\bibinfo{person}{Yiyue Qian}, \bibinfo{person}{Chunhui Zhang}, \bibinfo{person}{Yiming Zhang}, \bibinfo{person}{Qianlong Wen}, \bibinfo{person}{Yanfang Ye}, {and} \bibinfo{person}{Chuxu Zhang}.} \bibinfo{year}{2022}\natexlab{}.
\newblock \showarticletitle{Co-Modality Graph Contrastive Learning for Imbalanced Node Classification}.
\newblock \bibinfo{journal}{\emph{NeurIPS}}  \bibinfo{volume}{35} (\bibinfo{year}{2022}), \bibinfo{pages}{15862--15874}.
\newblock


\bibitem[Qin et~al\mbox{.}(2024a)]%
        {qin2024polycf}
\bibfield{author}{\bibinfo{person}{Yifang Qin}, \bibinfo{person}{Wei Ju}, \bibinfo{person}{Xiao Luo}, \bibinfo{person}{Yiyang Gu}, {and} \bibinfo{person}{Ming Zhang}.} \bibinfo{year}{2024}\natexlab{a}.
\newblock \showarticletitle{PolyCF: Towards the Optimal Spectral Graph Filters for Collaborative Filtering}.
\newblock \bibinfo{journal}{\emph{arXiv preprint arXiv:2401.12590}} (\bibinfo{year}{2024}).
\newblock


\bibitem[Qin et~al\mbox{.}(2024b)]%
        {qin2024learning}
\bibfield{author}{\bibinfo{person}{Yifang Qin}, \bibinfo{person}{Wei Ju}, \bibinfo{person}{Hongjun Wu}, \bibinfo{person}{Xiao Luo}, {and} \bibinfo{person}{Ming Zhang}.} \bibinfo{year}{2024}\natexlab{b}.
\newblock \showarticletitle{Learning graph ODE for continuous-time sequential recommendation}.
\newblock \bibinfo{journal}{\emph{TKDE}} (\bibinfo{year}{2024}).
\newblock


\bibitem[Qin et~al\mbox{.}(2023a)]%
        {qin2023disenpoi}
\bibfield{author}{\bibinfo{person}{Yifang Qin}, \bibinfo{person}{Yifan Wang}, \bibinfo{person}{Fang Sun}, \bibinfo{person}{Wei Ju}, \bibinfo{person}{Xuyang Hou}, \bibinfo{person}{Zhe Wang}, \bibinfo{person}{Jia Cheng}, \bibinfo{person}{Jun Lei}, {and} \bibinfo{person}{Ming Zhang}.} \bibinfo{year}{2023}\natexlab{a}.
\newblock \showarticletitle{DisenPOI: Disentangling Sequential and Geographical Influence for Point-of-Interest Recommendation}. In \bibinfo{booktitle}{\emph{WSDM}}. \bibinfo{pages}{508--516}.
\newblock


\bibitem[Qin et~al\mbox{.}(2023b)]%
        {qin2023diffusion}
\bibfield{author}{\bibinfo{person}{Yifang Qin}, \bibinfo{person}{Hongjun Wu}, \bibinfo{person}{Wei Ju}, \bibinfo{person}{Xiao Luo}, {and} \bibinfo{person}{Ming Zhang}.} \bibinfo{year}{2023}\natexlab{b}.
\newblock \showarticletitle{A diffusion model for poi recommendation}.
\newblock \bibinfo{journal}{\emph{TOIS}} \bibinfo{volume}{42}, \bibinfo{number}{2} (\bibinfo{year}{2023}), \bibinfo{pages}{1--27}.
\newblock


\bibitem[Qiu et~al\mbox{.}(2020)]%
        {qiu2020gcc}
\bibfield{author}{\bibinfo{person}{Jiezhong Qiu}, \bibinfo{person}{Qibin Chen}, \bibinfo{person}{Yuxiao Dong}, \bibinfo{person}{Jing Zhang}, \bibinfo{person}{Hongxia Yang}, \bibinfo{person}{Ming Ding}, \bibinfo{person}{Kuansan Wang}, {and} \bibinfo{person}{Jie Tang}.} \bibinfo{year}{2020}\natexlab{}.
\newblock \showarticletitle{Gcc: Graph contrastive coding for graph neural network pre-training}. In \bibinfo{booktitle}{\emph{KDD}}. \bibinfo{pages}{1150--1160}.
\newblock


\bibitem[Qu et~al\mbox{.}(2023)]%
        {qu2023st}
\bibfield{author}{\bibinfo{person}{Yansong Qu}, \bibinfo{person}{Jian Rong}, \bibinfo{person}{Zhenlong Li}, {and} \bibinfo{person}{Kaiqun Chen}.} \bibinfo{year}{2023}\natexlab{}.
\newblock \showarticletitle{ST-A-PGCL: Spatiotemporal adaptive periodical graph contrastive learning for traffic prediction under real scenarios}.
\newblock \bibinfo{journal}{\emph{Knowledge-Based Systems}}  \bibinfo{volume}{272} (\bibinfo{year}{2023}), \bibinfo{pages}{110591}.
\newblock


\bibitem[Rajadhyaksha and Chitkara(2023)]%
        {rajadhyaksha2023graph}
\bibfield{author}{\bibinfo{person}{Nishant Rajadhyaksha} {and} \bibinfo{person}{Aarushi Chitkara}.} \bibinfo{year}{2023}\natexlab{}.
\newblock \showarticletitle{Graph Contrastive Learning for Multi-omics Data}.
\newblock \bibinfo{journal}{\emph{arXiv preprint arXiv:2301.02242}} (\bibinfo{year}{2023}).
\newblock


\bibitem[Ren et~al\mbox{.}(2023)]%
        {ren2023disentangled}
\bibfield{author}{\bibinfo{person}{Xubin Ren}, \bibinfo{person}{Lianghao Xia}, \bibinfo{person}{Jiashu Zhao}, \bibinfo{person}{Dawei Yin}, {and} \bibinfo{person}{Chao Huang}.} \bibinfo{year}{2023}\natexlab{}.
\newblock \showarticletitle{Disentangled contrastive collaborative filtering}. In \bibinfo{booktitle}{\emph{SIGIR}}. \bibinfo{pages}{1137--1146}.
\newblock


\bibitem[Ren et~al\mbox{.}(2021)]%
        {ren2021label}
\bibfield{author}{\bibinfo{person}{Yuxiang Ren}, \bibinfo{person}{Jiyang Bai}, {and} \bibinfo{person}{Jiawei Zhang}.} \bibinfo{year}{2021}\natexlab{}.
\newblock \showarticletitle{Label contrastive coding based graph neural network for graph classification}. In \bibinfo{booktitle}{\emph{DASFAA}}. \bibinfo{pages}{123--140}.
\newblock


\bibitem[Rohani and Eslahchi(2019)]%
        {rohani2019drug}
\bibfield{author}{\bibinfo{person}{Narjes Rohani} {and} \bibinfo{person}{Changiz Eslahchi}.} \bibinfo{year}{2019}\natexlab{}.
\newblock \showarticletitle{Drug-drug interaction predicting by neural network using integrated similarity}.
\newblock \bibinfo{journal}{\emph{Scientific reports}} \bibinfo{volume}{9}, \bibinfo{number}{1} (\bibinfo{year}{2019}), \bibinfo{pages}{13645}.
\newblock


\bibitem[Rong et~al\mbox{.}(2019)]%
        {rong2019dropedge}
\bibfield{author}{\bibinfo{person}{Yu Rong}, \bibinfo{person}{Wenbing Huang}, \bibinfo{person}{Tingyang Xu}, {and} \bibinfo{person}{Junzhou Huang}.} \bibinfo{year}{2019}\natexlab{}.
\newblock \showarticletitle{Dropedge: Towards deep graph convolutional networks on node classification}.
\newblock \bibinfo{journal}{\emph{arXiv preprint arXiv:1907.10903}} (\bibinfo{year}{2019}).
\newblock


\bibitem[Schiappa et~al\mbox{.}(2023)]%
        {schiappa2023self}
\bibfield{author}{\bibinfo{person}{Madeline~C Schiappa}, \bibinfo{person}{Yogesh~S Rawat}, {and} \bibinfo{person}{Mubarak Shah}.} \bibinfo{year}{2023}\natexlab{}.
\newblock \showarticletitle{Self-supervised learning for videos: A survey}.
\newblock \bibinfo{journal}{\emph{CSUR}} \bibinfo{volume}{55}, \bibinfo{number}{13s} (\bibinfo{year}{2023}), \bibinfo{pages}{1--37}.
\newblock


\bibitem[Schroff et~al\mbox{.}(2015)]%
        {schroff2015facenet}
\bibfield{author}{\bibinfo{person}{Florian Schroff}, \bibinfo{person}{Dmitry Kalenichenko}, {and} \bibinfo{person}{James Philbin}.} \bibinfo{year}{2015}\natexlab{}.
\newblock \showarticletitle{FaceNet: A Unified Embedding for Face Recognition and Clustering}. In \bibinfo{booktitle}{\emph{CVPR}}. \bibinfo{pages}{815--823}.
\newblock


\bibitem[Schultz and Joachims(2003)]%
        {schultz2003learning}
\bibfield{author}{\bibinfo{person}{Matthew Schultz} {and} \bibinfo{person}{Thorsten Joachims}.} \bibinfo{year}{2003}\natexlab{}.
\newblock \showarticletitle{Learning a distance metric from relative comparisons}.
\newblock \bibinfo{journal}{\emph{NeurIPS}}  \bibinfo{volume}{16} (\bibinfo{year}{2003}), \bibinfo{pages}{41--48}.
\newblock


\bibitem[Shah et~al\mbox{.}(2022)]%
        {shah2022max}
\bibfield{author}{\bibinfo{person}{Anshul Shah}, \bibinfo{person}{Suvrit Sra}, \bibinfo{person}{Rama Chellappa}, {and} \bibinfo{person}{Anoop Cherian}.} \bibinfo{year}{2022}\natexlab{}.
\newblock \showarticletitle{Max-margin contrastive learning}. In \bibinfo{booktitle}{\emph{AAAI}}, Vol.~\bibinfo{volume}{36}. \bibinfo{pages}{8220--8230}.
\newblock


\bibitem[Sharma et~al\mbox{.}(2022)]%
        {sharma2022survey}
\bibfield{author}{\bibinfo{person}{Kartik Sharma}, \bibinfo{person}{Yeon-Chang Lee}, \bibinfo{person}{Sivagami Nambi}, \bibinfo{person}{Aditya Salian}, \bibinfo{person}{Shlok Shah}, \bibinfo{person}{Sang-Wook Kim}, {and} \bibinfo{person}{Srijan Kumar}.} \bibinfo{year}{2022}\natexlab{}.
\newblock \showarticletitle{A survey of graph neural networks for social recommender systems}.
\newblock \bibinfo{journal}{\emph{CSUR}} (\bibinfo{year}{2022}).
\newblock


\bibitem[Shen et~al\mbox{.}(2021)]%
        {shen2021powerful}
\bibfield{author}{\bibinfo{person}{Yifei Shen}, \bibinfo{person}{Yongji Wu}, \bibinfo{person}{Yao Zhang}, \bibinfo{person}{Caihua Shan}, \bibinfo{person}{Jun Zhang}, \bibinfo{person}{B~Khaled Letaief}, {and} \bibinfo{person}{Dongsheng Li}.} \bibinfo{year}{2021}\natexlab{}.
\newblock \showarticletitle{How powerful is graph convolution for recommendation?}. In \bibinfo{booktitle}{\emph{CIKM}}. \bibinfo{pages}{1619--1629}.
\newblock


\bibitem[Shi et~al\mbox{.}(2020)]%
        {shi2020graphaf}
\bibfield{author}{\bibinfo{person}{Chence Shi}, \bibinfo{person}{Minkai Xu}, \bibinfo{person}{Zhaocheng Zhu}, \bibinfo{person}{Weinan Zhang}, \bibinfo{person}{Ming Zhang}, {and} \bibinfo{person}{Jian Tang}.} \bibinfo{year}{2020}\natexlab{}.
\newblock \showarticletitle{Graphaf: a flow-based autoregressive model for molecular graph generation}.
\newblock \bibinfo{journal}{\emph{ICLR}} (\bibinfo{year}{2020}).
\newblock


\bibitem[Shwartz-Ziv and LeCun(2023)]%
        {shwartz2023compress}
\bibfield{author}{\bibinfo{person}{Ravid Shwartz-Ziv} {and} \bibinfo{person}{Yann LeCun}.} \bibinfo{year}{2023}\natexlab{}.
\newblock \showarticletitle{To Compress or Not to Compress--Self-Supervised Learning and Information Theory: A Review}.
\newblock \bibinfo{journal}{\emph{arXiv preprint arXiv:2304.09355}} (\bibinfo{year}{2023}).
\newblock


\bibitem[Sun et~al\mbox{.}(2019)]%
        {sun2019infograph}
\bibfield{author}{\bibinfo{person}{Fan-Yun Sun}, \bibinfo{person}{Jordan Hoffmann}, \bibinfo{person}{Vikas Verma}, {and} \bibinfo{person}{Jian Tang}.} \bibinfo{year}{2019}\natexlab{}.
\newblock \showarticletitle{Infograph: Unsupervised and semi-supervised graph-level representation learning via mutual information maximization}.
\newblock \bibinfo{journal}{\emph{arXiv preprint arXiv:1908.01000}} (\bibinfo{year}{2019}).
\newblock


\bibitem[Sun et~al\mbox{.}(2021b)]%
        {sun2021mocl}
\bibfield{author}{\bibinfo{person}{Mengying Sun}, \bibinfo{person}{Jing Xing}, \bibinfo{person}{Huijun Wang}, \bibinfo{person}{Bin Chen}, {and} \bibinfo{person}{Jiayu Zhou}.} \bibinfo{year}{2021}\natexlab{b}.
\newblock \showarticletitle{MoCL: data-driven molecular fingerprint via knowledge-aware contrastive learning from molecular graph}. In \bibinfo{booktitle}{\emph{KDD}}. \bibinfo{pages}{3585--3594}.
\newblock


\bibitem[Sun et~al\mbox{.}(2021a)]%
        {sun2021sugar}
\bibfield{author}{\bibinfo{person}{Qingyun Sun}, \bibinfo{person}{Jianxin Li}, \bibinfo{person}{Hao Peng}, \bibinfo{person}{Jia Wu}, \bibinfo{person}{Yuanxing Ning}, \bibinfo{person}{Philip~S Yu}, {and} \bibinfo{person}{Lifang He}.} \bibinfo{year}{2021}\natexlab{a}.
\newblock \showarticletitle{Sugar: Subgraph neural network with reinforcement pooling and self-supervised mutual information mechanism}. In \bibinfo{booktitle}{\emph{WWW}}. \bibinfo{pages}{2081--2091}.
\newblock


\bibitem[Sun et~al\mbox{.}(2022)]%
        {sun2022rumor}
\bibfield{author}{\bibinfo{person}{Tiening Sun}, \bibinfo{person}{Zhong Qian}, \bibinfo{person}{Sujun Dong}, \bibinfo{person}{Peifeng Li}, {and} \bibinfo{person}{Qiaoming Zhu}.} \bibinfo{year}{2022}\natexlab{}.
\newblock \showarticletitle{Rumor detection on social media with graph adversarial contrastive learning}. In \bibinfo{booktitle}{\emph{WWW}}. \bibinfo{pages}{2789--2797}.
\newblock


\bibitem[Suresh et~al\mbox{.}(2021)]%
        {suresh2021adversarial}
\bibfield{author}{\bibinfo{person}{Susheel Suresh}, \bibinfo{person}{Pan Li}, \bibinfo{person}{Cong Hao}, {and} \bibinfo{person}{Jennifer Neville}.} \bibinfo{year}{2021}\natexlab{}.
\newblock \showarticletitle{Adversarial graph augmentation to improve graph contrastive learning}.
\newblock \bibinfo{journal}{\emph{NeurIPS}}  \bibinfo{volume}{34} (\bibinfo{year}{2021}), \bibinfo{pages}{15920--15933}.
\newblock


\bibitem[Tabassum et~al\mbox{.}(2018)]%
        {tabassum2018social}
\bibfield{author}{\bibinfo{person}{Shazia Tabassum}, \bibinfo{person}{Fabiola~SF Pereira}, \bibinfo{person}{Sofia Fernandes}, {and} \bibinfo{person}{Jo{\~a}o Gama}.} \bibinfo{year}{2018}\natexlab{}.
\newblock \showarticletitle{Social network analysis: An overview}.
\newblock \bibinfo{journal}{\emph{DMKD}} \bibinfo{volume}{8}, \bibinfo{number}{5} (\bibinfo{year}{2018}), \bibinfo{pages}{e1256}.
\newblock


\bibitem[Tao et~al\mbox{.}(2023)]%
        {tao2023prediction}
\bibfield{author}{\bibinfo{person}{Wen Tao}, \bibinfo{person}{Yuansheng Liu}, \bibinfo{person}{Xuan Lin}, \bibinfo{person}{Bosheng Song}, {and} \bibinfo{person}{Xiangxiang Zeng}.} \bibinfo{year}{2023}\natexlab{}.
\newblock \showarticletitle{Prediction of multi-relational drug--gene interaction via Dynamic hyperGraph Contrastive Learning}.
\newblock \bibinfo{journal}{\emph{Briefings in Bioinformatics}} \bibinfo{volume}{24}, \bibinfo{number}{6} (\bibinfo{year}{2023}).
\newblock


\bibitem[Thakoor et~al\mbox{.}(2022)]%
        {thakoor2022large}
\bibfield{author}{\bibinfo{person}{Shantanu Thakoor}, \bibinfo{person}{Corentin Tallec}, \bibinfo{person}{Mohammad~Gheshlaghi Azar}, \bibinfo{person}{Mehdi Azabou}, \bibinfo{person}{Eva~L Dyer}, \bibinfo{person}{Remi Munos}, \bibinfo{person}{Petar Veli{\v{c}}kovi{\'c}}, {and} \bibinfo{person}{Michal Valko}.} \bibinfo{year}{2022}\natexlab{}.
\newblock \showarticletitle{Large-scale representation learning on graphs via bootstrapping}. In \bibinfo{booktitle}{\emph{ICLR}}.
\newblock


\bibitem[Thakoor et~al\mbox{.}(2021)]%
        {thakoor2021bootstrapped}
\bibfield{author}{\bibinfo{person}{Shantanu Thakoor}, \bibinfo{person}{Corentin Tallec}, \bibinfo{person}{Mohammad~Gheshlaghi Azar}, \bibinfo{person}{R{\'e}mi Munos}, \bibinfo{person}{Petar Veli{\v{c}}kovi{\'c}}, {and} \bibinfo{person}{Michal Valko}.} \bibinfo{year}{2021}\natexlab{}.
\newblock \showarticletitle{Bootstrapped representation learning on graphs}. In \bibinfo{booktitle}{\emph{ICLR Workshop}}.
\newblock


\bibitem[Tian et~al\mbox{.}(2023)]%
        {tian2023scgcc}
\bibfield{author}{\bibinfo{person}{Sheng-Wen Tian}, \bibinfo{person}{Jian-Cheng Ni}, \bibinfo{person}{Yu-Tian Wang}, \bibinfo{person}{Chun-Hou Zheng}, {and} \bibinfo{person}{Cun-Mei Ji}.} \bibinfo{year}{2023}\natexlab{}.
\newblock \showarticletitle{scGCC: Graph Contrastive Clustering with Neighborhood Augmentations for scRNA-seq Data Analysis}.
\newblock \bibinfo{journal}{\emph{JBHI}} (\bibinfo{year}{2023}).
\newblock


\bibitem[Tian et~al\mbox{.}(2020a)]%
        {tian2020contrastive}
\bibfield{author}{\bibinfo{person}{Yonglong Tian}, \bibinfo{person}{Dilip Krishnan}, {and} \bibinfo{person}{Phillip Isola}.} \bibinfo{year}{2020}\natexlab{a}.
\newblock \showarticletitle{Contrastive multiview coding}. In \bibinfo{booktitle}{\emph{ECCV}}. \bibinfo{pages}{776--794}.
\newblock


\bibitem[Tian et~al\mbox{.}(2020b)]%
        {tian2020makes}
\bibfield{author}{\bibinfo{person}{Yonglong Tian}, \bibinfo{person}{Chen Sun}, \bibinfo{person}{Ben Poole}, \bibinfo{person}{Dilip Krishnan}, \bibinfo{person}{Cordelia Schmid}, {and} \bibinfo{person}{Phillip Isola}.} \bibinfo{year}{2020}\natexlab{b}.
\newblock \showarticletitle{What makes for good views for contrastive learning?}
\newblock \bibinfo{journal}{\emph{NeurIPS}}  \bibinfo{volume}{33} (\bibinfo{year}{2020}), \bibinfo{pages}{6827--6839}.
\newblock


\bibitem[Tu et~al\mbox{.}(2023)]%
        {tu2023hierarchically}
\bibfield{author}{\bibinfo{person}{Wenxuan Tu}, \bibinfo{person}{Sihang Zhou}, \bibinfo{person}{Xinwang Liu}, \bibinfo{person}{Chunpeng Ge}, \bibinfo{person}{Zhiping Cai}, {and} \bibinfo{person}{Yue Liu}.} \bibinfo{year}{2023}\natexlab{}.
\newblock \showarticletitle{Hierarchically contrastive hard sample mining for graph self-supervised pretraining}.
\newblock \bibinfo{journal}{\emph{TNNLS}} (\bibinfo{year}{2023}).
\newblock


\bibitem[Veli{\v{c}}kovi{\'c} et~al\mbox{.}(2017)]%
        {velivckovic2018graph}
\bibfield{author}{\bibinfo{person}{Petar Veli{\v{c}}kovi{\'c}}, \bibinfo{person}{Guillem Cucurull}, \bibinfo{person}{Arantxa Casanova}, \bibinfo{person}{Adriana Romero}, \bibinfo{person}{Pietro Lio}, {and} \bibinfo{person}{Yoshua Bengio}.} \bibinfo{year}{2017}\natexlab{}.
\newblock \showarticletitle{Graph attention networks}.
\newblock In \bibinfo{booktitle}{\emph{ICLR}}.
\newblock


\bibitem[Veli{\v{c}}kovi{\'c} et~al\mbox{.}(2019)]%
        {velivckovic2019deep}
\bibfield{author}{\bibinfo{person}{Petar Veli{\v{c}}kovi{\'c}}, \bibinfo{person}{William Fedus}, \bibinfo{person}{William~L Hamilton}, \bibinfo{person}{Pietro Li{\`o}}, \bibinfo{person}{Yoshua Bengio}, {and} \bibinfo{person}{R~Devon Hjelm}.} \bibinfo{year}{2019}\natexlab{}.
\newblock \showarticletitle{Deep graph infomax}. In \bibinfo{booktitle}{\emph{ICLR}}.
\newblock


\bibitem[Wan et~al\mbox{.}(2021a)]%
        {wan2021contrastive}
\bibfield{author}{\bibinfo{person}{Sheng Wan}, \bibinfo{person}{Shirui Pan}, \bibinfo{person}{Jian Yang}, {and} \bibinfo{person}{Chen Gong}.} \bibinfo{year}{2021}\natexlab{a}.
\newblock \showarticletitle{Contrastive and generative graph convolutional networks for graph-based semi-supervised learning}. In \bibinfo{booktitle}{\emph{AAAI}}, Vol.~\bibinfo{volume}{35}. \bibinfo{pages}{10049--10057}.
\newblock


\bibitem[Wan et~al\mbox{.}(2021b)]%
        {wan2021contrastive_2}
\bibfield{author}{\bibinfo{person}{Sheng Wan}, \bibinfo{person}{Yibing Zhan}, \bibinfo{person}{Liu Liu}, \bibinfo{person}{Baosheng Yu}, \bibinfo{person}{Shirui Pan}, {and} \bibinfo{person}{Chen Gong}.} \bibinfo{year}{2021}\natexlab{b}.
\newblock \showarticletitle{Contrastive graph poisson networks: Semi-supervised learning with extremely limited labels}.
\newblock \bibinfo{journal}{\emph{NeurIPS}}  \bibinfo{volume}{34} (\bibinfo{year}{2021}), \bibinfo{pages}{6316--6327}.
\newblock


\bibitem[Wang et~al\mbox{.}(2019b)]%
        {wang2019semi}
\bibfield{author}{\bibinfo{person}{Daixin Wang}, \bibinfo{person}{Jianbin Lin}, \bibinfo{person}{Peng Cui}, \bibinfo{person}{Quanhui Jia}, \bibinfo{person}{Zhen Wang}, \bibinfo{person}{Yanming Fang}, \bibinfo{person}{Quan Yu}, \bibinfo{person}{Jun Zhou}, \bibinfo{person}{Shuang Yang}, {and} \bibinfo{person}{Yuan Qi}.} \bibinfo{year}{2019}\natexlab{b}.
\newblock \showarticletitle{A semi-supervised graph attentive network for financial fraud detection}. In \bibinfo{booktitle}{\emph{ICDM}}. \bibinfo{pages}{598--607}.
\newblock


\bibitem[Wang and Liu(2021)]%
        {wang2021understanding}
\bibfield{author}{\bibinfo{person}{Feng Wang} {and} \bibinfo{person}{Huaping Liu}.} \bibinfo{year}{2021}\natexlab{}.
\newblock \showarticletitle{Understanding the behaviour of contrastive loss}. In \bibinfo{booktitle}{\emph{CVPR}}. \bibinfo{pages}{2495--2504}.
\newblock


\bibitem[Wang et~al\mbox{.}(2023a)]%
        {wang2023molecular}
\bibfield{author}{\bibinfo{person}{Jinxian Wang}, \bibinfo{person}{Jihong Guan}, {and} \bibinfo{person}{Shuigeng Zhou}.} \bibinfo{year}{2023}\natexlab{a}.
\newblock \showarticletitle{Molecular property prediction by contrastive learning with attention-guided positive sample selection}.
\newblock \bibinfo{journal}{\emph{Bioinformatics}} \bibinfo{volume}{39}, \bibinfo{number}{5} (\bibinfo{year}{2023}), \bibinfo{pages}{btad258}.
\newblock


\bibitem[Wang et~al\mbox{.}(2024a)]%
        {wang2024assessing}
\bibfield{author}{\bibinfo{person}{Jianfei Wang}, \bibinfo{person}{Cuiqing Jiang}, \bibinfo{person}{Lina Zhou}, {and} \bibinfo{person}{Zhao Wang}.} \bibinfo{year}{2024}\natexlab{a}.
\newblock \showarticletitle{Assessing financial distress of SMEs through event propagation: An adaptive interpretable graph contrastive learning model}.
\newblock \bibinfo{journal}{\emph{Decision Support Systems}} (\bibinfo{year}{2024}), \bibinfo{pages}{114195}.
\newblock


\bibitem[Wang et~al\mbox{.}(2024b)]%
        {wang2024predicting}
\bibfield{author}{\bibinfo{person}{Jingru Wang}, \bibinfo{person}{Yihang Xiao}, \bibinfo{person}{Xuequn Shang}, {and} \bibinfo{person}{Jiajie Peng}.} \bibinfo{year}{2024}\natexlab{b}.
\newblock \showarticletitle{Predicting drug--target binding affinity with cross-scale graph contrastive learning}.
\newblock \bibinfo{journal}{\emph{Briefings in Bioinformatics}} \bibinfo{volume}{25}, \bibinfo{number}{1} (\bibinfo{year}{2024}), \bibinfo{pages}{bbad516}.
\newblock


\bibitem[Wang et~al\mbox{.}(2023b)]%
        {wang2023cross}
\bibfield{author}{\bibinfo{person}{Qizhou Wang}, \bibinfo{person}{Guansong Pang}, \bibinfo{person}{Mahsa Salehi}, \bibinfo{person}{Wray Buntine}, {and} \bibinfo{person}{Christopher Leckie}.} \bibinfo{year}{2023}\natexlab{b}.
\newblock \showarticletitle{Cross-domain graph anomaly detection via anomaly-aware contrastive alignment}. In \bibinfo{booktitle}{\emph{AAAI}}. \bibinfo{pages}{4676--4684}.
\newblock


\bibitem[Wang et~al\mbox{.}(2022c)]%
        {wang2022uncovering}
\bibfield{author}{\bibinfo{person}{Ruijia Wang}, \bibinfo{person}{Xiao Wang}, \bibinfo{person}{Chuan Shi}, {and} \bibinfo{person}{Le Song}.} \bibinfo{year}{2022}\natexlab{c}.
\newblock \showarticletitle{Uncovering the structural fairness in graph contrastive learning}.
\newblock \bibinfo{journal}{\emph{NeurIPS}}  \bibinfo{volume}{35} (\bibinfo{year}{2022}), \bibinfo{pages}{32465--32473}.
\newblock


\bibitem[Wang et~al\mbox{.}(2019a)]%
        {wang2019neural}
\bibfield{author}{\bibinfo{person}{Xiang Wang}, \bibinfo{person}{Xiangnan He}, \bibinfo{person}{Meng Wang}, \bibinfo{person}{Fuli Feng}, {and} \bibinfo{person}{Tat-Seng Chua}.} \bibinfo{year}{2019}\natexlab{a}.
\newblock \showarticletitle{Neural graph collaborative filtering}. In \bibinfo{booktitle}{\emph{SIGIR}}. \bibinfo{pages}{165--174}.
\newblock


\bibitem[Wang et~al\mbox{.}(2021)]%
        {wang2021self}
\bibfield{author}{\bibinfo{person}{Xiao Wang}, \bibinfo{person}{Nian Liu}, \bibinfo{person}{Hui Han}, {and} \bibinfo{person}{Chuan Shi}.} \bibinfo{year}{2021}\natexlab{}.
\newblock \showarticletitle{Self-supervised heterogeneous graph neural network with co-contrastive learning}. In \bibinfo{booktitle}{\emph{KDD}}. \bibinfo{pages}{1726--1736}.
\newblock


\bibitem[Wang et~al\mbox{.}(2022a)]%
        {wang2022improving}
\bibfield{author}{\bibinfo{person}{Yuyang Wang}, \bibinfo{person}{Rishikesh Magar}, \bibinfo{person}{Chen Liang}, {and} \bibinfo{person}{Amir Barati~Farimani}.} \bibinfo{year}{2022}\natexlab{a}.
\newblock \showarticletitle{Improving molecular contrastive learning via faulty negative mitigation and decomposed fragment contrast}.
\newblock \bibinfo{journal}{\emph{JCIM}} \bibinfo{volume}{62}, \bibinfo{number}{11} (\bibinfo{year}{2022}), \bibinfo{pages}{2713--2725}.
\newblock


\bibitem[Wang et~al\mbox{.}(2022b)]%
        {wang2022disencite}
\bibfield{author}{\bibinfo{person}{Yifan Wang}, \bibinfo{person}{Yiping Song}, \bibinfo{person}{Shuai Li}, \bibinfo{person}{Chaoran Cheng}, \bibinfo{person}{Wei Ju}, \bibinfo{person}{Ming Zhang}, {and} \bibinfo{person}{Sheng Wang}.} \bibinfo{year}{2022}\natexlab{b}.
\newblock \showarticletitle{DisenCite: Graph-Based Disentangled Representation Learning for Context-Specific Citation Generation}. In \bibinfo{booktitle}{\emph{AAAI}}, Vol.~\bibinfo{volume}{36}. \bibinfo{pages}{11449--11458}.
\newblock


\bibitem[Wei et~al\mbox{.}(2023)]%
        {wei2023boosting}
\bibfield{author}{\bibinfo{person}{Chunyu Wei}, \bibinfo{person}{Yu Wang}, \bibinfo{person}{Bing Bai}, \bibinfo{person}{Kai Ni}, \bibinfo{person}{David Brady}, {and} \bibinfo{person}{Lu Fang}.} \bibinfo{year}{2023}\natexlab{}.
\newblock \showarticletitle{Boosting graph contrastive learning via graph contrastive saliency}. In \bibinfo{booktitle}{\emph{ICML}}. \bibinfo{pages}{36839--36855}.
\newblock


\bibitem[Wu et~al\mbox{.}(2023)]%
        {wu2023graph}
\bibfield{author}{\bibinfo{person}{Cheng Wu}, \bibinfo{person}{Chaokun Wang}, \bibinfo{person}{Jingcao Xu}, \bibinfo{person}{Ziyang Liu}, \bibinfo{person}{Kai Zheng}, \bibinfo{person}{Xiaowei Wang}, \bibinfo{person}{Yang Song}, {and} \bibinfo{person}{Kun Gai}.} \bibinfo{year}{2023}\natexlab{}.
\newblock \showarticletitle{Graph Contrastive Learning with Generative Adversarial Network}. In \bibinfo{booktitle}{\emph{KDD}}. \bibinfo{pages}{2721--2730}.
\newblock


\bibitem[Wu et~al\mbox{.}(2021b)]%
        {wu2021self}
\bibfield{author}{\bibinfo{person}{Jiancan Wu}, \bibinfo{person}{Xiang Wang}, \bibinfo{person}{Fuli Feng}, \bibinfo{person}{Xiangnan He}, \bibinfo{person}{Liang Chen}, \bibinfo{person}{Jianxun Lian}, {and} \bibinfo{person}{Xing Xie}.} \bibinfo{year}{2021}\natexlab{b}.
\newblock \showarticletitle{Self-supervised graph learning for recommendation}. In \bibinfo{booktitle}{\emph{SIGIR}}. \bibinfo{pages}{726--735}.
\newblock


\bibitem[Wu et~al\mbox{.}(2021a)]%
        {wu2021self_survey}
\bibfield{author}{\bibinfo{person}{Lirong Wu}, \bibinfo{person}{Haitao Lin}, \bibinfo{person}{Cheng Tan}, \bibinfo{person}{Zhangyang Gao}, {and} \bibinfo{person}{Stan~Z Li}.} \bibinfo{year}{2021}\natexlab{a}.
\newblock \showarticletitle{Self-supervised learning on graphs: Contrastive, generative, or predictive}.
\newblock \bibinfo{journal}{\emph{TKDE}} \bibinfo{volume}{35}, \bibinfo{number}{4} (\bibinfo{year}{2021}), \bibinfo{pages}{4216--4235}.
\newblock


\bibitem[Wu et~al\mbox{.}(2022a)]%
        {wu2022attraction}
\bibfield{author}{\bibinfo{person}{Man Wu}, \bibinfo{person}{Shirui Pan}, {and} \bibinfo{person}{Xingquan Zhu}.} \bibinfo{year}{2022}\natexlab{a}.
\newblock \showarticletitle{Attraction and repulsion: Unsupervised domain adaptive graph contrastive learning network}.
\newblock \bibinfo{journal}{\emph{TETCI}} (\bibinfo{year}{2022}).
\newblock


\bibitem[Wu et~al\mbox{.}(2022b)]%
        {wu2022graph}
\bibfield{author}{\bibinfo{person}{Shiwen Wu}, \bibinfo{person}{Fei Sun}, \bibinfo{person}{Wentao Zhang}, \bibinfo{person}{Xu Xie}, {and} \bibinfo{person}{Bin Cui}.} \bibinfo{year}{2022}\natexlab{b}.
\newblock \showarticletitle{Graph neural networks in recommender systems: a survey}.
\newblock \bibinfo{journal}{\emph{CSUR}} \bibinfo{volume}{55}, \bibinfo{number}{5} (\bibinfo{year}{2022}), \bibinfo{pages}{1--37}.
\newblock


\bibitem[Wu et~al\mbox{.}(2019)]%
        {wu2019session}
\bibfield{author}{\bibinfo{person}{Shu Wu}, \bibinfo{person}{Yuyuan Tang}, \bibinfo{person}{Yanqiao Zhu}, \bibinfo{person}{Liang Wang}, \bibinfo{person}{Xing Xie}, {and} \bibinfo{person}{Tieniu Tan}.} \bibinfo{year}{2019}\natexlab{}.
\newblock \showarticletitle{Session-based recommendation with graph neural networks}. In \bibinfo{booktitle}{\emph{AAAI}}, Vol.~\bibinfo{volume}{33}. \bibinfo{pages}{346--353}.
\newblock


\bibitem[Wu et~al\mbox{.}(2020)]%
        {wu2020comprehensive}
\bibfield{author}{\bibinfo{person}{Zonghan Wu}, \bibinfo{person}{Shirui Pan}, \bibinfo{person}{Fengwen Chen}, \bibinfo{person}{Guodong Long}, \bibinfo{person}{Chengqi Zhang}, {and} \bibinfo{person}{Philip~S Yu}.} \bibinfo{year}{2020}\natexlab{}.
\newblock \showarticletitle{A comprehensive survey on graph neural networks}.
\newblock \bibinfo{journal}{\emph{TNNLS}} \bibinfo{volume}{32}, \bibinfo{number}{1} (\bibinfo{year}{2020}), \bibinfo{pages}{4--24}.
\newblock


\bibitem[Xia et~al\mbox{.}(2022b)]%
        {xia2022simgrace}
\bibfield{author}{\bibinfo{person}{Jun Xia}, \bibinfo{person}{Lirong Wu}, \bibinfo{person}{Jintao Chen}, \bibinfo{person}{Bozhen Hu}, {and} \bibinfo{person}{Stan~Z Li}.} \bibinfo{year}{2022}\natexlab{b}.
\newblock \showarticletitle{Simgrace: A simple framework for graph contrastive learning without data augmentation}. In \bibinfo{booktitle}{\emph{WWW}}. \bibinfo{pages}{1070--1079}.
\newblock


\bibitem[Xia et~al\mbox{.}(2022c)]%
        {xia2022progcl}
\bibfield{author}{\bibinfo{person}{Jun Xia}, \bibinfo{person}{Lirong Wu}, \bibinfo{person}{Ge Wang}, \bibinfo{person}{Jintao Chen}, {and} \bibinfo{person}{Stan~Z Li}.} \bibinfo{year}{2022}\natexlab{c}.
\newblock \showarticletitle{ProGCL: Rethinking Hard Negative Mining in Graph Contrastive Learning}. In \bibinfo{booktitle}{\emph{ICML}}. \bibinfo{pages}{24332--24346}.
\newblock


\bibitem[Xia et~al\mbox{.}(2022a)]%
        {xia2022hypergraph}
\bibfield{author}{\bibinfo{person}{Lianghao Xia}, \bibinfo{person}{Chao Huang}, \bibinfo{person}{Yong Xu}, \bibinfo{person}{Jiashu Zhao}, \bibinfo{person}{Dawei Yin}, {and} \bibinfo{person}{Jimmy Huang}.} \bibinfo{year}{2022}\natexlab{a}.
\newblock \showarticletitle{Hypergraph contrastive collaborative filtering}. In \bibinfo{booktitle}{\emph{SIGIR}}. \bibinfo{pages}{70--79}.
\newblock


\bibitem[Xia et~al\mbox{.}(2021)]%
        {xia2021self}
\bibfield{author}{\bibinfo{person}{Xin Xia}, \bibinfo{person}{Hongzhi Yin}, \bibinfo{person}{Junliang Yu}, \bibinfo{person}{Yingxia Shao}, {and} \bibinfo{person}{Lizhen Cui}.} \bibinfo{year}{2021}\natexlab{}.
\newblock \showarticletitle{Self-supervised graph co-training for session-based recommendation}. In \bibinfo{booktitle}{\emph{CIKM}}. \bibinfo{pages}{2180--2190}.
\newblock


\bibitem[Xie et~al\mbox{.}(2023)]%
        {xie2023mtgcl}
\bibfield{author}{\bibinfo{person}{Ming-Yu Xie}, \bibinfo{person}{Shaowu Zhang}, \bibinfo{person}{Tong Zhang}, \bibinfo{person}{Yan Li}, {and} \bibinfo{person}{Xiaodong Cui}.} \bibinfo{year}{2023}\natexlab{}.
\newblock \showarticletitle{MTGCL: Multi-Task Graph Contrastive Learning for Identifying Cancer Driver Genes from Multi-omics Data}.
\newblock \bibinfo{journal}{\emph{bioRxiv}} (\bibinfo{year}{2023}), \bibinfo{pages}{2023--10}.
\newblock


\bibitem[Xie et~al\mbox{.}(2022)]%
        {xie2022self}
\bibfield{author}{\bibinfo{person}{Yaochen Xie}, \bibinfo{person}{Zhao Xu}, \bibinfo{person}{Jingtun Zhang}, \bibinfo{person}{Zhengyang Wang}, {and} \bibinfo{person}{Shuiwang Ji}.} \bibinfo{year}{2022}\natexlab{}.
\newblock \showarticletitle{Self-supervised learning of graph neural networks: A unified review}.
\newblock \bibinfo{journal}{\emph{TPAMI}} \bibinfo{volume}{45}, \bibinfo{number}{2} (\bibinfo{year}{2022}), \bibinfo{pages}{2412--2429}.
\newblock


\bibitem[Xiong et~al\mbox{.}(2023)]%
        {xiong2023scgcl}
\bibfield{author}{\bibinfo{person}{Zehao Xiong}, \bibinfo{person}{Jiawei Luo}, \bibinfo{person}{Wanwan Shi}, \bibinfo{person}{Ying Liu}, \bibinfo{person}{Zhongyuan Xu}, {and} \bibinfo{person}{Bo Wang}.} \bibinfo{year}{2023}\natexlab{}.
\newblock \showarticletitle{scGCL: an imputation method for scRNA-seq data based on graph contrastive learning}.
\newblock \bibinfo{journal}{\emph{Bioinformatics}} \bibinfo{volume}{39}, \bibinfo{number}{3} (\bibinfo{year}{2023}), \bibinfo{pages}{btad098}.
\newblock


\bibitem[Xu et~al\mbox{.}(2023)]%
        {xu2023gan}
\bibfield{author}{\bibinfo{person}{Baowen Xu}, \bibinfo{person}{Xuelei Wang}, \bibinfo{person}{Zhenjie Liu}, {and} \bibinfo{person}{Liwen Kang}.} \bibinfo{year}{2023}\natexlab{}.
\newblock \showarticletitle{A GAN Combined with Graph Contrastive Learning for Traffic Forecasting}. In \bibinfo{booktitle}{\emph{CNIOT}}. \bibinfo{pages}{866--873}.
\newblock


\bibitem[Xu et~al\mbox{.}(2021)]%
        {xu2021infogcl}
\bibfield{author}{\bibinfo{person}{Dongkuan Xu}, \bibinfo{person}{Wei Cheng}, \bibinfo{person}{Dongsheng Luo}, \bibinfo{person}{Haifeng Chen}, {and} \bibinfo{person}{Xiang Zhang}.} \bibinfo{year}{2021}\natexlab{}.
\newblock \showarticletitle{Infogcl: Information-aware graph contrastive learning}.
\newblock \bibinfo{journal}{\emph{NeurIPS}}  \bibinfo{volume}{34} (\bibinfo{year}{2021}), \bibinfo{pages}{30414--30425}.
\newblock


\bibitem[Yang et~al\mbox{.}(2023b)]%
        {yang2023poisoning}
\bibfield{author}{\bibinfo{person}{Junwei Yang}, \bibinfo{person}{Hanwen Xu}, \bibinfo{person}{Srbuhi Mirzoyan}, \bibinfo{person}{Tong Chen}, \bibinfo{person}{Zixuan Liu}, \bibinfo{person}{Wei Ju}, \bibinfo{person}{Luchen Liu}, \bibinfo{person}{Ming Zhang}, {and} \bibinfo{person}{Sheng Wang}.} \bibinfo{year}{2023}\natexlab{b}.
\newblock \showarticletitle{Poisoning scientific knowledge using large language models}.
\newblock \bibinfo{journal}{\emph{bioRxiv}} (\bibinfo{year}{2023}), \bibinfo{pages}{2023--11}.
\newblock


\bibitem[Yang et~al\mbox{.}(2023a)]%
        {yang2023debiased}
\bibfield{author}{\bibinfo{person}{Yuhao Yang}, \bibinfo{person}{Chao Huang}, \bibinfo{person}{Lianghao Xia}, \bibinfo{person}{Chunzhen Huang}, \bibinfo{person}{Da Luo}, {and} \bibinfo{person}{Kangyi Lin}.} \bibinfo{year}{2023}\natexlab{a}.
\newblock \showarticletitle{Debiased Contrastive Learning for Sequential Recommendation}. In \bibinfo{booktitle}{\emph{WWW}}. \bibinfo{pages}{1063--1073}.
\newblock


\bibitem[Yang et~al\mbox{.}(2022)]%
        {yang2022knowledge}
\bibfield{author}{\bibinfo{person}{Yuhao Yang}, \bibinfo{person}{Chao Huang}, \bibinfo{person}{Lianghao Xia}, {and} \bibinfo{person}{Chenliang Li}.} \bibinfo{year}{2022}\natexlab{}.
\newblock \showarticletitle{Knowledge graph contrastive learning for recommendation}. In \bibinfo{booktitle}{\emph{SIGIR}}. \bibinfo{pages}{1434--1443}.
\newblock


\bibitem[Yao et~al\mbox{.}(2023)]%
        {yao2023semi}
\bibfield{author}{\bibinfo{person}{Kainan Yao}, \bibinfo{person}{Xiaowen Wang}, \bibinfo{person}{Wannian Li}, \bibinfo{person}{Hongming Zhu}, \bibinfo{person}{Yizhi Jiang}, \bibinfo{person}{Yulong Li}, \bibinfo{person}{Tongxuan Tian}, \bibinfo{person}{Zhaoyi Yang}, \bibinfo{person}{Qi Liu}, {and} \bibinfo{person}{Qin Liu}.} \bibinfo{year}{2023}\natexlab{}.
\newblock \showarticletitle{Semi-supervised heterogeneous graph contrastive learning for drug--target interaction prediction}.
\newblock \bibinfo{journal}{\emph{Computers in Biology and Medicine}}  \bibinfo{volume}{163} (\bibinfo{year}{2023}), \bibinfo{pages}{107199}.
\newblock


\bibitem[Yi et~al\mbox{.}(2023a)]%
        {yi2023redundancy}
\bibfield{author}{\bibinfo{person}{Siyu Yi}, \bibinfo{person}{Wei Ju}, \bibinfo{person}{Yifang Qin}, \bibinfo{person}{Xiao Luo}, \bibinfo{person}{Luchen Liu}, \bibinfo{person}{Yongdao Zhou}, {and} \bibinfo{person}{Ming Zhang}.} \bibinfo{year}{2023}\natexlab{a}.
\newblock \showarticletitle{Redundancy-Free Self-Supervised Relational Learning for Graph Clustering}.
\newblock \bibinfo{journal}{\emph{TNNLS}} (\bibinfo{year}{2023}).
\newblock


\bibitem[Yi et~al\mbox{.}(2023b)]%
        {yi2023towards}
\bibfield{author}{\bibinfo{person}{Si-Yu Yi}, \bibinfo{person}{Zhengyang Mao}, \bibinfo{person}{Wei Ju}, \bibinfo{person}{Yong-Dao Zhou}, \bibinfo{person}{Luchen Liu}, \bibinfo{person}{Xiao Luo}, {and} \bibinfo{person}{Ming Zhang}.} \bibinfo{year}{2023}\natexlab{b}.
\newblock \showarticletitle{Towards Long-Tailed Recognition for Graph Classification via Collaborative Experts}.
\newblock \bibinfo{journal}{\emph{TBD}} (\bibinfo{year}{2023}).
\newblock


\bibitem[Yi et~al\mbox{.}(2022)]%
        {yi2022multimodal}
\bibfield{author}{\bibinfo{person}{Zixuan Yi}, \bibinfo{person}{Xi Wang}, \bibinfo{person}{Iadh Ounis}, {and} \bibinfo{person}{Craig Macdonald}.} \bibinfo{year}{2022}\natexlab{}.
\newblock \showarticletitle{Multi-modal graph contrastive learning for micro-video recommendation}. In \bibinfo{booktitle}{\emph{SIGIR}}. \bibinfo{pages}{1807--1811}.
\newblock


\bibitem[Yin et~al\mbox{.}(2022a)]%
        {yin2022deal}
\bibfield{author}{\bibinfo{person}{Nan Yin}, \bibinfo{person}{Li Shen}, \bibinfo{person}{Baopu Li}, \bibinfo{person}{Mengzhu Wang}, \bibinfo{person}{Xiao Luo}, \bibinfo{person}{Chong Chen}, \bibinfo{person}{Zhigang Luo}, {and} \bibinfo{person}{Xian-Sheng Hua}.} \bibinfo{year}{2022}\natexlab{a}.
\newblock \showarticletitle{DEAL: An Unsupervised Domain Adaptive Framework for Graph-level Classification}. In \bibinfo{booktitle}{\emph{ACMMM}}. \bibinfo{pages}{3470--3479}.
\newblock


\bibitem[Yin et~al\mbox{.}(2023a)]%
        {yin2023coco}
\bibfield{author}{\bibinfo{person}{Nan Yin}, \bibinfo{person}{Li Shen}, \bibinfo{person}{Mengzhu Wang}, \bibinfo{person}{Long Lan}, \bibinfo{person}{Zeyu Ma}, \bibinfo{person}{Chong Chen}, \bibinfo{person}{Xian-Sheng Hua}, {and} \bibinfo{person}{Xiao Luo}.} \bibinfo{year}{2023}\natexlab{a}.
\newblock \showarticletitle{CoCo: A Coupled Contrastive Framework for Unsupervised Domain Adaptive Graph Classification}. In \bibinfo{booktitle}{\emph{ICML}}. \bibinfo{pages}{40040--40053}.
\newblock


\bibitem[Yin et~al\mbox{.}(2023b)]%
        {yin2023omg}
\bibfield{author}{\bibinfo{person}{Nan Yin}, \bibinfo{person}{Li Shen}, \bibinfo{person}{Mengzhu Wang}, \bibinfo{person}{Xiao Luo}, \bibinfo{person}{Zhigang Luo}, {and} \bibinfo{person}{Dacheng Tao}.} \bibinfo{year}{2023}\natexlab{b}.
\newblock \showarticletitle{OMG: Towards Effective Graph Classification Against Label Noise}.
\newblock \bibinfo{journal}{\emph{TKDE}} (\bibinfo{year}{2023}).
\newblock


\bibitem[Yin et~al\mbox{.}(2022b)]%
        {yin2022autogcl}
\bibfield{author}{\bibinfo{person}{Yihang Yin}, \bibinfo{person}{Qingzhong Wang}, \bibinfo{person}{Siyu Huang}, \bibinfo{person}{Haoyi Xiong}, {and} \bibinfo{person}{Xiang Zhang}.} \bibinfo{year}{2022}\natexlab{b}.
\newblock \showarticletitle{Autogcl: Automated graph contrastive learning via learnable view generators}. In \bibinfo{booktitle}{\emph{AAAI}}, Vol.~\bibinfo{volume}{36}. \bibinfo{pages}{8892--8900}.
\newblock


\bibitem[You et~al\mbox{.}(2021)]%
        {you2021graph}
\bibfield{author}{\bibinfo{person}{Yuning You}, \bibinfo{person}{Tianlong Chen}, \bibinfo{person}{Yang Shen}, {and} \bibinfo{person}{Zhangyang Wang}.} \bibinfo{year}{2021}\natexlab{}.
\newblock \showarticletitle{Graph contrastive learning automated}. In \bibinfo{booktitle}{\emph{ICML}}. \bibinfo{pages}{12121--12132}.
\newblock


\bibitem[You et~al\mbox{.}(2020a)]%
        {you2020graph}
\bibfield{author}{\bibinfo{person}{Yuning You}, \bibinfo{person}{Tianlong Chen}, \bibinfo{person}{Yongduo Sui}, \bibinfo{person}{Ting Chen}, \bibinfo{person}{Zhangyang Wang}, {and} \bibinfo{person}{Yang Shen}.} \bibinfo{year}{2020}\natexlab{a}.
\newblock \showarticletitle{Graph contrastive learning with augmentations}.
\newblock \bibinfo{journal}{\emph{NeurIPS}}  \bibinfo{volume}{33} (\bibinfo{year}{2020}), \bibinfo{pages}{5812--5823}.
\newblock


\bibitem[You et~al\mbox{.}(2020b)]%
        {you2020does}
\bibfield{author}{\bibinfo{person}{Yuning You}, \bibinfo{person}{Tianlong Chen}, \bibinfo{person}{Zhangyang Wang}, {and} \bibinfo{person}{Yang Shen}.} \bibinfo{year}{2020}\natexlab{b}.
\newblock \showarticletitle{When does self-supervision help graph convolutional networks?}. In \bibinfo{booktitle}{\emph{ICML}}. \bibinfo{pages}{10871--10880}.
\newblock


\bibitem[Yu et~al\mbox{.}(2023b)]%
        {yu2023xsimgcl}
\bibfield{author}{\bibinfo{person}{Junliang Yu}, \bibinfo{person}{Xin Xia}, \bibinfo{person}{Tong Chen}, \bibinfo{person}{Lizhen Cui}, \bibinfo{person}{Nguyen Quoc~Viet Hung}, {and} \bibinfo{person}{Hongzhi Yin}.} \bibinfo{year}{2023}\natexlab{b}.
\newblock \showarticletitle{XSimGCL: Towards extremely simple graph contrastive learning for recommendation}.
\newblock \bibinfo{journal}{\emph{TKDE}} (\bibinfo{year}{2023}).
\newblock


\bibitem[Yu et~al\mbox{.}(2021)]%
        {yu2021self}
\bibfield{author}{\bibinfo{person}{Junliang Yu}, \bibinfo{person}{Hongzhi Yin}, \bibinfo{person}{Jundong Li}, \bibinfo{person}{Qinyong Wang}, \bibinfo{person}{Nguyen Quoc~Viet Hung}, {and} \bibinfo{person}{Xiangliang Zhang}.} \bibinfo{year}{2021}\natexlab{}.
\newblock \showarticletitle{Self-supervised multi-channel hypergraph convolutional network for social recommendation}. In \bibinfo{booktitle}{\emph{WWW}}. \bibinfo{pages}{413--424}.
\newblock


\bibitem[Yu et~al\mbox{.}(2022)]%
        {yu2022graph}
\bibfield{author}{\bibinfo{person}{Junliang Yu}, \bibinfo{person}{Hongzhi Yin}, \bibinfo{person}{Xin Xia}, \bibinfo{person}{Tong Chen}, \bibinfo{person}{Lizhen Cui}, {and} \bibinfo{person}{Quoc Viet~Hung Nguyen}.} \bibinfo{year}{2022}\natexlab{}.
\newblock \showarticletitle{Are graph augmentations necessary? simple graph contrastive learning for recommendation}. In \bibinfo{booktitle}{\emph{SIGIR}}. \bibinfo{pages}{1294--1303}.
\newblock


\bibitem[Yu et~al\mbox{.}(2023a)]%
        {yu2023semi}
\bibfield{author}{\bibinfo{person}{Lu Yu}, \bibinfo{person}{Wen Wang}, \bibinfo{person}{Yanbei Liu}, \bibinfo{person}{Xiao Wang}, {and} \bibinfo{person}{Jun Wu}.} \bibinfo{year}{2023}\natexlab{a}.
\newblock \showarticletitle{A Semi-Supervised Graph Neural Network with Confidence Discrimination}. In \bibinfo{booktitle}{\emph{ICPR}}. \bibinfo{pages}{318--323}.
\newblock


\bibitem[Yuan et~al\mbox{.}(2021)]%
        {yuan2021explainability}
\bibfield{author}{\bibinfo{person}{Hao Yuan}, \bibinfo{person}{Haiyang Yu}, \bibinfo{person}{Jie Wang}, \bibinfo{person}{Kang Li}, {and} \bibinfo{person}{Shuiwang Ji}.} \bibinfo{year}{2021}\natexlab{}.
\newblock \showarticletitle{On explainability of graph neural networks via subgraph explorations}. In \bibinfo{booktitle}{\emph{ICML}}. \bibinfo{pages}{12241--12252}.
\newblock


\bibitem[Yuan et~al\mbox{.}(2023a)]%
        {yuan2023alex}
\bibfield{author}{\bibinfo{person}{Jingyang Yuan}, \bibinfo{person}{Xiao Luo}, \bibinfo{person}{Yifang Qin}, \bibinfo{person}{Zhengyang Mao}, \bibinfo{person}{Wei Ju}, {and} \bibinfo{person}{Ming Zhang}.} \bibinfo{year}{2023}\natexlab{a}.
\newblock \showarticletitle{Alex: Towards effective graph transfer learning with noisy labels}. In \bibinfo{booktitle}{\emph{ACMMM}}. \bibinfo{pages}{3647--3656}.
\newblock


\bibitem[Yuan et~al\mbox{.}(2023b)]%
        {yuan2023learning}
\bibfield{author}{\bibinfo{person}{Jingyang Yuan}, \bibinfo{person}{Xiao Luo}, \bibinfo{person}{Yifang Qin}, \bibinfo{person}{Yusheng Zhao}, \bibinfo{person}{Wei Ju}, {and} \bibinfo{person}{Ming Zhang}.} \bibinfo{year}{2023}\natexlab{b}.
\newblock \showarticletitle{Learning on Graphs under Label Noise}. In \bibinfo{booktitle}{\emph{ICASSP}}. \bibinfo{pages}{1--5}.
\newblock


\bibitem[Yuan et~al\mbox{.}(2024)]%
        {yuan2024towards}
\bibfield{author}{\bibinfo{person}{Yige Yuan}, \bibinfo{person}{Bingbing Xu}, \bibinfo{person}{Huawei Shen}, \bibinfo{person}{Qi Cao}, \bibinfo{person}{Keting Cen}, \bibinfo{person}{Wen Zheng}, {and} \bibinfo{person}{Xueqi Cheng}.} \bibinfo{year}{2024}\natexlab{}.
\newblock \showarticletitle{Towards generalizable graph contrastive learning: An information theory perspective}.
\newblock \bibinfo{journal}{\emph{Neural Networks}}  \bibinfo{volume}{172} (\bibinfo{year}{2024}), \bibinfo{pages}{106125}.
\newblock


\bibitem[Yue et~al\mbox{.}(2022)]%
        {yue2022label}
\bibfield{author}{\bibinfo{person}{Han Yue}, \bibinfo{person}{Chunhui Zhang}, \bibinfo{person}{Chuxu Zhang}, {and} \bibinfo{person}{Hongfu Liu}.} \bibinfo{year}{2022}\natexlab{}.
\newblock \showarticletitle{Label-invariant augmentation for semi-supervised graph classification}.
\newblock \bibinfo{journal}{\emph{NeurIPS}}  \bibinfo{volume}{35} (\bibinfo{year}{2022}), \bibinfo{pages}{29350--29361}.
\newblock


\bibitem[Zang et~al\mbox{.}(2023)]%
        {zang2023hierarchical}
\bibfield{author}{\bibinfo{person}{Xuan Zang}, \bibinfo{person}{Xianbing Zhao}, {and} \bibinfo{person}{Buzhou Tang}.} \bibinfo{year}{2023}\natexlab{}.
\newblock \showarticletitle{Hierarchical molecular graph self-supervised learning for property prediction}.
\newblock \bibinfo{journal}{\emph{Communications Chemistry}} \bibinfo{volume}{6}, \bibinfo{number}{1} (\bibinfo{year}{2023}), \bibinfo{pages}{34}.
\newblock


\bibitem[Zbontar et~al\mbox{.}(2021)]%
        {zbontar2021barlow}
\bibfield{author}{\bibinfo{person}{Jure Zbontar}, \bibinfo{person}{Li Jing}, \bibinfo{person}{Ishan Misra}, \bibinfo{person}{Yann LeCun}, {and} \bibinfo{person}{St{\'e}phane Deny}.} \bibinfo{year}{2021}\natexlab{}.
\newblock \showarticletitle{Barlow twins: Self-supervised learning via redundancy reduction}. In \bibinfo{booktitle}{\emph{ICML}}. \bibinfo{pages}{12310--12320}.
\newblock


\bibitem[Zeng et~al\mbox{.}(2023)]%
        {zeng2023imgcl}
\bibfield{author}{\bibinfo{person}{Liang Zeng}, \bibinfo{person}{Lanqing Li}, \bibinfo{person}{Ziqi Gao}, \bibinfo{person}{Peilin Zhao}, {and} \bibinfo{person}{Jian Li}.} \bibinfo{year}{2023}\natexlab{}.
\newblock \showarticletitle{Imgcl: Revisiting graph contrastive learning on imbalanced node classification}. In \bibinfo{booktitle}{\emph{AAAI}}, Vol.~\bibinfo{volume}{37}. \bibinfo{pages}{11138--11146}.
\newblock


\bibitem[Zhang et~al\mbox{.}(2023a)]%
        {zhang2023graph}
\bibfield{author}{\bibinfo{person}{Hangfan Zhang}, \bibinfo{person}{Jinghui Chen}, \bibinfo{person}{Lu Lin}, \bibinfo{person}{Jinyuan Jia}, {and} \bibinfo{person}{Dinghao Wu}.} \bibinfo{year}{2023}\natexlab{a}.
\newblock \showarticletitle{Graph contrastive backdoor attacks}. In \bibinfo{booktitle}{\emph{ICML}}. \bibinfo{pages}{40888--40910}.
\newblock


\bibitem[Zhang et~al\mbox{.}(2022d)]%
        {zhang2022localized}
\bibfield{author}{\bibinfo{person}{Hengrui Zhang}, \bibinfo{person}{Qitian Wu}, \bibinfo{person}{Yu Wang}, \bibinfo{person}{Shaofeng Zhang}, \bibinfo{person}{Junchi Yan}, {and} \bibinfo{person}{Philip~S Yu}.} \bibinfo{year}{2022}\natexlab{d}.
\newblock \showarticletitle{Localized contrastive learning on graphs}.
\newblock \bibinfo{journal}{\emph{arXiv preprint arXiv:2212.04604}} (\bibinfo{year}{2022}).
\newblock


\bibitem[Zhang and Ma(2022)]%
        {zhang2022rethinking}
\bibfield{author}{\bibinfo{person}{Junbo Zhang} {and} \bibinfo{person}{Kaisheng Ma}.} \bibinfo{year}{2022}\natexlab{}.
\newblock \showarticletitle{Rethinking the augmentation module in contrastive learning: Learning hierarchical augmentation invariance with expanded views}. In \bibinfo{booktitle}{\emph{CVPR}}. \bibinfo{pages}{16650--16659}.
\newblock


\bibitem[Zhang and Chen(2018)]%
        {zhang2018link}
\bibfield{author}{\bibinfo{person}{Muhan Zhang} {and} \bibinfo{person}{Yixin Chen}.} \bibinfo{year}{2018}\natexlab{}.
\newblock \showarticletitle{Link prediction based on graph neural networks}.
\newblock \bibinfo{journal}{\emph{NeurIPS}}  \bibinfo{volume}{31} (\bibinfo{year}{2018}), \bibinfo{pages}{5171--5181}.
\newblock


\bibitem[Zhang et~al\mbox{.}(2018)]%
        {zhang2018end}
\bibfield{author}{\bibinfo{person}{Muhan Zhang}, \bibinfo{person}{Zhicheng Cui}, \bibinfo{person}{Marion Neumann}, {and} \bibinfo{person}{Yixin Chen}.} \bibinfo{year}{2018}\natexlab{}.
\newblock \showarticletitle{An end-to-end deep learning architecture for graph classification}. In \bibinfo{booktitle}{\emph{AAAI}}, Vol.~\bibinfo{volume}{32}. \bibinfo{pages}{4438--4445}.
\newblock


\bibitem[Zhang et~al\mbox{.}(2022b)]%
        {zhang2022contrastive}
\bibfield{author}{\bibinfo{person}{Qinggang Zhang}, \bibinfo{person}{Junnan Dong}, \bibinfo{person}{Keyu Duan}, \bibinfo{person}{Xiao Huang}, \bibinfo{person}{Yezi Liu}, {and} \bibinfo{person}{Linchuan Xu}.} \bibinfo{year}{2022}\natexlab{b}.
\newblock \showarticletitle{Contrastive knowledge graph error detection}. In \bibinfo{booktitle}{\emph{CIKM}}. \bibinfo{pages}{2590--2599}.
\newblock


\bibitem[Zhang et~al\mbox{.}(2023c)]%
        {zhang2023automated}
\bibfield{author}{\bibinfo{person}{Qianru Zhang}, \bibinfo{person}{Chao Huang}, \bibinfo{person}{Lianghao Xia}, \bibinfo{person}{Zheng Wang}, \bibinfo{person}{Zhonghang Li}, {and} \bibinfo{person}{Siuming Yiu}.} \bibinfo{year}{2023}\natexlab{c}.
\newblock \showarticletitle{Automated spatio-temporal graph contrastive learning}. In \bibinfo{booktitle}{\emph{WWW}}. \bibinfo{pages}{295--305}.
\newblock


\bibitem[Zhang et~al\mbox{.}(2023d)]%
        {zhang2023spatial}
\bibfield{author}{\bibinfo{person}{Qianru Zhang}, \bibinfo{person}{Chao Huang}, \bibinfo{person}{Lianghao Xia}, \bibinfo{person}{Zheng Wang}, \bibinfo{person}{Siu~Ming Yiu}, {and} \bibinfo{person}{Ruihua Han}.} \bibinfo{year}{2023}\natexlab{d}.
\newblock \showarticletitle{Spatial-temporal graph learning with adversarial contrastive adaptation}. In \bibinfo{booktitle}{\emph{ICML}}. \bibinfo{pages}{41151--41163}.
\newblock


\bibitem[Zhang et~al\mbox{.}(2022a)]%
        {zhang2022unsupervised}
\bibfield{author}{\bibinfo{person}{Sixiao Zhang}, \bibinfo{person}{Hongxu Chen}, \bibinfo{person}{Xiangguo Sun}, \bibinfo{person}{Yicong Li}, {and} \bibinfo{person}{Guandong Xu}.} \bibinfo{year}{2022}\natexlab{a}.
\newblock \showarticletitle{Unsupervised graph poisoning attack via contrastive loss back-propagation}. In \bibinfo{booktitle}{\emph{WWW}}. \bibinfo{pages}{1322--1330}.
\newblock


\bibitem[Zhang et~al\mbox{.}(2024)]%
        {zhang2024motif}
\bibfield{author}{\bibinfo{person}{Shichang Zhang}, \bibinfo{person}{Ziniu Hu}, \bibinfo{person}{Arjun Subramonian}, {and} \bibinfo{person}{Yizhou Sun}.} \bibinfo{year}{2024}\natexlab{}.
\newblock \showarticletitle{Motif-driven contrastive learning of graph representations}.
\newblock \bibinfo{journal}{\emph{TKDE}} (\bibinfo{year}{2024}).
\newblock


\bibitem[Zhang et~al\mbox{.}(2023b)]%
        {zhang2023contrastive}
\bibfield{author}{\bibinfo{person}{Yifei Zhang}, \bibinfo{person}{Yankai Chen}, \bibinfo{person}{Zixing Song}, {and} \bibinfo{person}{Irwin King}.} \bibinfo{year}{2023}\natexlab{b}.
\newblock \showarticletitle{Contrastive cross-scale graph knowledge synergy}. In \bibinfo{booktitle}{\emph{KDD}}. \bibinfo{pages}{3422--3433}.
\newblock


\bibitem[Zhang et~al\mbox{.}(2022c)]%
        {zhang2022protgnn}
\bibfield{author}{\bibinfo{person}{Zaixi Zhang}, \bibinfo{person}{Qi Liu}, \bibinfo{person}{Hao Wang}, \bibinfo{person}{Chengqiang Lu}, {and} \bibinfo{person}{Cheekong Lee}.} \bibinfo{year}{2022}\natexlab{c}.
\newblock \showarticletitle{Protgnn: Towards self-explaining graph neural networks}. In \bibinfo{booktitle}{\emph{AAAI}}, Vol.~\bibinfo{volume}{36}. \bibinfo{pages}{9127--9135}.
\newblock


\bibitem[Zhao et~al\mbox{.}(2023)]%
        {zhao2023dynamic}
\bibfield{author}{\bibinfo{person}{Yusheng Zhao}, \bibinfo{person}{Xiao Luo}, \bibinfo{person}{Wei Ju}, \bibinfo{person}{Chong Chen}, \bibinfo{person}{Xian-Sheng Hua}, {and} \bibinfo{person}{Ming Zhang}.} \bibinfo{year}{2023}\natexlab{}.
\newblock \showarticletitle{Dynamic hypergraph structure learning for traffic flow forecasting}. In \bibinfo{booktitle}{\emph{ICDE}}. \bibinfo{pages}{2303--2316}.
\newblock


\bibitem[Zheng et~al\mbox{.}(2024)]%
        {zheng2024subgraph}
\bibfield{author}{\bibinfo{person}{Jiahao Zheng}, \bibinfo{person}{Yuedong Yang}, {and} \bibinfo{person}{Zhiming Dai}.} \bibinfo{year}{2024}\natexlab{}.
\newblock \showarticletitle{Subgraph extraction and graph representation learning for single cell Hi-C imputation and clustering}.
\newblock \bibinfo{journal}{\emph{Briefings in Bioinformatics}} \bibinfo{volume}{25}, \bibinfo{number}{1} (\bibinfo{year}{2024}), \bibinfo{pages}{bbad379}.
\newblock


\bibitem[Zheng et~al\mbox{.}(2023)]%
        {zheng2023casangcl}
\bibfield{author}{\bibinfo{person}{Zixi Zheng}, \bibinfo{person}{Yanyan Tan}, \bibinfo{person}{Hong Wang}, \bibinfo{person}{Shengpeng Yu}, \bibinfo{person}{Tianyu Liu}, {and} \bibinfo{person}{Cheng Liang}.} \bibinfo{year}{2023}\natexlab{}.
\newblock \showarticletitle{CasANGCL: pre-training and fine-tuning model based on cascaded attention network and graph contrastive learning for molecular property prediction}.
\newblock \bibinfo{journal}{\emph{Briefings in Bioinformatics}} \bibinfo{volume}{24}, \bibinfo{number}{1} (\bibinfo{year}{2023}), \bibinfo{pages}{bbac566}.
\newblock


\bibitem[Zhou et~al\mbox{.}(2023a)]%
        {zhou2023smgcl}
\bibfield{author}{\bibinfo{person}{Hui Zhou}, \bibinfo{person}{Maoguo Gong}, \bibinfo{person}{Shanfeng Wang}, \bibinfo{person}{Yuan Gao}, {and} \bibinfo{person}{Zhongying Zhao}.} \bibinfo{year}{2023}\natexlab{a}.
\newblock \showarticletitle{Smgcl: Semi-supervised multi-view graph contrastive learning}.
\newblock \bibinfo{journal}{\emph{Knowledge-Based Systems}}  \bibinfo{volume}{260} (\bibinfo{year}{2023}), \bibinfo{pages}{110120}.
\newblock


\bibitem[Zhou et~al\mbox{.}(2023b)]%
        {zhou2023detecting}
\bibfield{author}{\bibinfo{person}{Ming Zhou}, \bibinfo{person}{Dan Zhang}, \bibinfo{person}{Yuandong Wang}, \bibinfo{person}{Yangli-Ao Geng}, {and} \bibinfo{person}{Jie Tang}.} \bibinfo{year}{2023}\natexlab{b}.
\newblock \showarticletitle{Detecting Social Bot on the Fly using Contrastive Learning}. In \bibinfo{booktitle}{\emph{CIKM}}. \bibinfo{pages}{4995--5001}.
\newblock


\bibitem[Zhu et~al\mbox{.}(2021c)]%
        {zhu2021transfer}
\bibfield{author}{\bibinfo{person}{Qi Zhu}, \bibinfo{person}{Carl Yang}, \bibinfo{person}{Yidan Xu}, \bibinfo{person}{Haonan Wang}, \bibinfo{person}{Chao Zhang}, {and} \bibinfo{person}{Jiawei Han}.} \bibinfo{year}{2021}\natexlab{c}.
\newblock \showarticletitle{Transfer learning of graph neural networks with ego-graph information maximization}.
\newblock \bibinfo{journal}{\emph{NeurIPS}}  \bibinfo{volume}{34} (\bibinfo{year}{2021}), \bibinfo{pages}{1766--1779}.
\newblock


\bibitem[Zhu et~al\mbox{.}(2022)]%
        {zhu2022survey}
\bibfield{author}{\bibinfo{person}{Yanqiao Zhu}, \bibinfo{person}{Yuanqi Du}, \bibinfo{person}{Yinkai Wang}, \bibinfo{person}{Yichen Xu}, \bibinfo{person}{Jieyu Zhang}, \bibinfo{person}{Qiang Liu}, {and} \bibinfo{person}{Shu Wu}.} \bibinfo{year}{2022}\natexlab{}.
\newblock \showarticletitle{A survey on deep graph generation: Methods and applications}. In \bibinfo{booktitle}{\emph{LoG}}. \bibinfo{pages}{47--1}.
\newblock


\bibitem[Zhu et~al\mbox{.}(2023)]%
        {zhu2023graphcontrol}
\bibfield{author}{\bibinfo{person}{Yun Zhu}, \bibinfo{person}{Yaoke Wang}, \bibinfo{person}{Haizhou Shi}, \bibinfo{person}{Zhenshuo Zhang}, {and} \bibinfo{person}{Siliang Tang}.} \bibinfo{year}{2023}\natexlab{}.
\newblock \showarticletitle{GraphControl: Adding Conditional Control to Universal Graph Pre-trained Models for Graph Domain Transfer Learning}.
\newblock \bibinfo{journal}{\emph{arXiv preprint arXiv:2310.07365}} (\bibinfo{year}{2023}).
\newblock


\bibitem[Zhu et~al\mbox{.}(2021a)]%
        {zhu2021empirical}
\bibfield{author}{\bibinfo{person}{Yanqiao Zhu}, \bibinfo{person}{Yichen Xu}, \bibinfo{person}{Qiang Liu}, {and} \bibinfo{person}{Shu Wu}.} \bibinfo{year}{2021}\natexlab{a}.
\newblock \showarticletitle{An Empirical Study of Graph Contrastive Learning}. In \bibinfo{booktitle}{\emph{NeurIPS Datasets and Benchmarks}}.
\newblock


\bibitem[Zhu et~al\mbox{.}(2020)]%
        {zhu2020deep}
\bibfield{author}{\bibinfo{person}{Yanqiao Zhu}, \bibinfo{person}{Yichen Xu}, \bibinfo{person}{Feng Yu}, \bibinfo{person}{Qiang Liu}, \bibinfo{person}{Shu Wu}, {and} \bibinfo{person}{Liang Wang}.} \bibinfo{year}{2020}\natexlab{}.
\newblock \showarticletitle{Deep graph contrastive representation learning}.
\newblock \bibinfo{journal}{\emph{arXiv preprint arXiv:2006.04131}} (\bibinfo{year}{2020}).
\newblock


\bibitem[Zhu et~al\mbox{.}(2021b)]%
        {zhu2021graph}
\bibfield{author}{\bibinfo{person}{Yanqiao Zhu}, \bibinfo{person}{Yichen Xu}, \bibinfo{person}{Feng Yu}, \bibinfo{person}{Qiang Liu}, \bibinfo{person}{Shu Wu}, {and} \bibinfo{person}{Liang Wang}.} \bibinfo{year}{2021}\natexlab{b}.
\newblock \showarticletitle{Graph contrastive learning with adaptive augmentation}. In \bibinfo{booktitle}{\emph{WWW}}. \bibinfo{pages}{2069--2080}.
\newblock


\bibitem[Zong et~al\mbox{.}(2022)]%
        {zong2022const}
\bibfield{author}{\bibinfo{person}{Yongshuo Zong}, \bibinfo{person}{Tingyang Yu}, \bibinfo{person}{Xuesong Wang}, \bibinfo{person}{Yixuan Wang}, \bibinfo{person}{Zhihang Hu}, {and} \bibinfo{person}{Yu Li}.} \bibinfo{year}{2022}\natexlab{}.
\newblock \showarticletitle{conST: an interpretable multi-modal contrastive learning framework for spatial transcriptomics}.
\newblock \bibinfo{journal}{\emph{bioRxiv}} (\bibinfo{year}{2022}), \bibinfo{pages}{2022--01}.
\newblock


\bibitem[Z{\"u}gner et~al\mbox{.}(2018)]%
        {zugner2018adversarial}
\bibfield{author}{\bibinfo{person}{Daniel Z{\"u}gner}, \bibinfo{person}{Amir Akbarnejad}, {and} \bibinfo{person}{Stephan G{\"u}nnemann}.} \bibinfo{year}{2018}\natexlab{}.
\newblock \showarticletitle{Adversarial attacks on neural networks for graph data}. In \bibinfo{booktitle}{\emph{KDD}}. \bibinfo{pages}{2847--2856}.
\newblock


\end{thebibliography}


\end{document}